%% file: VQA.tex
\documentclass[preprint,5p,times,twocolumn]{elsarticle}

\usepackage{amssymb}
\usepackage[framemethod=tikz]{mdframed}
\usepackage{amsfonts} 
\usepackage{lipsum}  
\usepackage{amsmath} 
\usepackage{tabularx}
\usepackage{array}
\usepackage{ragged2e}
\usepackage{multirow}
\usepackage{colortbl}
\usepackage{rotating}
\usepackage{booktabs}
\usepackage{lscape}
\usepackage{pdflscape}
\usepackage{footnote}
\usepackage{threeparttable}
\usepackage{soul}
\usepackage{subcaption}

\journal{Information Fusion}

\begin{document}

\begin{frontmatter}



\title{From Image to Language: A Critical Analysis of Visual Question Answering (VQA) Approaches, Challenges, and Opportunities}


\author[label1,label2]{Md Farhan Ishmam}
\ead{farhanishmam@iut-dhaka.edu}
\author[label2,label3] {Md Sakib Hossain Shovon}
\ead{sakib.aiub.cs@gmail.com}
\author[label2,label3]{M.F. Mridha}
\ead{firoz.mridha@aiub.edu}
\author[label4]{and Nilanjan Dey}
\ead{nilanjan.dey@tint.edu.in}

\affiliation[label1]{organization={Department of Computer Science and Engineering, Islamic University of Technology},       
city={Dhaka},
            country={Bangladesh}}
            
\affiliation[label2]{organization={Advanced Machine Intelligence Research Lab},
            city={Dhaka},
            country={Bangladesh}}

\affiliation[label3]{organization={Department of Computer Science and Engineering, American International University},
            city={Dhaka},
            country={Bangladesh}}

\affiliation[label4]{organization={Department of Computer Science and Engineering, Techno International New Town},
            city={Kolkata},
            country={India}}

\begin{abstract}
The multimodal task of Visual Question Answering (VQA) encompassing elements of Computer Vision (CV) and Natural Language Processing (NLP), aims to generate answers to questions on any visual input. Over time, the scope of VQA has expanded from datasets focusing on an extensive collection of natural images to datasets featuring synthetic images, video, 3D environments, and various other visual inputs. The emergence of large pre-trained networks has shifted the early VQA approaches relying on feature extraction and fusion schemes to vision language pre-training (VLP) techniques. However, there is a lack of comprehensive surveys that encompass both traditional VQA architectures and contemporary VLP-based methods. Furthermore, the VLP challenges in the lens of VQA haven't been thoroughly explored, leaving room for potential open problems to emerge. Our work presents a survey in the domain of VQA that delves into the intricacies of VQA datasets and methods over the field's history, introduces a detailed taxonomy to categorize the facets of VQA, and highlights the recent trends, challenges, and scopes for improvement. We further generalize VQA to multimodal question answering, explore tasks related to VQA, and present a set of open problems for future investigation. The work aims to navigate both beginners and experts by shedding light on the potential avenues of research and expanding the boundaries of the field.
\end{abstract}




\end{frontmatter}

\section{Introduction}
\label{sec:Intro}

Over the past decade, advancements in deep learning-based systems led to breakthroughs in visual and textual comprehension enabling AI models to rival human performance in these domains. While understanding an image has been a domain of expertise for humans, it hasn't been so for AI. An AI-generated answer has been viewed as lackluster and simplistic. However, in recent times, the ability to distinguish between human-generated and AI-generated texts has become increasingly difficult, exemplified by the fact that the specific line in question has been crafted by an AI system. Visual Question Answering (VQA) is an AI problem that inherited all its principles and methodologies from the realms of Computer Vision (CV) and Natural Language Processing (NLP). But as VQA evolved, it carved its own identity with unique traits and nuances.  

Visual Question Answering (VQA) has been \textit{traditionally} defined as the problem of answering a question with an image as the context \cite{antol2015vqa}. The current scope of VQA is not limited to a single image as the visual input but can be generalized to any form of visual input e.g. set of images \cite{bansal2020visualImageSet} or videos \cite{xu2017videoCaptionQA,zhong2022video}. However, the term VQA is interchangeably used to define VQA on a single image and generalized visual input. Being a multimodal domain, visual question answering naturally works with visual and textual modalities but also works with auditory input as seen in multimodal video question answering \cite{lei2018tvqa}. VQA is also closely related to other vision-language problems starting from early works like image-sentence retrieval \cite{mezaris2003ontology} to more recent tasks -- Visual Reasoning \cite{zellers2019recognitionVCR}, Image Captioning \cite{hossain2019comprehensiveImageCaptioning}, Visual Dialogue \cite{das2017visualDialog}, etc. 

\begin{figure*}[ht]
    \includegraphics[width=\linewidth]{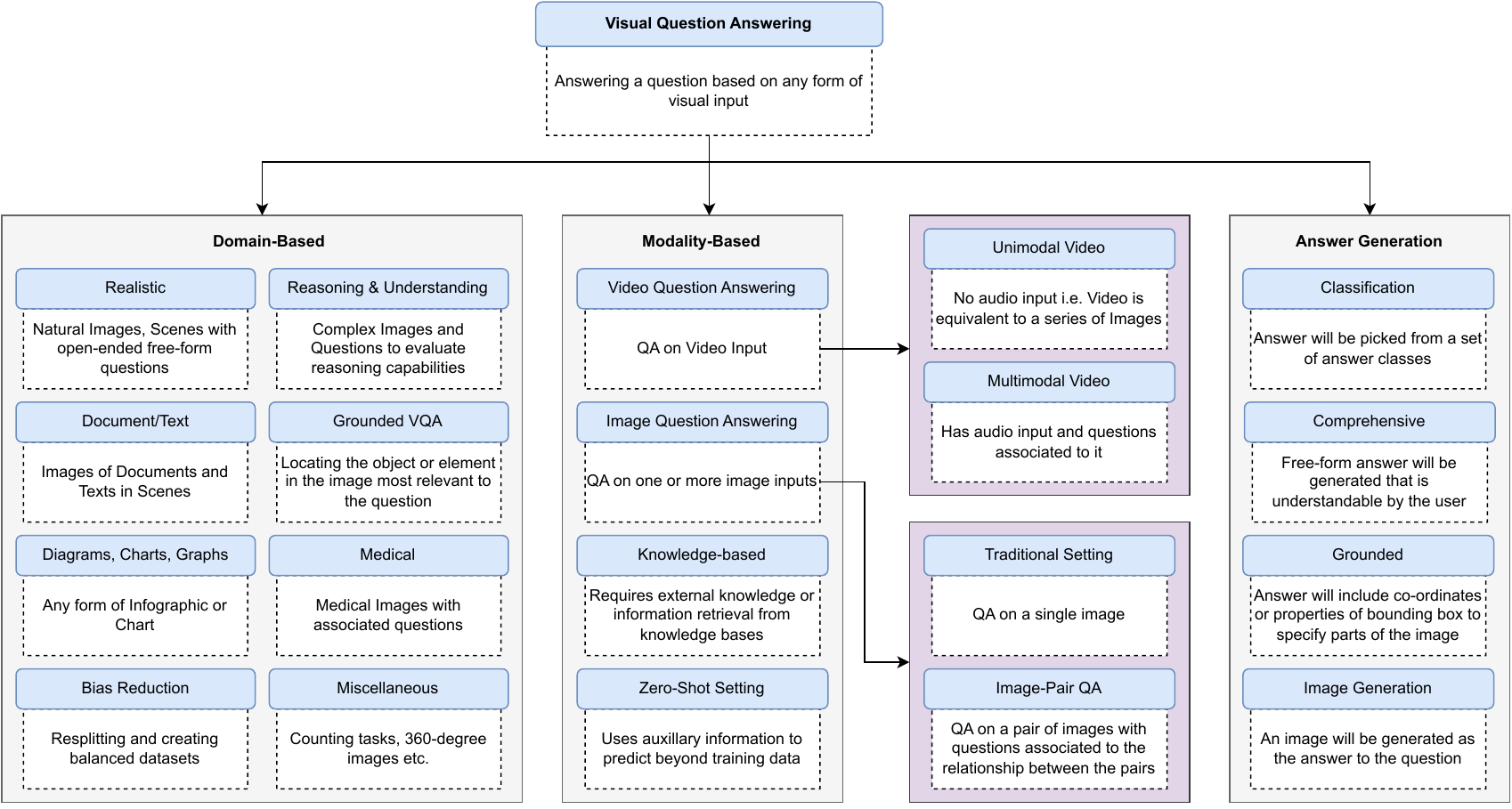}
    \caption{Taxonomy and definitions of several VQA tasks based on domain (where and how VQA is applied), the modality (type and source of information), and answer generation (type of output and how is produced).}
    \label{fig:vqaTasks}
\end{figure*}

In the early days, VQA struggled with defining a dataset that could be used as a benchmark for the problem. Initially, the problem definitions were restrictive but were later expanded to free-form open-ended question answering with the advent of the VQA dataset \cite{antol2015vqa}, regarded as the first standard dataset in this domain. VQA 2.0 \cite{goyal2017making}, the successor to the VQA dataset, also gained immense popularity while addressing some of the major limitations of the VQA dataset. The subsequent datasets increased the task complexity by requiring a high level of reasoning \cite{johnson2017clevr, hudson2019gqa, zhang2019raven}, incorporated external knowledge \cite{wang2015explicit, wang2017fvqa, schwenk2022okvqa}, and extended the task to different variations of visual input \cite{xu2017videoCaptionQA, methani2020plotqa, mishra2019ocr, biten2019scene}. Recently VQA datasets saw substantial work in Video Question Answering \cite{zhong2022video}, Medical VQA \cite{lin2023medical}, and datasets on plots, figures, and graphs \cite{mathew2021docvqa, masry2022chartqa, mathew2022infographicvqa}. 

The VQA methodologies have also undergone several phases but have permanently shifted to deep learning-based methods. The earlier methods \cite{antol2015vqa, ren2015exploring, malinowski2015ask} primarily relied on a visual and textual encoder to extract features from the multimodal inputs followed by some form of fusion strategy to combine the encodings. The fused output is then passed to a classifier or generator depending on answer generation being treated as a classification \cite{ren2015exploring} or generative problem \cite{malinowski2015ask}. Modern VQA methods have shifted from this joint encoding scheme to Vision Language Pre-training (VLP) \cite{gan2022visionVLPSurvey, chen2023vlpSurvey} employing transformer \cite{vaswani2017attention} architectures trained on generalized tasks with large image-text pair datasets and then fine-tuned to downstream tasks like VQA \cite{li2019visualbert, li2020unicoder, li2020oscar}. 

We believe that the rapidly evolving field of VQA is far from saturation and has plenty of open problems, research challenges, and scope for improvement. While existing VQA surveys thoroughly explored pre-VLP architectures \cite{kafle2017visual, wu2017visualSurvey}, contemporary VLP surveys \cite{gan2022visionVLPSurvey, chen2023vlpSurvey} do not explicitly delve into the domain through the lens of VQA. Our survey bridges the gap by extending the traditional VQA surveys to encompass VLP techniques, thereby providing a comprehensive overview of the whole domain. In this review, we introduced a comprehensive taxonomy of VQA problems, datasets, and methods in order to organize the vast amount of research work throughout the years. In Section-\ref{sec:Applications}, we explore various applications of VQA in visually impaired assistance, medical domain, educational domain, visual chatbots, etc. Section-\ref{sec:defScope} defines the problem of VQA and the scope of the domain. Then, section-\ref{sec:surveys} delves into the existing generalized surveys in VQA along with specialized VQA surveys on fusion techniques, language bias, video QA, etc., and related surveys in VLP, Image Captioning, Multimodal Machine Learning, Zero-Shot Learning, etc. 

Our work aspires to serve as a beginner's roadmap to VQA by highlighting the major works in VQA datasets, methods, and metrics over the last decade covered in section-\ref{sec:Datasets}, \ref{sec:Method}, and \ref{sec:evaluationMetrics} respectively. The review also aims to navigate future researchers by directing them toward the research challenges faced in modern VQA as seen in section \ref{sec:Challenges}. Furthermore, section-\ref{sec:vqaBiggerPicture} positions VQA in the broader domain of multimodal learning and explores related domains and sub-domains. Finally, section-\ref{sec:trendsOpenProblemsFuture} highlights the trends, presents unsolved open problems, and discusses the future of the domain. We hope to see this survey leading to potentially ground-breaking work in VQA and its related domains. 

\section{VQA Applications}

\begin{figure}[ht]
    \includegraphics[width=\linewidth]{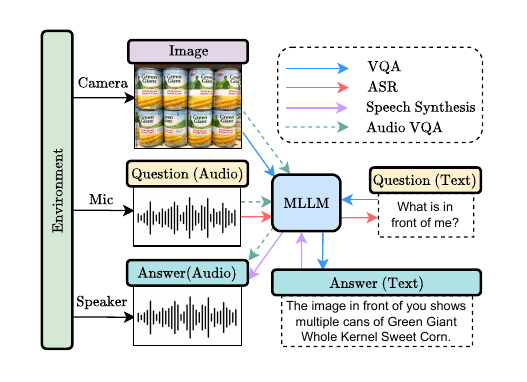}
    \caption{Overview of a visually impaired assistance system similar to the VizWiz Mobile App \cite{bigham2010vizwiz} utilizing Multimodal Large Language Models (MLLMs). The app allows visually impaired users to take pictures of their environment along with a voice-recorded question. Automatic Speech Recognition (ASR) processes the audio input, generating a textual question that is sent to the MLLM along with the image. The MLLM then produces a comprehensive answer, which is converted back to audio through Speech Synthesis. Alternatively, the audio can be sent directly to the MLLM for Audio VQA, and dedicated modules for ASR, Speech Synthesis, and VQA can also be used separately.}
    \label{fig:vizWiz}
\end{figure}

\label{sec:Applications}
While the current applications of VQA methodologies are scarce, there are a lot of potential areas that can improve significantly using VQA. Real-world applications of VQA with its emerging use cases have been extensively studied by \citet{barra2021visual}. In the subsequent subsections, we shall explore some of the domains that use VQA.

\subsection{Assistance of the Visually Impaired and VizWiz Challenges}
\label{sec:assistVisually}
Vision can be considered the most important and frequently utilized sensor of the human body. Assistive technologies for the visually impaired \cite{bigham2010vizwiz} have been a field of interest even before the popularity of automatic VQA systems. Inspired by computer vision datasets and techniques, the continuous research enabled the creation of the VizWiz dataset by \citet{gurari2018vizwiz}. The primary purpose of the dataset is to train VQA models that can automatically provide assistance to visually impaired people by answering questions on real-world images based on data collected by a mobile application. 

Works on visual challenges faced by visually impaired people date back to 2010 when \citet{bigham2010vizwiz} proposed the VizWiz mobile application that relied on Amazon Mechanical Turk workers to manually respond to visual questions. The users will send an image and voice-recorded question to the server which will respond with an answer produced by the workers. Although the task of question answering was done manually, it showcased the necessity of an automated VQA system. The VizWiz organization established a common goal of assisting visually impaired people through the development of datasets and algorithms for assistive technologies and carried out substantial research work \cite{gurari2018vizwiz, gurari2019vizwiz, tseng2022vizwiz} in the following decade.

The launch of the VizWiz app was followed by several similar works for the visually impaired e.g. perceiving fashion \cite{burton2012crowdsourcing} which deals with answering subjective fashion-based questions, exploring the visual challenges faced in daily life \cite{brady2013visual} through the creation of a dataset with more than 40k visual questions on 5k visually impaired users, and assisting users on a video stream in longer conversations \cite{lasecki2013answering} by highlighting the problem of Video Question Answering. While the works are prominent in the pre-deep learning VQA era, they strongly advocated for the case of development of an automatic system in this domain and subsequently was introduced by \citet{gurari2017crowdverge} through Crowdverse, an automatic system for predicting the answerability of a visual question. As visually impaired people have no knowledge of the image taken by the app, the answerability of that particular image is the first problem in creating a completely automatic system. 

In the following year, \citet{gurari2018vizwiz} proposed the VizWiz grand challenge through the creation of the VizWiz dataset with two tasks - answerability of a visual question from the VizWiz app and automatically generating an answer if the question is answerable. With more than 30k visual questions, the VizWiz dataset paved the path for subsequent VQA models that aimed to assist the visually impaired. The challenge had been substantially difficult compared to other VQA problems as the images taken in the app do no not have ideal conditions. A standard image from the traditional datasets like VQAv2 \cite{goyal2017making}  will be clear with good lighting conditions while the images from the VizWiz dataset can be blurry, noisy, have unfavorable lighting conditions, and in often cases, are unanswerable by humans. \citet{gurari2019vizwiz} also proposed the VizWiz-private dataset that deals with images containing private content in the same setup as VizWiz. 

Recently, the mobile application Be My Eyes\footnotemark[1] was launched to to provide automated assistance to the visually impaired. With a user base exceeding 500,000 individuals from 150 countries, the app utilizes the multimodal LLM GPT-4 \cite{openai2023gpt4} to generate VQA responses. The VizWiz challenge has also been expanded to other domains e.g image captioning, VizWiz-Captions by \citet{gurari2020captioning}, a few-shot localization setting by \citet{tseng2022vizwiz} and Visual Grounding by \citet{chen2022grounding}.

\footnotetext[1]{https://www.bemyeyes.com/}

\begin{figure*}[ht]
\centering
    \includegraphics[width=0.8\linewidth]{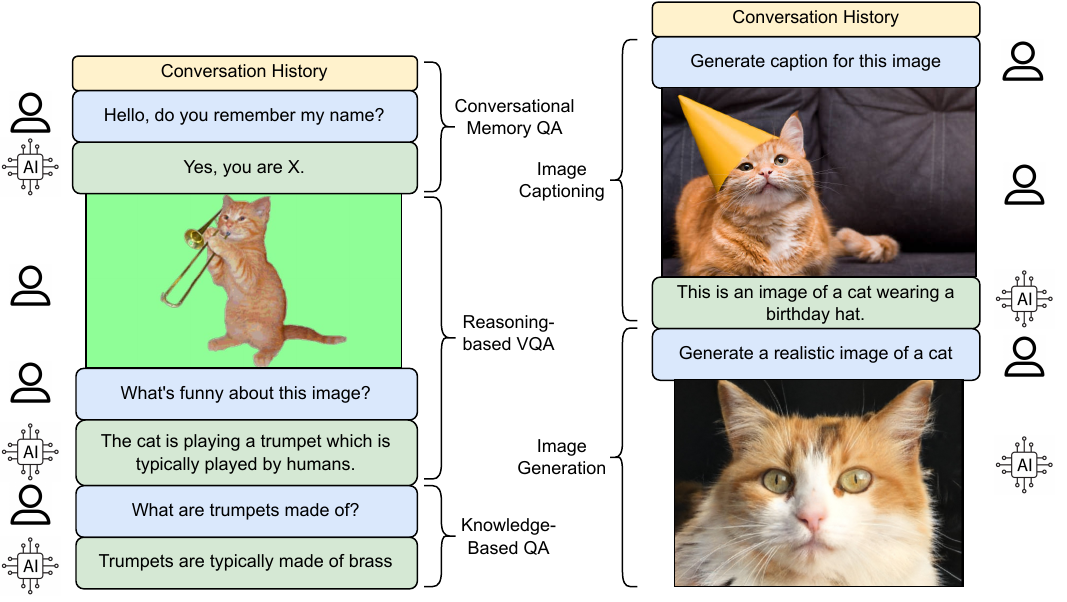}
    \caption{
    A standard visual dialog system should be capable of a multitude of unimodal and bimodal tasks. Some of the tasks include but are not limited to tracking conversational history during question answering, performing any form of textual or visual reasoning, being able to answer questions that require external knowledge, generating image captions, and generating synthetic images. 
    }
    \label{fig:visDial}
\end{figure*}

\subsection{Med-VQA}
\label{sec:medical}
The Med-VQA subdomain focuses on images and questions related to the medical field and has gained immense popularity due to the versatility of its use cases. Existing work shows medical VQA models acting as radiologists, pathologists, or knowledgeable medical assistants \cite{lin2023medical}. Furthermore, these models can potentially decrease the workload of healthcare workers prone to fatigue and burnout. Healthcare burnout has been a concerning issue for decades as studies found a negative correlation between workers experiencing burnout and the quality of healthcare \cite{salyers2017relationshipHealthcareBurnout}. Hence, integrating VQA systems in the medical domain can reduce instances of disease misdiagnosis and spread of misinformation. Consequently, the safety and overall quality of healthcare services is enhanced.

Automated VQA systems can also serve as a reliable consultancy system for patients seeking professional advice, following up on previous checkups, or fact-checking medical information. Online Med-VQA systems have the potential to make healthcare accessible to a broader demographic by eliminating physical barriers and reducing costs.  An integral application of Medical VQA lies in emergency healthcare scenarios, particularly when non-experts need to take swift actions within a short time. For instance, in the event of a snakebite, an individual can capture an image of the wound to verify whether the snake was venomous. The person can further ask about the steps required to treat a venomous snakebite. 

An ideal medical VQA system should be able to generate a comprehensive answer for free-form and open-ended questions on medical images encompassing the whole or majority of the domain. The model must also possess knowledge and reasoning capabilities beyond the scope of its training data as unseen cases are frequently encountered in the medical domain. The systems should also exhibit lower false positive rates in the identification of medical conditions, similar to the low false positive rates of the corresponding classifiers. Additionally, the systems should be reliable both in terms of answer quality and accessibility. The accuracy of the system's responses should closely align with that of professionals in the domain. Current Med-VQA models most satisfy these criteria suggesting that a future where Med-VQA systems are fully integrated into global healthcare services may be closer than we can imagine.

\subsection{Education}
\label{sec:education}

The education sector is responsible for nurturing the researchers of the next generation. The integration of AI in education can potentially improve the accessibility and scope of education. It should be noted that the scope of this field is quite broad starting from teaching pre-scholars to taking courses for the professionals. Question-answering is a fundamental technique used to reinforce and evaluate the learned concepts. VQA systems can be great learning assistants to students, hobbyists, and professionals.

\citet{educationalRobotSystem} proposed an automatic robot system that uses VQA to teach pre-scholars and act as their companion. The robot captures images of objects from the environment and asks questions regarding them using various multimodal modules for QA generation. The work has been expanded to a separate subdomain called Educational Robotics \cite{anwar2019systematicEduRobotics}, often relying on VQA-based techniques with comprehensive answer generation. 

During the pandemic, \citet{eduBot} proposed visual chatbots to assist students in education. Current, state-of-the-art generative AI models \cite{wu2023visual, openai2023gpt4} can provide better assistance as they excel at Visual Dialog \cite{das2017visualDialog}. Another promising technique for educative VQA is the gamification of VQA systems as proposed by \citet{gamificationVQA} where game-based systems can incentivize students to learn new visual concepts while the core of the assessment will rely on automated VQA. Visual question generation \cite{guidingVisualQuestionGeneration} can complement VQA systems to create new content for evaluating learned concepts. 

VQA on infographics \cite{mathew2022infographicvqa}, diagrams \cite{kembhavi2016diagramAI2D}, plots \cite{methani2020plotqa}, and documents \cite{mathew2021docvqa} are also opening scopes of improvement in the education domain, especially for business analysts and similar professions.  \citet{bongini2020visualCultural} introduced VQA to automate museum guides and can be used as a source of knowledge by active learners to know more about cultural heritage and artworks. Several VQA works \cite{kembhavi2017youTextbookVQA, ding2023vqapdf,tanaka2023slidevqa} were proposed for textbooks, PDFs, and slides which are rich sources of knowledge for both students and professionals. In the future, we can expect to see education domain-specific generative models built on top of foundational models \cite{bommasani2021opportunitiesRisksFoundationalModels} to fully automate any form and modality of educative question-answering.

\subsection{Visual Dialogue/ChatBot}
\label{sec:VisualDialogChatbot}

Visual Chatbots are generative AI models that are capable of performing Visual Dialogue \cite{das2017visualDialog} and have shown substantial potential fueled by the popularity of Large Language Models (LLMs). Before exploring the realms of generative AI, a term that will be repeatedly used throughout the literature, let us first define what generative AI or genAI is. GenAI models are capable of creating new content including but not limited to texts, images, and videos. Foundational models \cite{bommasani2021opportunitiesRisksFoundationalModels} can be defined as generalized AI models trained on a large amount of data and capable of excelling in a variety of AI tasks. The foundational models are usually fine-tuned for domain-specific downstream tasks like Visual Question Answering (VQA). 

Detouring from the field of generative AI, we will explore language modeling in natural language processing. Language Models primarily deal with assigning probability values to words being trained on a text corpus. Large Language Models (LLMs) are upscaled versions of language models with model parameters ranging from a few hundred million to billions. The GPT-based models \cite{radford2018improving} are notable among the foundational models being used as LLMs following the success of the revolutionary transformer architecture \cite{vaswani2017attention} in drastically improving the performance of language models and enabling parallelization on sequential data. 

A proprietary GPT-based model, ChatGPT, derived from GPT-3 \cite{brown2020language}, popularized conversational AI which piqued users' interest in visual conversational AI. \citet{wu2023visual} introduced Visual ChatGPT by using ChatGPT with computer vision-based models and developing a modular architecture for vision-language tasks. The promising results from Visual ChatGPT followed by groundbreaking results from the multimodal LLM GPT-4 \cite{openai2023gpt4} ensure that Visual Chatbots will soon be deployed in real-world applications. A key point to note is that these GPT-based models are \emph{generative} while VQA models are traditionally \emph{discrimative} -- both of these properties will be explored in section-\ref{sec:genVQA}.

\subsection{Miscellaneous}

Apart from the aforementioned applications of VQA, there are several miscellaneous use cases worth exploring. The Visual Chatbots described in section-\ref{sec:VisualDialogChatbot} may not be considered an end-product but can be integrated into various systems to provide customer service. The domains of application include but are not limited to product recommendation, troubleshooting, and website tutorials. Furthermore, VQA can be extensively used for qualitative data analysis from visual diagrams, visual accessibility, streamlining the user experience, etc. QA on diagrams enabled us to develop better ways to interact with digital visualizations that dominate today's businesses.

\citet{toor2019biometric} proposed a VQA system called C2VQA-BOARS for the surveillance of individual biometric data on both images and videos. VQA can also be part of information retrieval systems applicable in several domains like medical, geospatial, military, remote sensing, etc. \citet{sarkar2021vqaAidPostDisasterDamageAssessment, sarkar2023samPostDisasterDamageAssessment} utilized VQA architectures to assess post-disaster damage which can useful in prioritizing aid to regions affected by disasters. Furthermore, current VLP techniques used in VQA improve the performance of image-text alignment in general. Paired with big data systems, such VLP systems will be the key to the utilization of large visual corpora and establishing meaningful relationships between visual and textual pairs. Consequently, any application domain dependent on conversing between images and texts will benefit from the state-of-the-art VLP systems. 

\section{Definition and Scope}
\label{sec:defScope}

The task definition of VQA has evolved throughout the years -- starting from a single image-based question answering to QA on any type of visual input. The visual input can take forms including but not limited to images \cite{antol2015vqa}, video \cite{xu2017videoCaptionQA,zhong2022video}, gif \cite{jang2017tgif}, set of images \cite{bansal2020visualImageSet}, diagrams \cite{methani2020plotqa}, slides \cite{tanaka2023slidevqa}, and $360^{\circ}$ images \cite{chou2020visual360deg}. VQA systems are widely used to solve subtasks of other complex multimodal problems. EmbodiedQA \cite{das2018embodied}, discussed in section-\ref{sec:EmbodiedQA}, requires a Reinforcement Learning (RL) agent to use VQA to answer questions in a 3D synthetic environment. Visual Dialog \cite{das2017visualDialog}, explored in section-\ref{sec:VisualDialogChatbot}, can also have VQA subtasks as depicted in fig-\ref{fig:visDial} but needs to incorporate information retrieval, commonsense reasoning, and conversational memory along with the standard setting of VQA. 

The scope of VQA is bounded by any form of multimodal question-answering problem related to generating a textual answer, $Y_t$, to a textual question,$X_q$ on a visual input, $X_v$. We can formally define VQA as $\mathcal{V}$, such that,

\begin{equation}
    \mathcal{V}:X_{v}, X_{t} \rightarrow Y_{t}
    \label{eq:vqa}
\end{equation}

The flexibility of having any form of visual input allows VQA to have various sub-tasks that can be based on the domain, modality, or answer generation procedure. Referring to fig-\ref{fig:vqaTasks}, most of the works in VQA are done on realistic or natural images. The performance of the model on generalized visual input is primarily evaluated by the ability of the model to correctly answer natural images. Most of the natural images are sourced from the COCO dataset by \citet{lin2014microsoft} while MSVD and MSR-VTT \cite{xu2016msr} have been popular sources for natural videos. The sources of VideoQA tend to be more diverse due to the large variation of video content in general.

A popular setting of VQA uses graphs, charts, documents, and texts in natural scenes as the visual input. Document and text interpretation by VQA models have work techniques similar to Optical Character Recognition (OCR) \cite{mori1999opticalOCR} systems i.e. by extracting textual information from the images and using them as a context to answer the question. As seen in section-\ref{sec:medical}, VQA is also gaining popularity in the medical domain, primarily due to the scope of developing potential educational, research, and service-related medical applications.

A separate branch of VQA is dedicated to criticizing existing datasets and methods by highlighting biases and the limited ability of VQA models to reason. Sec-\ref{sec:biasReasoning} will discuss the datasets applicable for reasoning and bias mitigation. Commonsense reasoning is often treated as a separate task known as Visual Reasoning \cite{suhr2017corpus}. Ideally, VQA models must have a reasonably high capability of understanding both visual and textual concepts. \citet{johnson2017clevr} challenged existing models with the CLEVR dataset that incorporated complex questions requiring a good depth of image and textual semantic understanding. On the other hand, models might exhibit linguistic bias due to dataset distribution \cite{shrestha2020negativeLinguisticBias}. Linguistic bias causes VQA models to learn subtle correlations between visual and/or textual entities in the dataset causing the models to generate a correct answer without following the proper path of reasoning. Bias is also exhibited by the annotators due to various factors including but not limited to age, sex, gender, and location \cite{hirota2022genderRacialBiasVQA}. 

VQA also has several miscellaneous domains e.g. counting-based questions by \citet{acharya2019tallyqa}, QA on change detection by \citet{yuan2022change}, QA on 360$^\circ$ images \cite{chou2020visual360deg}, etc. Further variations of VQA based on the modality will be discussed formally in \ref{sec:modalitiesVQA}. The definition of VQA also varies based on the procedure of answer generation and will be discussed extensively in sec-\ref{sec:answerGeneration}. While traditionally, the problem of VQA has been treated as a classification problem \cite{ren2015exploring}, contemporary LLM-based architectures are shifting towards generative answers \cite{guo2023imagesFrozenZeroShot}.

\section{Surveys in VQA}
\label{sec:surveys}

Surveys and reviews are gateways to new research openings and introduce newcomers to a particular domain. A comprehensive survey is beneficial to beginners by navigating them through the intricacies of the domain and providing high-level research direction. Similarly, domain experts and experienced researchers equally benefit from surveys that highlight the challenges, recent trends, and open problems in the field. Surveys can also save time in reviewing literature and help organize works in the domain. 

Over the years, VQA has garnered significant attention in the research community resulting in numerous high-quality surveys \cite{wu2017visualSurvey, kafle2017visual}. The surveys can be roughly categorized into \textbf{generalized surveys} that provide a \emph{bird's eye view} on the whole domain itself and \textbf{specialized or critical surveys} that extensively focus on a part of the domain. Hence, generalized surveys tend to explore the breadth of the domain by expanding on newer topics while specialized surveys explore the depth of certain topics through critical analysis. Although our work tries to provide a broad overview of the field, we also provide an in-depth analysis of some of the challenges in the domain. 

Researchers also benefited by exploring surveys on tasks closely related to VQA \cite{hossain2019comprehensiveImageCaptioning} or more generalized surveys that encompass various multimodal problems along with VQA \cite{baltruvsaitis2018multimodalMachineLearning}. The VLP techniques used by contemporary models are closing the gap between various vision-language problems. Hence, researchers will equally benefit by exploring VLP surveys \cite{gan2022visionVLPSurvey, chen2023vlpSurvey} that can be considered specialized reviews in the domain. In recent years, there has been a steady decline in the "VQA-specific" surveys due to the rise of VLP. To the best of our knowledge, there's no existing review that encompasses the domain of VQA from the early deep-learning era relying on CNNs and LSTMs to the modern VLP techniques. Our work solely aims to bridge the gap between these two classes of survey and provide a single comprehensive review encapsulating the evolution of the domain. 

\subsection{Generalized Surveys}
The generalized surveys emphasize the traditional single image-based setting \cite{antol2015vqa} and explore the associated datasets, models, metrics, challenges, and opportunities. Table-\ref{tab:surveyGen} highlights some of the prominent generalized surveys along with the challenges, open problems, and key contributions. Generalized surveys primarily serve as a guide to entry-level researchers in the domain, are more comprehensive in nature, and introduce basic forms of categorization. 

Recently, the domain of VQA has been revolutionized by both generative AI and new approaches to the zero-shot setting. Such, trends haven't been explored by the existing surveys which might halt the fast-paced research in the domain. Additionally, most of the taxonomy introduced in these surveys has been outdated as the domain is rapidly evolving. The surveys rarely placed VQA in the world of multi-modal problems and compared it with related domains.

\subsection{Specialized Surveys}
Specialized surveys in VQA can be dedicated to a particular sub-domain of VQA \cite{zhong2022video, lin2023medical}, particular phases of VQA \cite{zhang2019information, lu2023multi}, or any topic of interest \cite{barra2021visual}. In addition to surveys strictly on VQA, related surveys on generalized topics like Vision and Language problems \cite{kafle2019challenges} or Multimodal Learning \cite{baltruvsaitis2018multimodalMachineLearning} can provide researchers a broader view of the domain. Such surveys are key to establishing new challenges and open problems in the domain. Researchers will also benefit from surveys that are closely related to VQA, \emph{e.g.} Image Captioning \cite{hossain2019comprehensiveImageCaptioning}. Being a multimodal problem that deals with image understanding and text generation, image captioning methodologies are often embraced in VQA due to the high levels of similarity. Furthermore, surveys on Vision Language Pre-training (VLP) \cite{gan2022visionVLPSurvey, chen2023vlpSurvey} are valuable guides to researchers wishing to explore contemporary architectures.
\input{TabSurveySpec}
\input{TabSurveyGen}
\section{Datasets}
\label{sec:Datasets}

The domain of VQA has a large number of rich datasets, introduced throughout the years. The challenge of establishing a VQA dataset dates back to the pre-deep learning era of VQA. The early VQA methods relied on manual answer generation \cite{bigham2010vizwiz} or used probabilistic methods \cite{malinowski2014multi}. Prior to establishing a standard dataset, the domain faced challenges in formulating a problem statement as most of the early datasets had restricted settings \cite{malinowski2014multi}. As deep learning-based techniques evolved rapidly in both vision and language, the necessity of establishing a benchmark dataset for VQA became evident. Several benchmarks \cite{ren2015exploring, antol2015vqa} were proposed in the early deep learning era of VQA followed by datasets that criticized various aspects of these benchmarks \cite{johnson2017clevr, pandhre2017shapes, agrawal2018don}. The latter class of datasets, often referred to as diagnostic datasets, challenged aspects like visual reasoning, image understanding, and language bias in existing VQA benchmarks.

Concurrently, the early deep-learning era saw prominent works in Knowledge-based datasets \cite{wang2015explicit, wang2017fvqa}. that focused on extracting knowledge from an external source or answering fact-based questions. The problem was often extended to answering questions beyond the training data \cite{teney2016zero} resulting in defining VQA in zero-shot settings. VQA also experienced growth in the number of application-based datasets primarily VizWiz \cite{gurari2018vizwiz} and VQA-Med \cite{hasan2018overview} -- both described in-depth in section-\ref{sec:Applications}. Recent dataset trends are focused on Video Question Answering \cite{xu2017videoCaptionQA,zhong2022video} and Figure-Text-based question answering \cite{methani2020plotqa, mathew2021docvqa}. Video QA continues to rapidly evolve as a sub-domain with notable trends towards multimodal Video QA datasets encompassing other modalities along with vision. The following sub-sections will explore the broader categories of VQA datasets.

\begin{figure*}[ht]
    \label{fig:timelineDS}
    \centering
    \includegraphics[width = 0.85\textwidth]{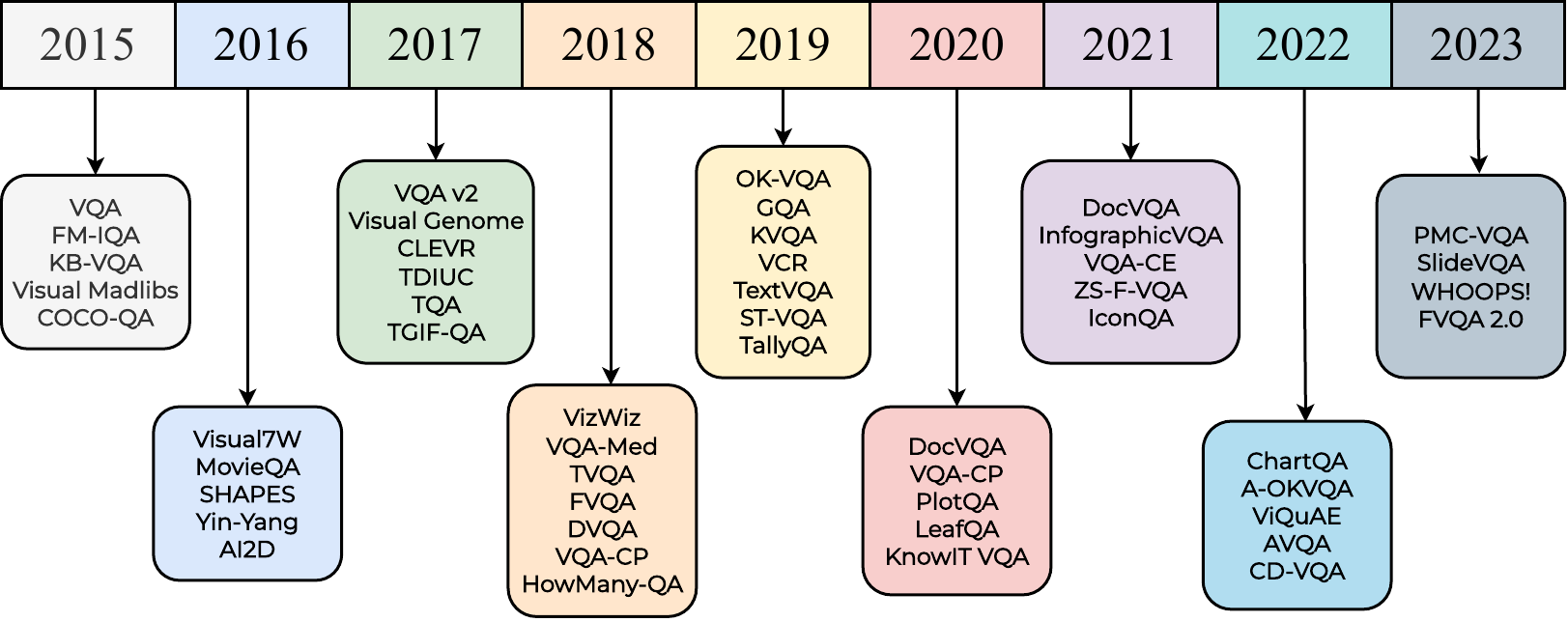}
    \caption{Timeline of popular VQA datasets} 
\end{figure*}

\subsection{Traditional Datasets}

The datasets and task specifications evolve in every domain; often leading to outcomes that differ significantly from the initial setting. VQA is no exception as it started off as question-answering on natural images but is currently, extended to any form of visual input. Although the constraints on the input and the nature of the output have changed, the use of a single image remains a popular VQA setting. 

Traditional datasets or benchmark datasets aim to create a standard for training and evaluating VQA models by providing a large amount of data that closely resembles real-world scenarios i.e. utilizing natural images with human-annotated QA pairs \cite{ren2015exploring, antol2015vqa}. However, the datasets are restricted to a single image only. Traditional datasets are popular performance benchmarks as they provide an overall assessment of the model. Coupled with the fact that the traditional datasets bear a strong resemblance to the deployed environment, they are extremely popular among the VQA community. However, these datasets have been subject to several limitations, notably, language bias \cite{goyal2017making}, limited reasoning on both visual and textual modalities \cite{johnson2017clevr}, and not incorporating questions beyond the knowledge of the training data \cite{teney2016zero}. 

\subsubsection{DAQUAR and COCO-QA}
\citet{malinowski2014multi} introduced the DAQUAR dataset as an early VQA benchmark and afterward, initiated the Visual Turing Challenge \cite{malinowski2014towards}. DAQUAR was a relatively small dataset restricted to indoor images but was extended with additional annotated answers to the DAQUAR-Consensus dataset \cite{malinowski2015ask}. Following approaches for textually annotating large image datasets done in image captioning \cite{chen2015microsoft}, \citet{ren2015exploring} proposed the COCO-QA dataset by algorithmically generating QA pairs for the COCO dataset \cite{lin2014microsoft}. COCO has been the primary source of natural images for all the subsequent traditional datasets and remains the most popular source of images for single-image QA. 

\subsubsection{VQA v1 and v2}
\citet{antol2015vqa} introduced the VQA dataset by using unrestricted or free-form questions on both real and synthetic data. The VQA dataset, later known as VQA v1 or 1.0, became a domain benchmark but was quickly criticized due to linguistic bias \cite{zhang2016yin, goyal2017making}. VQA was termed "unbalanced" meaning that the models trained on the dataset were emphasized on the question while often disregarding the image. The problem was countered by VQAv2 \cite{goyal2017making} by introducing a counterexample that will produce a different answer to the same question and hence, force the model to comprehend or \emph{look} at the image. Both VQA and VQA v2 have been popular benchmarks for model evaluation throughout the years. 

\subsubsection{Visual Genome and Visual7W}
The Visual Genome project has been the source of two rich VQA datasets, Visual7W \cite{zhu2016visual7w} and Visual Genome \cite{krishna2017visual}. With 1.7M QA pairs, the visual genome project is one of the richest datasets in both VQA and the introduced task of Visual Grounding \cite{zhu2016visual7w, deng2018visualGrounding}. For every image in the visual genome dataset, there is a scene graph that is used to automatically generate candidate answers. 

In order to introduce diversity in questions and create VQA challenges that correspond to computer vision challenges, the datasets introduced the "W"s in creating QA annotations. The 7 "W"s -- "what", "who", "when", "where", "why", "how", and "which" are commonly used in political journalism for complete storytelling \cite{kuhn2003political}. The visual7W dataset uses all 7 "W"s while the visual genome dataset uses 6 "W"s, omitting "which". Visual7W was described as a subset of the Visual Genome project with a few additional annotations. 

\subsubsection{Other Datasets}
\citet{yu2015visual} took a different approach to VQA by introducing fill-in-the-blanks with multiple options. \citet{gao2015you} annotated the QA pairs in Chinese and then translated them to English to create a multilingual dataset. The work was pivotal in introducing non-English VQA systems and subsequently inspired many non-English VQA datasets \cite{rafi2022deepBanglaVQA, chandrasekar2022indicHindiVQA, kamel2023vaqa}. \citet{kafle2017analysis} introduced 12 categories of questions with a new evaluation scheme to fight biases due to the abundance of a certain question category. 

Ok-VQA \cite{marino2019okVQA} has been a popular benchmark on the single-image VQA, primarily used to evaluate many modern Zero-Shot VQA models \cite{guo2023imagesFrozenZeroShot, tiong2022plug} and Multimodal LLMs \cite{openai2023gpt4, peng2023kosmos2}. The dataset will be categorized separately due to incorporating external knowledge but the readers are encouraged to use this dataset as a benchmark for standard, few-shot, and zero-shot settings of VQA models. 

\subsection{Knowledge-based (KB) Datasets}

The VQA datasets were usually limited by the training data i.e. models generalizing beyond the training data wasn't expected. However, in realistic scenarios expect the models to possess general knowledge and common sense. KB-VQA \cite{wang2015explicit} is defined as the setting of VQA that deals with answering knowledge-based questions by extracting the data from a secondary source i.e. it employs a form of information retrieval to the standard setting of VQA. The source is often referred to as a secondary source of knowledge as we are considering training data as the primary knowledge source for a VQA model. Additionally, there can be multiple ways to represent a secondary source e.g. knowledge bases, knowledge graphs, etc.

A knowledge base (KB) can be defined as a collection of knowledge triplets, $(e_1, r, e2_2)$, such that, $e_1$ and $e_2$ represent two entities and $r$ represent the relationship between them. The set of triplets can also be formulated as a graph called a Knowledge Graph (KG). However, unless specified, we use the term knowledge base to represent any external knowledge source including knowledge graphs. Knowledge bases typically have large-scale structured data stored in relational database management systems (RDBMS) accessible using any form of query language. KB-VQA systems can have single, multiple, or no specified KBs. 

\subsubsection{KB-specified Datasets}

\citet{wang2015explicit} proposed a \emph{test} dataset to evaluate the performance of VQA models on external data and thereby formulated the first knowledge-based setting of VQA systems. The proposed Ahab model relied on DBpedia \cite{auer2007dbpedia} as the source of external knowledge. The KB-VQA dataset was relatively small and focused on pre-defined template-based questions. The FVQA dataset proposed by \citet{wang2017fvqa} extended the previous setting by ensuring that each question will have a supporting fact that can be extracted from the associated KB. Furthermore, utilizing multiple knowledge bases \cite{tandon2014acquiringWebchild, liu2004conceptnet} along with DBpedia resulted in the reduction of dependency on a single KB. 

\citet{lu2018rvqa} proposed the R-VQA dataset based on the extraction of relational facts using samples from Visual Genome \cite{krishna2017visual} and the associated knowledge graph (KG). R-VQA aimed to address the semantic gap between languages and images during information extraction from images or knowledge sources. \citet{lin2023fvqa2} further addressed the imbalance of the FVQA by introducing adversarial samples, ensuring a higher degree of robustness. However, all the KB-specified challenges are dependent on their associated KBs resulting in the inability to generalize beyond the specified KBs.

\subsubsection{Open Knowledge VQA (OK-VQA) Datasets}

\citet{marino2019okVQA} proposed a generalized dataset like VQA \cite{antol2015vqa} but for KB systems. The dataset inherits all the attributes of traditional datasets along with questions related to external knowledge. The diverse set of questions with the challenges of incorporating external knowledge, resulted in a large yet difficult dataset. As the dataset wasn't restricted to a particular KB, it shaped a new challenge of VQA with \emph{open knowledge} or \emph{open domain}. 

\citet{jain2021select} argued that models trained on OK-VQA were able to achieve high scores while predicting based on the answer distributions. They introduced their own version of OK-VQA as OK-VQA$_{S3}$ with another new dataset, S3VQA. Following the steps of OK-VQA, \citet{schwenk2022okvqa} introduced A-OKVQA as a difficult benchmark by incorporating reasoning-based questions and adding \emph{rationale} to the QA annotations.

\input{TabDsTradQuant}
\input{TabDsTradCont}

Open knowledge-based datasets have been popular evaluation benchmarks of the recent VQA models \cite{li2022blip, chen2022pali}, few-shot models \cite{song2022clipFewShot, alayrac2022flamingo}, zero-shot models \cite{guo2023imagesFrozenZeroShot, tiong2022plug}, and multi-modal LLMs \cite{huang2023languageKosmos1, openai2023gpt4}. The zero-shot setting closely aligns with the task of open-knowledge VQA as both models rely on predicting beyond what they have seen during training. The performance difference between standard evaluation and open knowledge evaluation is significantly lower for zero-shot models \cite{guo2023imagesFrozenZeroShot} than standard VQA models. We believe that future VQA datasets will draw inspiration from OK-VQA to establish the groundwork for generalized VQA evaluation.

\subsubsection{KB-VQA on Named Entities (KB-VQA-NE)}
Challenges in VQA attempt to replicate realistic scenarios in order to develop models mimicking human capabilities or going beyond that. Recognizing a person, object, or landmark is a commonly encountered real-life task. In this context, recognition refers to recalling the name of the entity. For instance, a picture of the Eiffel Tower is shown and asked, "What is the name of this tower?" with additional questions, "Where is this tower located?". The equivalent KB-VQA problem settings have been termed as Knowledge-aware VQA (KVQA) \cite{shah2019kvqa} and Knowledge-based VQA on named Entities (KVQAE) \cite{lerner2022viquae}. Instead of using multiple pre-established terms, we will refer to the class of datasets as KB-VQA-NE by simply using the extension Named Entity (NE) to KB-VQA.

\citet{shah2019kvqa} introduced a large-scale dataset with 183k questions related to people from Wikidata \cite{vrandevcic2014wikidata}. A knowledge graph containing the relationship between entities was used to retrieve information on those particular entities. \citet{lerner2022viquae} extended the setting to landmarks and named objects along with highlighting few-shot and zero-shot methods in the sub-domain. KB-VQA-NE datasets can also be considered a reformulation of the face-recognition and object-recognition problems in computer vision for VQA systems. However, questions provide the flexibility of performing different tasks and can make the KB-VQA-NE systems an excellent choice for realistic use cases.

\subsection{Reasoning and Bias Reduction (RBR) Datasets}
\label{sec:biasReasoning}

With the advent of large standardized VQA datasets \cite{antol2015vqa}, people attempted to criticize the benchmarks by highlighting the underwhelming performance of the associated models on complex reasoning-based questions \cite{johnson2017clevr}. As the training data didn't include complex questions, the trained models were unable to comprehend the new challenge. Several datasets \cite{johnson2017clevr,hudson2019gqa,lu2021iconqa} were proposed to enhance the reasoning capabilities of the models and incorporate challenges related to visual and textual reasoning. Some of the corresponding datasets are often termed diagnostic datasets \cite{johnson2017clevr} but will simply be called reasoning datasets in this literature. 

VQA datasets aim to train models with a comprehensive understanding of both visual and textual modalities but without exploiting any other correlation in the training data. As previously discussed in section-\ref{sec:Intro}, datasets can exhibit different forms of association within their data entries and these associations can be easily captured by VQA models, potentially leading to a form of overfitting. Usually, the models learn to correlate questions to answers based on the answer distribution of certain question categories and hence, exhibit linguistic bias - a form of multimodal shortcut.  

The linguistic bias can also be viewed as the model's inability to understand or extract features from the image. The associated model can be analogous to a \emph{blind model} i.e. the visual input doesn't affect the generated answer. The model's inability could be attributed to either the model's architectural incompetence or the dataset's inability to train the model's visual reasoning. Several works \cite{zhang2016yin} were presented to highlight the linguistic bias and imbalance of the benchmarks \cite{antol2015vqa, ren2015exploring}, one of them being the popular iteration VQA v2 \cite{goyal2017making}.

The seemingly unrelated concepts of bias and inability to reason are, in fact, two sides of the same coin. A biased model will take various multimodal shortcuts to conclude a particular answer that will subsequently cause the model's inaptitude for reasoning. Similarly, a model incapable of reasoning will rely on multimodal shortcuts to generate the desired answer. Dataset redistribution might mitigate the linguistic bias resulting in better visual and textual comprehension \cite{dancette2021beyondVQACE, agrawal2018don}. Recent surveys \cite{ma2023robustVQASurvey} also group datasets on bias and multimodal shortcuts in VQA with commonsense and reasoning datasets \cite{hudson2019gqa, gao2022cric}. 

\subsubsection{Reasoning-based Datasets}

\citet{andreas2016neural} introduced a synthetic dataset comprising various arrangements of colored shapes with compositional questions to evaluate visual reasoning. \citet{johnson2017clevr} introduced the popular CLEVR dataset, termed \emph{diagnostic dataset}. The dataset was instrumental in highlighting several drawbacks of VQA models trained on the traditional datasets \cite{antol2015vqa, ren2015exploring}. The CLEVR dataset has been extended to various other tasks including referring expressions \cite{liu2019clevrRef}, visual dialog \cite{kottur2019clevrDialog}, explainable AI \cite{arras2022clevrXAI}, natural language explanations \cite{salewski2020clevrX}, domain robustness \cite{li2023superCLEVR}, and more.

In the following years, several large-scale datasets \cite{hudson2019gqa, zellers2019recognitionVCR} aimed to replicate the generalization capabilities of established benchmarks \cite{antol2015vqa, krishna2017visual}, while incorporating reasoning capabilities. \citet{zellers2019recognitionVCR} introduced a novel task termed Visual Common Sense reasoning that emphasizes the rationale along with the answer. The GQA dataset \cite{hudson2019gqa} addressed both reasoning and linguistic bias along with being a large-scale generalized benchmark. The dataset has been a popular choice of evaluation for state-of-the-art VQA models and zero-shot methods. \citet{zhang2019raven} used questions from general intelligence tests. \citet{bitton2023breakingWHOOPS} used weird, unconventional images to test a model's reasoning capabilities.

\subsubsection{Bias Reduction Datasets}
\citet{zhang2016yin} simplified the linguistic bias and imbalance in VQA datasets by introducing a binary classification problem on the VQA abstract dataset \cite{antol2015vqa}. The popular VQA v2 \cite{goyal2017making} can also be considered a bias-reduction dataset as it reduces the imbalance in the VQAv1 dataset \cite{antol2015vqa} using counterexamples. 

\input{TabDsKB}
\input{TabDsRB}

To grasp the concept of counterexamples, let's consider a dataset where there is an image of a man wearing glasses with the associated question, "Who is wearing the glasses?" and answer, "The Man". Hence, there is an imbalanced distribution of men wearing glasses resulting in the model associating glasses with men. A counterexample of women wearing glasses with the associated QA pair can help mitigate the dataset imbalance.  

Redistributing the VQA datasets has been a popular choice to address biases. Redistribution means splitting the training, validation, and test sets differently. \citet{agrawal2018don} ensured different answer distributions in training and val-test splits. \cite{dancette2021beyondVQACE} redistributed by identifying training counterexamples from val-test splits. \citet{chen2021zero} proposed a redistribution of the F-VQA dataset \cite{wang2017fvqa} in order to evaluate models in zero-shot settings.

\input{TabDsFT}

\subsubsection{Counting Datasets}
Counting-based challenges in VQA are closely related to the reasoning challenges. In fact, many reasoning datasets \cite{johnson2017clevr} employ questions that require counting different objects to evaluate visual reasoning. \citet{trott2017interpretableHowManyQA} proposed a counting dataset by filtering "how many" questions from traditional datasets \cite{goyal2017making, krishna2017visual} and highlighted the limitations of VQA models in counting. \citet{acharya2019tallyqa} extended the previous work by incorporating complex and manually annotated counting QAs with associated images. VQA models are still struggling with counting-based questions, opening opportunities for improvement.

\subsection{Vis-Text Datasets}
The task of VQA primarily used a single image as the visual input. With the increase in architectural complexity, researchers attempted to test the models on other vision-related tasks. The scope of VQA was broadened to any form of image including diagrams \cite{kembhavi2016diagramAI2D}, charts \cite{masry2022chartqa}, infographics \cite{mathew2022infographicvqa}, slides \cite{tanaka2023slidevqa}, etc. Throughout the literature, the term \emph{visualization} is used as an umbrella term to represent any kind of visual representation including but not limited to figures, charts, graphs, plots, diagrams, infographics, digital graphics, and slides. Similar works highlighting the inability of models to interpret texts in a natural image \cite{biten2019scene, singh2019towardsTextVQA} also grabbed the attention of researchers. 

Both the tasks of understanding visualizations and interpreting texts in natural scenes require some form of text-reading skill. While comprehending figures requires a higher level of structural understanding, works on document interpretation \cite{mathew2021docvqa} require a higher level of text recognition. Document processing VQA systems are likely to benefit from external modules like Optical Character Recognition (OCR). However, as the similarities between figure VQA and text VQA outweigh their dissimilarities, they have been grouped together.

\subsubsection{Visualization Datasets}
\citet{siegel2016figureseer} worked on parsing figures from research papers and established the groundwork for Vis-Text VQA. \citet{kembhavi2016diagramAI2D} proposed the task of interpreting scientific diagrams for QA using parse graphs. Digital chart interpretation was emphasized in the subsequent years as \citet{kahou2017figureqa} worked with 5 diagram classes - horizontal and vertical bar, line, dotted line, and pie charts. \citet{methani2020plotqa} and \citet{masry2022chartqa} shifted the focus to real-world charts that were either extracted or generated from actual data. \citet{chaudhry2020leaf} tried to evaluate the reasoning capabilities of figures by using GRE-based questions; drastically increasing the task complexity. 

Several datasets took a specialized approach to specific types of visualization. For instance, \citet{kafle2018dvqa} studied bar charts only for structural understanding and reasoning on a particular form of diagram. Similarly, \citet{mathew2022infographicvqa} worked on infographics which are significantly harder to analyze due to the content variety and the dominance of text in the image. Comprehending infographics requires high-level understanding and reasoning over both visual and textual content in an image. \citet{kembhavi2017youTextbookVQA} explored a unique QA task on middle-school science lessons. The evaluation MCQs at the end of a lesson were used as the questions. The dataset enabled us to assess the current state of machine intelligence in academics.

\subsubsection{Text VQA}
Although most visualizations contain a good amount of textual content, the models trained for QA on visualizations emphasize the structural understanding of the image and the relationship with the visualization entities. A separate set of VQA datasets focuses on reading and understanding text either from real scenes or digital images of texts. \citet{singh2019towardsTextVQA} and \citet{biten2019scene} proposed datasets with natural images of texts in realistic scenes extracted from large natural image datasets.

\citet{mathew2021docvqa} considered images of documents for QA in order to extensively test the reading capabilities of VQA models. A similar task was previously proposed by \citet{mishra2019ocr} on images of book covers. Their work relied on using an Optical Character Recognition (OCR) module and was later extended by \citet{zeng2021beyondOCRplus} on a more robust setting.  \citet{tanaka2023slidevqa} proposed the unique task of performing QA on slide decks. Both documents and slide decks require a high-level understanding of non-textual content like diagrams, charts, and table structures along with the textual content; removing the boundaries between visualization and text-based datasets.

\subsection{Miscellaneous Datasets}

As seen in fig-\ref{fig:vqaTasks}, VQA has been diversified, venturing into different domains and modalities. The trending sub-domain VideoQA \cite{xu2017videoCaptionQA,zhong2022video} extends the classical visual input from spatial to spatiotemporal. Furthermore, VideoQA expands the scope of VQA by incorporating other modalities, e.g., audio and knowledge base. Datasets on assisting the visually-impaired and medical images were explored in section-\ref{sec:assistVisually} and \ref{sec:medical} respectively. \citet{yuan2022change} changed the task setting of VQA by asking questions change-related questions on a pair of images and thereby merging VQA with Change Detection. \citet{bansal2020visualImageSet} extended the VQA setting to a set of images, unlike the traditional setting restricted to a single image.

\subsubsection{Video Question Answering (VideoQA)}
\label{sec:videoQA}
VideoQA emerged as a popular subfield of VQA with more datasets being proposed to expand the modalities. \citet{tapaswi2016movieqa} proposed QA on movies and \citet{lei2018tvqa} proposed the same task on TV series - both works process videos with audio feed. However, a lack of audio-related questions led to \citet{yang2022avqa} proposing a dataset to specifically address audio-visual questions. \citet{xu2017videoCaptionQA} introduced two popular VideoQA datasets generated from large-scale video description datasets. \citet{garcia2020knowit} introduced a knowledge-based setting for VideoQA similar to KB-VQA. 

Other interesting settings of VideoQA are gameplay videos by \citet{mun2017marioqa} and GIFs by \citet{jang2017tgif}. \citet{das2018embodied} formulated the task of embodiedQA, discussed elaborately in sec-\ref{sec:EmbodiedQA}, as an extension of VideoQA and an intersection of VQA with Reinforcement Learning (RL) by proposing the EQA dataset. VideoQA is a promising field to bring forward new datasets and challenges. An in-depth literature review on VideoQA is beyond the scope of this paper and readers are encouraged to study the VideoQA survey by \citet{zhong2022video}.

\begin{table*}[ht]
\centering
\caption{Miscellaneous Datasets in VQA}
\vspace{0.2cm}
\begin{threeparttable}
  
\begin{tabular}{|l|l|l|l|}
\hline
\textbf{Task Category}                                                   & \textbf{Name}                                               & \textbf{Year} & \textbf{Description}                                                                                                         \\ \hline
Video                                                                    & \begin{tabular}[c]{@{}l@{}}MSRVTT-QA, \\ MSVD-QA \cite{xu2017videoCaptionQA}\end{tabular} & 2017          & Automatic open-ended QA from video description datasets                                                                      \\ \hline
\multirow{2}{*}{Multimodal Video\tnote{a}}                                       & MovieQA \cite{tapaswi2016movieqa}                                           & 2016          & MCQ-based questions on movies                                                                                                \\ \cline{2-4} 
                                                                         & TVQA \cite{lei2018tvqa}                                                      & 2018          & MCQ-based questions on TV series                                                                                             \\ \hline
KB Video                                                                 & KnowIT VQA  \cite{garcia2020knowit}                                                & 2020          & MCQ-based on a single TV series                                                                                              \\ \hline
Gameplay Video                                                           & MarioQA \cite{mun2017marioqa}                                                     & 2017          & Open-ended QA on gameplay videos of a 2D Mario game                                                                          \\ \hline

Frame and Video                                                          & TGIF-QA \cite{jang2017tgif}                                                     & 2017                  & Hybrid annotation on Tumblr GIF (TGIF) dataset \cite{li2016tgif}                                                                              \\ \hline

\begin{tabular}[c]{@{}l@{}}Visually Impaired \\ Assisstance\end{tabular} & VizWiz \cite{gurari2018vizwiz}                                             & 2018          & \begin{tabular}[c]{@{}l@{}}Conversational questions on realistic images captured by\\ visually impaired people\end{tabular} \\ \hline
\multirow{2}{*}{Medical VQA}                                             & VQA-Med \cite{hasan2018overview}                                           & 2018          & Open-ended QA on radiology images                                                                                            \\ \cline{2-4} 
                                                                         & PMC-VQA \cite{zhang2023pmc}                                                   & 2023          & MCQ-based questions on a mixture of modalities\tnote{b}                                                                              \\ \hline
Image Set                                                               & IS-VQA \cite{bansal2020visualImageSet}                                                   & 2021          & Open-ended QA on indoor and outdoor scenes                                                                                              \\ \hline

Change Detection                                                         & CD-VQA \cite{yuan2022change}                                                    & 2022          & Automatically generated QA on the SECOND dataset \cite{yang2021asymmetric}                                                                             \\ \hline

\end{tabular}

\begin{tablenotes}
\vspace{0.5cm}
\item[a] Standard VideoQA doesn't have an audio feed while multimodal videos include the audio modality.
\item[b] "Modality" refers to the types of medical images e.g. radiology, pathology, etc.
\end{tablenotes}

\end{threeparttable}
\end{table*}

\section{Methods}
\label{sec:Method}

Over the years, VQA architectures evolved from non-deep learning paradigms relying on probabilistic techniques \cite{malinowski2014multi} to large-scale monolithic architectures \cite{li2022blip}. This section explores the methodologies in the traditional single-image VQA by highlighting the fundamentals, advanced techniques, and evolution throughout the history of the domain. 

\subsection{Fundamental Techniques}
    
    VQA has been approached in a multitude of ways but the standard approach can be broken down into three separate phases: \textbf{Feature extraction}, \textbf{Feature conjugation}, and \textbf{Answer generation}.
    
    Feature extraction aims to extract meaningful information from the multi-modal visual and textual inputs. While most architectures rely on two separate encoders for the two modalities, a few contemporary architectures employ a single multimodal encoder for feature extraction. If a monolithic architecture has not been used, the next phase will be feature conjugation which deals with the aggregation or combination of the unimodal features extracted in the previous phase. The fused output is forwarded for answer generation that can either be a classifier [eq-\ref{eq:classify}] or a natural language generator [eq-\ref{eq:generate}] depending on the problem formulation. 

    For the VQA model, $\mathcal{M}:(V,Q) \rightarrow A$, we define the visual and textual encoders are $enc_v$ and $enc_q$ respectively. The encodings are then combined using multimodal fusion, $\Phi$, discussed in section-\ref{sec:fusion}. For the visual features, $v_f$, and textual features, $q_f$, the visual, textual, and fused encodings can be represented as,
    \begin{equation}
    \label{eq:visualFeatureExtractionEquation}
        h_v = enc_v(v_f)
    \end{equation}
    \begin{equation}
    \label{eq:textualFeatureExtractionEquation}
        h_q = enc_q(q_f)
    \end{equation}   
    \begin{equation}
    \label{eq:fusionBaseEquation}
        h = \Phi(h_v, h_q)
    \end{equation}
    Finally, the fused output is passed to the answer generator to produce the answer, $a$, based on the formulation in section-\ref{sec:answerGeneration}.
    \begin{equation}
        a = gen_a(h)
    \end{equation}

    It should be noted that this is a basic formulation of the traditional VQA architectures relying on the fundamental techniques only while subsequent models have incorporated advanced techniques like utilization of attention and memory. The technique is often termed as \textbf{Joint Embedding} as the generated answer is based on the joint visual and textual embedding. The following subsections will elaborate on feature extraction, feature conjugation, and answer generation.

    \subsubsection{Visual Feature Extraction}
        The visual input is high-dimensional data that is costly to process. Visual feature extraction captures the important information from the high-dimensional visual input and produces a dimensionality-reduced vector representation of the image. The vector representation should hold an abridged form of information mapped to a space where similar features in different images produce similar vectors. Thus, feature extraction ensures that the semantic components of the visual input are conveyed to the model. Furthermore, transforming the visual input into a vector ensures mathematical manipulation and simplifies the process of grouping similar inputs. 
        
        In the era predating the widespread adoption of deep learning models, methods like explicit RGB vector, SVM \cite{cortes1995supportSVM}, HAAR \cite{haar1909theorie, viola2001rapidObjDetHaar}, HOG \cite{dalal2005histogramsHOG}, SIFT \cite{lowe1999object}, Singular value \cite{hong1991algebraic}, PCA \cite{hyvarinen1998image} or simply hand-crafting kernels to extract characteristics \cite{fukushima1982neocognitron, ciregan2012multi} was functional to a certain degree on specific tasks. While these methods were easily computable, they lacked strong generalization capabilities. With the onset of deep learning, Neural networks \cite{pomerleau1988alvinn, sarlashkar1998feature, lerner1999comparative} have become very popular among researchers to retrieve image features. Convolutional Neural Networks (CNN) \cite{krizhevsky2012imagenetAlexNet,lecun1998gradient} significantly outperformed regular Feed Forward Neural Networks for image feature extraction. Over the years, the CNNs boasted a higher number of layers with a larger parameter count. Models like LeNet \cite{lecun1998gradient}, AlexNet \cite{krizhevsky2012imagenetAlexNet}, VGG \cite{simonyan2014very}, ResNet \cite{he2016deep}, InceptionNet \cite{szegedy2015going} are just a few examples --- all fueled by the ImageNet \cite{deng2009imagenet} competition.

\begin{figure*}[htpb]
    \centering
    \includegraphics[width=0.95\linewidth]{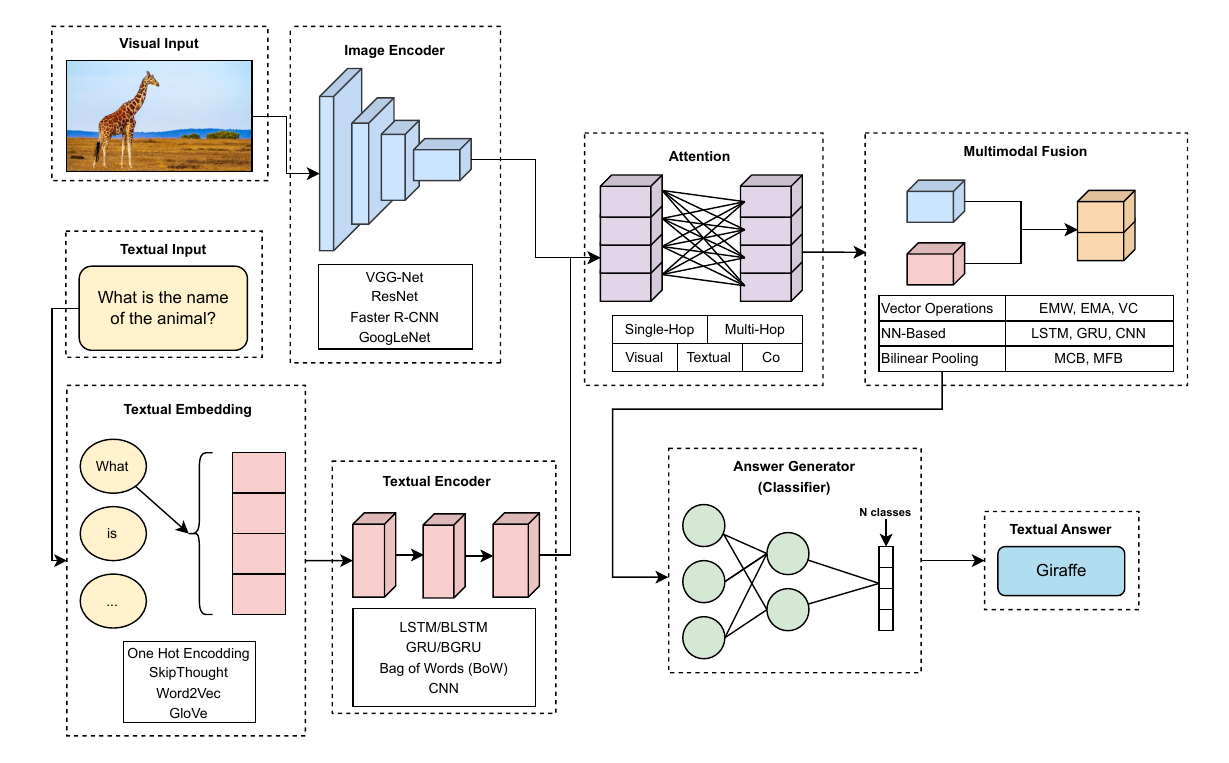}
    \caption{Overview of the traditional pre-transformer VQA architecture based on Joint Embedding and Attention. CNN-RNN-based encoder pairs, Multimodal Fusion, and classification head were used.}
    \label{fig:dualEncoder}
\end{figure*}

        Several VQA models \cite{ren2015exploring, antol2015vqa} use CNN models for visual feature extraction and employ transfer learning to some degree. Transfer learning is an early concept in machine learning research \cite{bozinovski1976influenceTransfer} denoting how an ML model trained in one task can transfer its \emph{knowledge} stored in the gradients to another model being trained on a different task. This approach is noticeably more efficient than training another model from scratch. The model learns structural information and low-level features in the lower layers and can be transferred to other related tasks. 

  Transfer learning is particularly useful for vision models as training such models from scratch requires a massive amount of training data, time, and computational resources as documented by \citet{szegedy2015going}. VQA models usually incorporate GoogLeNet \cite{szegedy2015going}, VGGNet \cite{simonyan2014very}, ResNet \cite{he2016deep}, and varients of R-CNN \cite{girshick2014rich, girshick2015fast, ren2015faster} as visual feature extractors. The CNN backbones are typically pre-trained on Imagenet \cite{deng2009imagenet} or COCO \cite{lin2014microsoft} datasets. 
        
        The usual approach is freezing the whole pre-trained model except the last few layers and fine-tuning those layers on the target dataset \cite{kafle2016answer}. The procedure significantly reduces the training time while requiring less computational resources and training data. The gradual shift from simpler architectures like GoogLeNet and VGGNet to latter architectures like ResNet and R-CNN provides key insights into the requirements of image feature extraction. Both deeper architectures and object localization can drastically improve the quality of extracted visual information.
        
        A recent development in computer vision is the emergence of visual transformers by \citet{dosovitskiy2020image} which breaks an image into $16\times16$ image patches and feeds to a Transformer architecture \cite{vaswani2017attention}. \citet{liu2021swin} later proposed the Swin Transformer which uses a shifting window to apply self-attention, mimicking the convolution operation of CNNs. The architecture offers relatively better performance in tasks that require detailed feature extraction e.g., semantic segmentation and object detection. 
        
        However, convolutions are still relevant as \citet{liu2022convnet} proposed ConvNext architecture that can perform on par or better than transformer-based models in tasks including classification, object detection, and semantic segmentation. The architecture borrows certain ideas from modern transformer models and applies them to classic CNN-based architecture. The aforementioned backbones are widely used by modern VQA models for visual feature extraction as these methods are proven to be quite effective \cite{hirota2021picture, xue2021probing, luo2022towards}. Although contemporary visual feature extraction techniques may adhere to better results, they require a significant amount of time and computational resources. 
 
    \subsubsection{Textual Feature Extraction}
        Textual data from the question should be processed to create vector representations of the words or sentences. Researchers have been seeking ways to improve such vector representations called \textbf{Word Embedding}. Unlike images containing spatial data that can be easily represented in the vector space, texts have semantic data which makes it challenging for a vector space representation. 
            
        One of the earlier approaches is to represent each word as a one-hot vector representation where each $n$ word in the corpus can be represented using an $n$ dimensional vector. However, this simple representation assumes that every word is independent to each other and disregards the similarity between word vectors. Hence, every word vector will be orthogonal in the vector space. Additionally, as the vector size depends on the vocabulary length, the representation can become quite long for larger vocabularies. Count-based methods were also popular in the early days of NLP. These methods constructed a matrix based on the frequency of word occurrences \cite{miller1991contextual} and can be further broken down by computing the singular value decomposition (SVD) \cite{eckart1936approximation} to get the best-fit approximation of lower dimensions. However, these earlier representations are limited by their inability to capture semantic similarity between words.
        
        If the semantic information is properly captured, words with similar meanings should be close in the vector space. Prediction-based models were employed to predict the \textit{next} word given a \textit{current} context. The associated probability can be represented as $P(x^t|x^{t-1},...,x^1)$  where $x^t$ represents the $t^{th}$ word. However, it is rare to use all the previous words as the context, and models like n-gram use the last $n$ words as the context. Consequently, the model learns the word representations by expressing each word in a $k$ dimensional space and capturing the semantic information. 
        
        Earlier works using Neural networks \cite{xu2000can, bengio2000neural} paved the way for sequential models such as Recurrent Neural Networks (RNNs) \cite{rumelhart1985learningRNN} for linguistic representations \cite{mikolov2013linguistic}. Methods like Word2Vec \cite{mikolov2013efficient}, CBOW, and Skip-Gram \cite{mikolov2013linguistic} were developed to learn word representations. These models were trained to predict a masked word given the context surrounding it. Through predictions, the models learn vector representations of words that are later used to extract textual features. Further research on the RNN architecture provided us with GRU \cite{chung2014empirical} and LSTMs \cite{hochreiter1997longLSTM} --- widely popular in VQA for question feature extraction as they are able to preserve long-range contextual information. The recent transformer architecture \cite{vaswani2017attention} outperforms regular LSTMs and GRUs and has been widely adopted in VQA \cite{yang2021comparative, biten2021latr, yang2021tap}. 
        
    \subsubsection{Feature Conjugation or Multimodal Fusion}
    \label{sec:fusion}
        After extracting the visual and textual features from the separate streams, the model has to combine them to form a joint feature vector. Before, the feature conjugation phase, the visual and textual inputs were processed more or less independently. Feature conjugation is also called fusion and is a form of multimodal fusion that deals with combining vectors from two separate modalities into a single modality. For VQA, the multimodal fusion usually aggregates the visual and textual inputs but might also fuse auditory data for sub-domains like Multimodal VideoQA.
        
        Generally, the input streams are independent to each other before fusion. But, techniques like attention might result in the streams affecting each other. For instance, \citet{chen2015abc} used question-guided attention i.e. based on the textual input a form of attention will be produced on the visual input. However, attention isn't a form of fusion. In fact, the easiest way to identify the fusion phase is to recognize the step from where the visual and textual features lose their own identity and cannot be differentiated from one another. Before feature conjugation, both the visual and textual features can be used in another network for another task unrelated to VQA. After feature conjugation, the modalities are merged and the resulting output encapsulates multimodal information of the two input streams.

        Vector operations have been popular choices of feature conjugation. The operations include Vector Concatenation (VC) \cite{zhou2015simple, jabri2016revisiting, huang2019novel}, Element-Wise Multiplication (EWM) \cite{antol2015vqa, zhu2016visual7w, teney2018tips}, and Element-Wise Addition (EWA) \cite{chen2015abc, wang2017fvqa, yang2016stacked}. The methodologies will produce a simple yet effective joint embedding that can be further passed through the latter layers of the model. Apart from vector operations, \textbf{neural network-based} approaches like using a CNN or RNN-based network can also be used for fusion. \textbf{Bilinear pooling} has been a popular choice for multimodal fusion. Prominent works include Multimodal Compact Bi-Linear Pooling (MCB)\cite{fukui2016multimodalVQAVisualGrounding} for VQA \cite{ben2017mutan} addresses the Multimodal Tucker Fusion model.

    \subsubsection{Answer Generation}
    \label{sec:answerGeneration}
    
    VQA can either be treated as a \emph{classification problem} if the model has to pick an answer from a fixed set of answers or a \emph{generative problem} if the model has to generate a natural language answer. When treated as a classifier or \emph{discriminative} model, a single-word answer is usually generated. Evaluation of such answers tends to be easier using simple evaluation metrics like VQA accuracy \cite{antol2015vqa}. On the other hand, generative models often produce longer answers that are tough to evaluate. 

    For a visual input $v$, a question input $q$, and generated answer $\hat{a}$, we consider a VQA model, $\mathcal{M}: (V,Q) \rightarrow A$ with the parameters $\theta$. Depending on the problem formulation, the model can either be treated as a classifier or a generator.

    To generate the answer $\hat{a}$ from a set of answers $\mathbb{A}$, the formulation of VQA as a classification task can be defined as:
    \begin{equation}
    \label{eq:classify}
        \hat{a} = \underset{a \in \mathbb{A} }{\arg \max \ } p(a|v,q; \theta)
    \end{equation}
    
    For the task of answer generation, given a set of answer tokens  $\hat{a} = \{\hat{a}_1,...,\hat{a}_n\}$, the generative formulation of VQA is defined as:
    \begin{equation}
    \label{eq:generate}
        \hat{a}_n = \underset{a \in \mathbb{A} }{\arg \max \ } p(a|v,q,\hat{a}_{<n}; \theta)
    \end{equation}

\input{TabModTrad}

\subsection{Advanced Techniques}
\label{sec:advancedTechniques}
    Although the fundamental techniques form the base of VQA, contemporary methodologies rely on several advanced techniques that can provide several advantages including performance improvement, model scalability, robustness, etc. The realm of advanced VQA techniques can be overwhelming. In this subsection, we discuss a few of these advanced methods.
    
    \subsubsection{Attention}
    \label{sec:attention}
    When an image or a text is given to a human, that person doesn't equally focus on the whole image or text. Depending on the content, the person might focus more on a particular region of the image or a particular portion of the text. A similar mechanism is employed in VQA following the success of attention in standard computer vision tasks like object recognition \cite{ba2014multipleObjRecogAttention} and multimodal tasks like image captioning \cite{jin2015aligningImgCapAttention, hossain2019comprehensiveImageCaptioning}. Attention can be described as \emph{soft weights}, where the model learns to focus more on certain parts of the input. 

    While there are multiple ways of implementing attention, the mechanism fundamentally relies on generating an attention vector by correlating the modalities followed by some form of normalization. The entire process is typically referred to as a single attention layer. Depending on the architecture, models can incorporate single or multiple attention layers. Certain types of multi-layer attention may be dependent on the attention output of the previous layers. Attention mechanisms can be classified based on the number of attention layers as \textbf{single-hop} and \textbf{multi-hop} attention. Each category can be further classified based on attention modality as \textbf{visual attention}, \textbf{textual attention}, and \textbf{co-attention}. \citet{zhang2019information} extensively analyzed the formulation of various forms of attention in the aforementioned categories. 
    
    \citet{lu2016hierarchical} utilizes image-question co-attention to join the embeddings. Word-to-region attention \cite{peng2019word} bridges the gap between keywords detected in questions with image regions. \citet{anderson2018bottom} introduced bottom-up top-down attention. \citet{malinowski2018learning} introduced “hard attention” to formulate visual and textual attention by filtering redundant visual features. The mechanism works as an effective feature selection method for visual features, showing great performance on cluttered and noisy data. \citet{rahman2021improved} utilizes bounding boxes along with the visual input to extract visual features and compute attention over the existing attention vectors as a form of co-attention.

    \subsubsection{Transformer}
    \label{sec:transformer}
    Transformers form the foundation of contemporary VQA architecture. \citet{vaswani2017attention} proposed that layers of self-attention are sufficient for language encoding and can outperform state-of-the-art sequential models in NLP tasks like Machine translation, language parsing, etc. The transformer architecture is fundamentally an encoder-decoder architecture based on a variant of attention called self-attention. The primary advantage of using transformers as text encoders is parallelization of the processing which leads to significantly reduced training time. In contrast, the sequential processing of RNN-variants \cite{rumelhart1985learningRNN, hochreiter1997longLSTM, chung2014empirical} made training them considerably slow. Furthermore, the transformer architecture was able to capture long-range dependencies surpassing the memory techniques used in LSTMs \cite{hochreiter1997longLSTM} and GRUs \cite{chung2014empirical}. Similar to RNNs the transformer architecture has several variants like BERT \cite{devlin2018bert}, GPT \cite{radford2018improving}, RoBERTa \cite{liu2019roberta}, T5 \cite{raffel2020exploring}, etc.

    Transformers were not limited to the textual domain as \citet{dosovitskiy2020image} proposed the Vision Transformer (ViT) which became a popular choice for visual feature extraction. Furthermore, transformer-based models were widely employed for fusing the visual and textual modalities. \citet{lu2019vilbert} relied on Cross-Modal Transformers (CMT) using cross-attention in their fusion scheme. The architectures differed by the number of transformer layers or the type of attention used. The CLIP architecture \cite{radford2018improving} aligns vision-language modalities and has been a popular module in zero-shot VQA methods. 
    
    The underlying equations behind the transformer architecture and its derivatives have been discussed extensively in different literature. For the sake of conciseness, the associated equations will not be covered in our work. Similar to CNNs and RNNs, we will treat transformers can black boxes. 
    
    \subsubsection{Vision Language Pre-training (VLP)}
    \label{sec:vlp}
    Following the success of pre-trained models in Computer Vision (CV) and Natural Language Processing (NLP) \cite{devlin2018bert, radford2018improving}, work began on vision-language pretraining for tasks like Image Captioning, VQA, and Visual Reasoning. The goal of pretraining is to train a large-scale generalized model on multiple tasks using a large volume of data. The resulting pre-trained model will be further fine-tuned on downstream tasks. Popular choices for pre-training tasks include Masked Language Modeling (MLM) \cite{taylor1953clozeMLM, devlin2018bert}, Masked Vision Modelling (MVM) \cite{chen2020uniter}, and Vision-language Marching (VLM) \cite{li2021alignALBEF}. Apart from VQA, popular downstream vision-language tasks include Visual Captioning \cite{vinyals2015show, hossain2019comprehensiveImageCaptioning}, Visual Entailment \cite{xie2019visualEntailment}, Visual Commonsense Reasoning \cite{zellers2019recognitionVCR}, Visual Grounding \cite{zhu2016visual7w}, etc; some of them will be discussed extensively in sec-\ref{sec:RelatedVLtasks}.

\begin{figure*}[htpb]
    \centering
    \includegraphics[width = 0.85\linewidth]{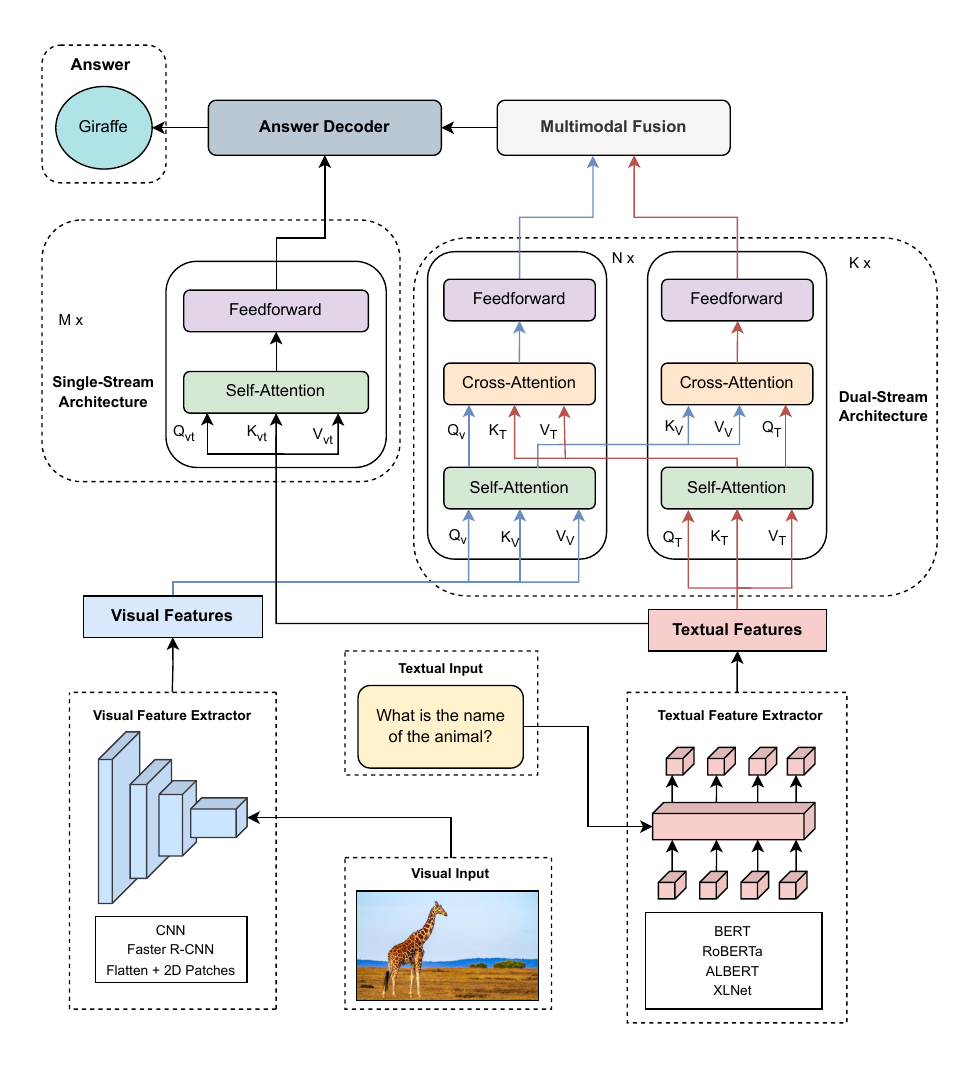}
    \caption{
    Generalized Vision Language Pre-training (VLP) network encapsulating both Single-Stream and Dual-Stream architectures. The $Q$, $K$, and $V$ vectors represent the query, key, and value vectors respectively for the transformer architecture \cite{vaswani2017attention}. The associated subscripts $v$ and $t$ stand for the visual and textual inputs respectively. The architecture incorporates two feature extractors followed by transformer blocks for encoding. The output is followed by fusion and decoder modules that can be optional based on the architectural design. 
    }
    \label{fig:vlpArchi}
\end{figure*}

\begin{table*}[p]
\label{tab:vlpTable}
\centering

\caption{Architectural overview of VLP methods.}
\begin{threeparttable}

\begin{tabular}{|l|ll|l|l|l|}
\hline
\textbf{Model Name} & \multicolumn{1}{l|}{\textbf{Text Encoder}} & \textbf{Visual Encoder} & \textbf{\#Streams} & \textbf{Fusion Encoder/Scheme} & \textbf{Attention} \\ \hline
ViLBERT \cite{lu2019vilbert}                                                        & \multicolumn{1}{l|}{BERT}                                   & FR-CNN                                   & \multicolumn{1}{l|}{Dual}             & \multicolumn{1}{l|}{Cross-Modal Transformer} & Co                 \\ \hline
VisualBERT \cite{li2019visualbert}                                                  & \multicolumn{1}{l|}{BERT}                                   & FR-CNN\tnote{a}                                  & \multicolumn{1}{l|}{Single}           & \multicolumn{1}{l|}{BERT}                        & Self               \\ \hline
Unicoder-VL \cite{li2020unicoder}                                                    & \multicolumn{1}{l|}{BERT}                                   & FR-CNN                                   & \multicolumn{1}{l|}{Single}           & \multicolumn{1}{l|}{BERT}                        & Self               \\ \hline
LXMERT \cite{tan2019lxmert}                                                       & \multicolumn{1}{l|}{BERT}                                   & FR-CNN                                   & \multicolumn{1}{l|}{Dual}             & \multicolumn{1}{l|}{Cross-Modal Transformer} & Self + Cross       \\ \hline
VL-BERT \cite{su2019vlBert}                                                    & \multicolumn{1}{l|}{BERT}                                   & FR-CNN + Resnet                          & \multicolumn{1}{l|}{Single}           & \multicolumn{1}{l|}{BERT}                        & Self               \\ \hline
Unified-VLP \cite{zhou2020unified}                                                   & \multicolumn{1}{l|}{UniLM}                                  & FR-CNN                                   & \multicolumn{1}{l|}{Single}           & \multicolumn{1}{l|}{Unified Transformer}         & Self               \\ \hline
UNITER \cite{chen2020uniter}                                                        & \multicolumn{1}{l|}{BERT}                                   & FR-CNN                                   & \multicolumn{1}{l|}{Single}           & \multicolumn{1}{l|}{BERT}                        & Self               \\ \hline
OSCAR \cite{li2020oscar}                                                          & \multicolumn{1}{l|}{BERT}                                   & FR-CNN                                   & \multicolumn{1}{l|}{Single}           & \multicolumn{1}{l|}{BERT}                        & Self               \\ \hline
ViLT \cite{kim2021vilt}                                                         & \multicolumn{1}{l|}{ViT}                                    & Patch Embedding                        & \multicolumn{1}{l|}{Single}           & \multicolumn{1}{l|}{ViT}                         & Self               \\ \hline
ALBEF \cite{li2021alignALBEF}                                                        & \multicolumn{1}{l|}{BERT}                                   & ViT                                      & \multicolumn{1}{l|}{Dual}             & \multicolumn{1}{l|}{Cross-Modal Transformer} & Self + Cross       \\ \hline
BLIP \cite{li2022blip}                                                          & \multicolumn{1}{l|}{BERT}                                   & ViT                                      & \multicolumn{1}{l|}{Dual}             & \multicolumn{1}{l|}{Cross-Modal Transformer} & Self + Cross       \\ \hline
VLMO \cite{bao2022vlmo}                                                         & \multicolumn{1}{l|}{Word-Piece \cite{wu2016googleWordPiece}}  & Patch Embedding                                                                  & \multicolumn{1}{l|}{Single}           & \multicolumn{1}{l|}{MoME Transformer}            & Self               \\ \hline
OFA \cite{wang2022ofa}                                                            & \multicolumn{1}{l|}{Byte Pair Encoding}                                    & ResNet                                   & \multicolumn{1}{l|}{Single}           & \multicolumn{1}{l|}{Unified Transformer}         & Self               \\ \hline
Unified-IO \cite{lu2022unified}                                                           & \multicolumn{1}{l|}{Sentence-Piece \cite{kudo2018sentencepiece}}                                    & VQ-GAN \cite{esser2021taming}                               & \multicolumn{1}{l|}{Single}           & \multicolumn{1}{l|}{T5-based \cite{raffel2020exploring} Transformer}         & Self               \\ \hline

PaLI \cite{chen2022pali}                                                          & \multicolumn{1}{l|}{mT5 \cite{xue2020mt5}}                                    & ViT                                   & \multicolumn{1}{l|}{Single}           & \multicolumn{1}{l|}{Unified Transformer}         & Self               \\ \hline

\end{tabular}
\begin{tablenotes}
   \item[a] Based on the ResNext architecture \cite{xie2017aggregatedResNext}  
\end{tablenotes}

\end{threeparttable}
\end{table*}

    \subsection{Taxonomy of VQA Architectures}

    Categorizing VQA architectures can be challenging due to the variety of approaches present in the literature. Earlier methods often adopted joint embedding architectures, where the two input streams are independently processed and fused. In contrast, modern VLP architectures incorporate transformer-based encoders, fusers, and decoders. Works \cite{malinowski2014towards} preceding the deep learning era will not be covered in the review. The domain of VQA can be subdivided into the traditional CNN-RNN-based, the subsequent CNN-BERT-based, and the VLP-based architectural paradigms that will be discussed in the following subsections.
    
    \subsubsection{CNN-RNN-based Architecture}

    Throughout the years, the foundations of VQA shifted from neural networks to transformers. Earlier deep-learning methods focused on answer generation from the fusion of visual and textual streams \cite{ren2015exploring, antol2015vqa}. The joint embedding scheme using two encoders is the foundation of traditional VQA architectures. In the pre-transformer era, VQA models employed usually relied on two separate feature extractors for the two modalities \cite{zhou2015simple, yang2016stacked} and incorporated some form of attention mechanism \cite{chen2015abc}. The visual and textual embeddings preserve their own identity before being fed to the multimodal fusion block. The answer generation module was generally formed of multiple fully connected layers with a classifier head to select the correct answer from the top $k$ answers of the dataset \cite{ren2015exploring, antol2015vqa}. 

    Deep learning-based methods started with \textbf{CNN-RNN-based} models where visual feature extraction was done with a variant of CNN \cite{lecun1998gradient} while textual encoding was performed using a variant of RNN \cite{rumelhart1985learningRNN}. CNNs are used due to the sparseness of the convolutional layers making feature extraction more efficient for visual inputs. GoogLeNet \cite{szegedy2015going}, VGGNet \cite{simonyan2014very}, ResNet \cite{he2016deep}, etc. were popular CNN variants for visual encoding followed by specialized object detection networks based on R-CNN \cite{girshick2014rich} like Fast R-CNN \cite{girshick2015fast} and Faster R-CNN \cite{ren2015faster}. The additional information provided by object detection networks resulted in richer visual features.
    
    Recurrent Neural Networks (RNNs) \cite{rumelhart1985learningRNN} were popular in working with sequential data like texts. The two most popular variants of RNNs were Long Short-Term Memory Networks (LSTMs) \cite{hochreiter1997longLSTM} and Gated Recurrent Units (GRUs) \cite{chung2014empirical}. The gated memory used in both variants solved the vanishing gradient problem in the vanilla RNNs. While LSTMs were popular choices for encoding the textual input in VQA \cite{ren2015exploring, antol2015vqa}, GRUs gained popularity in the upcoming years due to having fewer parameters and requiring less training time.

    Following eq- \ref{eq:visualFeatureExtractionEquation}, \ref{eq:textualFeatureExtractionEquation}, we can rewrite the visual and textual encoding equations for CNN-RNN-based architectures as

    \begin{equation}
        h_v = CNN(v)
    \end{equation}
    \begin{equation}
        h_q = RNN(q)
    \end{equation}

    During the multimodal fusions, vector operations like Element-wise Addition (EWA), Element-wise Multiplication (EWM) or Hadamard Product, and Vector Concatenation (VC) are used to join the modalities. Certain vector operations impose restrictions on the encoding dimensions. For instance, the dimensions for visual and textual encodings must match to perform EWA and EWM. Although VC has no such restriction, it increases the dimensions of the fused encoding. Following eq-\ref{eq:fusionBaseEquation}, the corresponding equations for the EWA, EWM, and VC are:

    \begin{equation}
        \Phi(h_v, h_q) = h_v + h_q
    \end{equation}
    \begin{equation}
        \Phi(h_v, h_q) = h_v \odot h_q
    \end{equation}
    \begin{equation}
        \Phi(h_v, h_q) = [h_v,h_q]
    \end{equation}

    \input{TabResOld}

    Neural network-based architectures like CNNs and LSTMs are also popular choices for fusing the multi-modal input. \citet{malinowski2015ask} fed CNN features and input text to an LSTM to generate the answer where the LSTM used the image features as the previous state input. While CNNs have been less popular fusion modules, \citet{ma2016learning} proposed a VQA architecture using CNNs only. The corresponding fusion equations for LSTMs and CNNs are:
    \begin{equation}
        \Phi(h_v, h_q) = LSTM(h_v, h_q)
    \end{equation}
    \begin{equation}
        \Phi(h_v, h_q) = CNN(h_v, h_q)
    \end{equation}

    Subsequent fusion strategies relied on bilinear pooling techniques like Multimodal Compact Bilinear pooling (MCB) \cite{fukui2016multimodalVQAVisualGrounding}, Multimodal Factorized Bilinear pooling (MFB) \cite{yu2017multi}, Multimodal Low-rank Bilinear attention (MLB) \citet{kim2016hadamard}, MUTAN \cite{ben2017mutan}, Multi-modal Factorized High-order pooling (MFH) \cite{yu2018beyond}, etc. Apart from the fusion strategies, the traditional architectures relied on various forms of attention mechanisms discussed extensively in sec-\ref{sec:attention}. Following sec-\ref{sec:answerGeneration}, answer generation was primarily treated as a classification task using the softmax or sigmoid function \cite{ren2015exploring} while many approaches used LSTMs for generative answers \cite{malinowski2015ask}.

\subsubsection{CNN-BERT-based Architecture}

    Transformers revolutionized the landscape of natural language processing (NLP) and were quickly popularized for textual encoding in VQA. The VQA methodologies were remolded by a variant of the transformer architecture called Bidirectional Encoder Representations from Transformers (BERT) \cite{devlin2018bert}. In the subsequent \textbf{CNN-BERT-based} architectures, BERT \cite{devlin2018bert} became the primary choice for textual encoding along with a latter variant of CNN \cite{girshick2014rich, girshick2015fast} for visual encoding. The language encoding equation based on eq-\ref{eq:textualFeatureExtractionEquation} can be redefined for this architectural paradigm as - 

    \begin{equation}
    \label{eq:BERTfeatureExtraction}
        h_q = BERT(q)
    \end{equation}
    
    Derivatives of the transformer like multi-modal transformers \cite{bao2022vlmo}, Cross-Modal Transformers (CMTs), and BERT were employed for the fusion of the multimodal input streams. CMTs relied on bi-directional cross-attention along with self-attention. Cross-attention calculates the attention weights based on the inputs of one modality and the comparison with the inputs of the other modality. 

\subsubsection{VLP-based Architecture}
    The scope of transformers was not limited to the textual modality only as \citet{dosovitskiy2020image} showed promising results with the Vision Transformer (ViT) architecture for visual feature extraction. ViT showed scope for researchers to incorporate transformer-based vision-language extractors introducing \textbf{dual-transformer-based} models and subsequently, led to the emergence of VLP architectures as discussed in section-\ref{sec:vlp}. Following eq-\ref{eq:visualFeatureExtractionEquation}, the visual encoding can be rewritten as,
    
    \begin{equation}
        h_v = ViT(v)
    \end{equation}

    VLP architectures can be either \textbf{dual-stream} or \textbf{single-stream} depending on how the input modalities are treated. Dual-stream architectures rely on separate multimodal streams sent to independent transformer blocks and the resulting encodings are fused at a subsequent phase. On the other hand, single-stream architectures feed the concatenated multimodal input to a single transformer block utilizing merged attention for fusion \cite{chen2023vlpSurvey}. 
    
    ViLT \cite{kim2021vilt} uses a single ViT transformer as the primary visual and textual feature extractor treating the whole input as a single stream while architectures like ALBEF \cite{li2021alignALBEF} and BLIP \cite{li2022blip} rely on two separate streams processed separately by ViT and BERT as visual and textual encoders respectively. Single streams architecture often exhibits \textit{monolithic} architectural paradigms i.e. there is a single unified transformer block \cite{chen2020uniter}. 
    
    Another crucial design choice is between the encoder-only architectural paradigm where the encoding is directly fed to the output layer and the encoder-decoder architectural paradigm where the encoding is fed to a decoder module for answer generation. An interesting work worth mentioning is VLMO \cite{bao2022vlmo} which unifies the dual and fusion encoders into a single Mixture of Modality Expert (MoME) transformer and achieved promising results.
    
\subsection{Performance Analysis}

The performance of VQA models has evolved rapidly in recent years. The VQA dataset \cite{antol2015vqa} had been a popular benchmark to evaluate the early joint embedding and attention-based models as illustrated in table-\ref{tab:performanceVQATrad}. The vanilla CNN-RNN-based \cite{ren2015exploring, malinowski2015ask} architectures achieved close to 60\% VQA accuracy on the test-dev and test-std splits. Neural module network-based architectures \cite{andreas2016neural} achieved similar accuracy. The architectures employed bilinear pooling and boosted the score beyond 65\%.

In a few years, VQAv2 \cite{goyal2017making} VQAv1 as the popular benchmark, and the performance of associated VQA models is discussed in Table-\ref{tab:performanceVLP}. It should be noted that the table contains exclusively VLP architectures as it is impossible to achieve those high levels of accuracy with pre-training. The earlier models achieved beyond 70\% accuracy on the test-dev and test-std splits of the VQAv2 dataset which is substantially higher than the non-VLP models in the VQA dataset. Furthermore, VQAv2 is more than VQAv1 due to the absence of linguistic bias.

Currently, the best models are achieving around 84\% accuracy but have severely underperformed in the zero-shot setting. The VLP model PaLI \cite{chen2022pali} achieved state-of-the-art results in the traditional setting while the multimodal LLM GPT-4 \cite{openai2023gpt4} achieved the same in the zero-shot setting. Based on the performance results, a shift towards multimodal LLMs with specialized fine-tuning strategies for VQA can be expected to be introduced in the coming years.

\input{TabResNew}

\section{Evaluation Metrics}
\label{sec:evaluationMetrics}
VQA is often associated with the Visual Turing Test \cite{malinowski2014towards} that requires a well-defined and sophisticated evaluation metric identical to human evaluation. Idealistically, the metric should produce a single score for each generated answer based on its similarity with a human-annotated answer. The scores can be further aggregated to produce a single score for a particular model.

Evaluation of natural language answers isn't straightforward. For instance, the answers "cat" and "feline" might be semantically similar but the answer "feline" is rarely used by humans; making the answer undesirable. On the other hand, semantically similar words shouldn't be harshly discouraged. The answers "cat" and "cats" are two separate yet semantically similar words. If one of the words produces a high metric score, then the other word should also produce a significantly high metric score. The following subsections will discuss the evaluation of classification and generative answers.

\begin{table*}[ht]
\label{tab:evalMetric}
\caption{Overview of various Evaluation Metrics used in VQA}
\centering
\begin{tabular}{|l|l|l|}
\hline
\textbf{Metric}                                                                             & \textbf{Measures}                                                                                  & \textbf{Use Case}                                                                        \\ \hline
Accuracy                                                                                    & Exact answer match                                                                                 & MCQs, Single answer annotation                                                           \\ \hline
VQA-Accuracy                                                                                & No. of answers matched (at most 3)                                                                 & Multi answer annotation                                                                  \\ \hline
Wu-Palmer Similarity (WUPS)                                                                 & Semantic connotation difference                                                                    & Softer form of accuracy                                                                  \\ \hline
\begin{tabular}[c]{@{}l@{}}BiLingual Evaluation \\ Understudy (BLEU) \cite{papineni2002bleu}\end{tabular}           & \begin{tabular}[c]{@{}l@{}}N-gram co-occurrences between\\ actual and predicted answer\end{tabular} & Long generative answers                                                                  \\ \hline
Mean Per Type (MPT)                                                                         & \begin{tabular}[c]{@{}l@{}}Arithmetic/Harmonic mean \\ for individual question type\end{tabular}   & Unbalanced question types                                                                \\ \hline
\begin{tabular}[c]{@{}l@{}}Average Normalized \\ Levenshtein Similarity (ANLS)\end{tabular} & Levenshtein distance                                                                               & \begin{tabular}[c]{@{}l@{}}Reasoning evaluation\\ Smooth error penalization\end{tabular} \\ \hline
F1-Score                                                                                    & Harmonic mean of precision and recall                                                              & Biased datasets                                                                          \\ \hline
Human Judgement                                                                             & Subjective human opinion                                                                           & Subjective answers, controversial topics                                                   \\ \hline
\end{tabular}
\end{table*}

\subsection{Evaluation of Classification Answers}
The standard accuracy metric has been a popular choice to evaluate classifier-based VQA models. Vanilla accuracy disregards the semantic similarities between words and only considers a question correct when the predicted answer \emph{exactly} matches the ground truth. WUPS score is a softer form of accuracy that accredits partially correct answers. VQA accuracy has been a popular metric that regards an answer to be correct based on the number of answer annotations it matches. In such datasets, a single question must have multiple answer annotations. The VQA dataset \cite{antol2015vqa} ensured 10 answer annotations per question and deemed an answer to be fully correct if it matches at least 3 of the 10 answer annotations. Otherwise, the answer is partially correct based on the number of matched annotations. 

\subsection{Evaluation of Generative Answers}
\label{sec:genAnsEval}
Generative answers are tougher to evaluate, especially if the generated answer is long. Machine translation metrics like BLEU score \cite{papineni2002bleu} are often employed to find the similarity between ground truth and the generated answer. Apart from assessing the similarity, the models face challenges in evaluating subjective and controversial answers. Controversial and unsafe answers are usually avoided in VQA with limited work covering these topics. Nevertheless, with the increasing popularity of multimodal LLMs \cite{huang2023languageKosmos1, openai2023gpt4}, both VQA and Visual Dialog systems need to develop robust metrics for generative answer evaluation.

\section{Challenges and Opportunities}
    \label{sec:Challenges}
    Being a multimodal task, VQA faces the classical multimodal challenges -- the representation of multimodal data, the fusion of the modalities, and co-learning between the two modalities \cite{baltruvsaitis2018multimodalMachineLearning}. However, this section will explore some of the domain-specific challenges along with possible mitigation strategies. 

    \subsection{Dataset Challenges}
    \label{sec:DatasetChallenges}

    The biggest and oldest challenge in VQA dates back to its early days when there was an unavailability of large-scale datasets that could be used as benchmarks. Researchers aimed to create a dataset that represents realistic or natural images with a diverse range of QA pairs. The seemingly straightforward challenge was first attempted by \citet{malinowski2014multi} with the DAQUAR dataset, although the dataset was restricted to indoor scenes only. The VQA dataset \citet{antol2015vqa} gained widespread popularity but exhibited imbalanced class distribution. The simple task of establishing a generalized open-ended benchmark turned out to be far more difficult. 
    
        \subsubsection{Generalized Open-Ended Benchmark}
        \label{sec:genDatasetBenchmark}
    
        To get a better grasp of the difficulty in establishing a benchmark, let's consider a hypothetical dataset that, in order to be idealistic, must contain visuals of every conceivable visual class and concept. For instance, if images of zebras are absent in our hypothetical dataset, the trained model will not \emph{know} what a zebra is. Similarly, the training set must cover concepts and activities like reading, walking, etc. The challenge will seem overwhelming when one can comprehend the vast number of existing visual classes and concepts. The hypothetical dataset must also be linguistically sound i.e. the associated model trained on the dataset must be able to comprehend questions containing \emph{any} word as long as the question is semantically correct. 
        
        Altogether, the hypothetical VQA dataset must be sufficiently large to encapsulate all visual and linguistic entities along with their associated intricacies. Although creating such a dataset might seem practically impossible, there can be a workaround by decoupling the training and testing phases. Prior to standard VQA training, contemporary methods pre-train the models on various vision and language tasks with larger datasets. As discussed in section-\ref{sec:vlp}, vision language pre-training (VLP) \cite{gan2022visionVLPSurvey,chen2023vlpSurvey} enables the models to transfer knowledge from other related tasks and understand concepts beyond the limited VQA training data. However, the challenge of establishing a generalized VQA dataset is now reformulated to the challenge of establishing a standard VLP dataset.
    
        \subsubsection{Evaluation Dataset}
        \label{sec:generalizedEvalDataset}
        
        The test split should be capable of rigorously evaluating the performance of models in various VQA sub-tasks by reflecting the obstacles in realistic scenarios. As VQA models become increasingly capable, the task of evaluation becomes increasingly difficult. Similar to designing a VQA benchmark, designing an ingenious evaluation dataset that is difficult for the state-of-the-art VLP models can be equally challenging. We should also consider that a large amount of training data might provide high accuracy in the benchmarks, but a higher accuracy cannot always be associated with a better architecture. The testing split should be able to differentiate such models and evaluate other aspects of the models apart from the accuracy score. 

        \subsubsection{VLP Dataset}
        \label{sec:vlpChallenges}
         VLP datasets are usually large-scale datasets of image-text pairs automatically extracted from online sources. As quality assurance is difficult for such large-scale datasets, it often leads to image-text misalignment issues and data redundancy \cite{wang2023too}. VLP datasets are also primarily English-based and translations of native English answers are usually erroneous \cite{chen2022pali}. Current training strategies relying on multilingual captioning \cite{thapliyal2022crossmodal} use human-annotated captions to overcome annotation artifacts. However, cross-lingual performance is still low compared to its native English counterparts \cite{chen2022pali}. Potential research directions may include introducing non-English VLP datasets or improving cross-lingual capabilities by enhancing cross-lingual model architecture.

   \subsection{Med-VQA Challenges}
   \label{sec:medVQAChallenges}
   The medical domain is a promising field for VQA with its own set of challenges that will be discussed in this section,

        \subsubsection{Med-VQA Dataset}
         Med-VQA faces challenges similar to classical VQA in creating a large-scale dataset that encompasses medical images from every category. Datasets dedicated to radiology \cite{lau2018datasetVQARAD} and pathology \cite{he2020pathvqa} contain actual X-rays, CT scans, MRIs, and textbook-extracted pathological images. However, as the medical domain is more diverse, the aforementioned datasets do not necessarily encapsulate the whole domain. Furthermore, certain factors can make the creation of medical VQA datasets more challenging than that of standard VQA. Firstly, collecting high-quality annotated medical data is difficult and expensive. Secondly, the generalization of medical data is not as straightforward as standard VQA data since images might have subtle changes that can drastically change the generated answer.

        \subsubsection{Multimodal Medical LLMs}
         Med-VQA is also experiencing the trend of using LLMs and their multimodal counterparts \cite{zhang2023biomedgpt}. LLMs can be used as modules for comprehensive answer generation while MLLMs can be deployed as an end-to-end VQA system. BiomedGPT \cite{zhang2023biomedgpt} and LLaVa-Med \cite{li2023llava} are examples of multimodal LLM-based architectures dedicated to the medical domain and have shown great performance in Med-VQA and Medical Image Dialog. Nevertheless, these LLMs come with their own set of problems, primarily hallucination \cite{li2023evaluatingHallucination} which can be more detrimental in the medical domain. Hallucinations occur when LLMs produce misinformation or erroneous responses with high confidence. The penalty of misinformation can be dire in the medical domain, potentially costing the life of a patient. 

        \subsubsection{Miscellaneous}
        Med-VQA is considerably closer to knowledge-based VQA (KB-VQA) as Med-VQA questions often require medical knowledge along with a high degree of image understanding and answer precision. Medical data is also constantly being updated which might make the models outdated if no form of online learning is incorporated. Online learning enables machine learning models to continuously update their parameters by training on real-time data. Establishing a system based on online learning in the medical domain has its challenges, given the fact that there can be no room for misinformation and error.

        \subsection{Model Evaluation Challenges}
        The improved performance of state-of-the-art models increased the difficulty in model evaluation. A higher accuracy score does not imply a better architecture and other factors should be considered along with accuracy. This subsection will be centralized around the evaluation practices in the current VQA literature.

        \subsubsection{Generalization, Robustness, and Consistency}
            Historically, models were evaluated based on their accuracy scores \cite{antol2015vqa}. However, relying solely on accuracy doesn't ensure the evaluation of a model's generalization capabilities, consistency, and robustness \cite{gupta2022grit}. Models with more parameters trained on larger datasets may have better accuracy scores but might be prone to noise and corruption effects on the input modalities. Furthermore, the models may not be able to predict beyond training data or consistently produce correct answers. 
        
            Generalization is defined as the ability of the model to extend its performance beyond the distribution of the dataset i.e. its extensibility in a novel setting. Zero-shot settings \cite{teney2016zero} or specialized metrics \cite{gupta2022grit} can be used to evaluate the generalization capabilities of a model. On the other hand, robustness is defined as the reluctance of the model to change its prediction when introduced to artifacts, corruptions, or noise in the multimodal inputs. Consistency refers to the logical coherence of the answers i.e. related to visual entailment.
        
            Both robustness and consistency in the textual domain have been thoroughly evaluated in the VQA literature \cite{huang2019assessing, jimenez2022carets} as VQA models often struggled with paraphrased questions, syntactic errors, grammatical errors, etc. Robustness also highlights the challenges of VQA models in realistic settings as seen in VizWiz \cite{gurari2018vizwiz}, emphasizing the fact that real-life data and questions are not always in idealistic conditions. The evaluation of the generalization, robustness, and consistency of VQA models is regaining interest with the trends of VLP and multimodal LLMs in VQA \cite{zhao2023evaluatingRobustnessVLP}. 
            
        \subsubsection{Generative Answer Evaluation Metric}
            VQA has been traditionally defined as a classification problem that simplifies answer generation by limiting it to the top $k$ answers of the dataset \cite{ren2015exploring}. Discriminative models dominated the classification problem formulation until the recent rise of generative models \cite{radford2018improving}. Generative models can produce high-quality long answers suitable to elucidate "how" and "why" type questions. Due to the shift from the classification formulation, VQA lost the simplicity of evaluating classes \cite{antol2015vqa} as free-form answers are harder to evaluate. Following the discussion in sec-\ref{sec:genAnsEval}, evaluation of generative answers is difficult considering subjectivity and the variations in answer style.
      
    \subsection{Rational VQA}
    A utopian goal in VQA is to design VQA models as rational as human beings which can be done by addressing the bias in datasets, questioning the reasoning capabilities of the models, and understanding the rationale behind answer generation. 

    \subsubsection{Bias in VQA}

    A dataset can inherently have redundant patterns that we do not want our model to follow. Training a VQA model is analogous to teaching children -- often when trying to teach a child to do something in a specific way, they may pick up a few "shortcuts" that will allow them to complete that particular task faster but in a different way. However, disregarding the intended way may affect their performance in a different yet similar task. A VQA model being trained from a biased dataset might perform well using shortcuts for that particular dataset or setting but consequently, will lose generalization capabilities by showing subpar performance in other settings.

    For a better illustration of dataset bias, let us consider a dataset where all the color-based questions on apples have the answer "red". A model trained on this dataset will associate the color red with apples. If the image of a green apple is shown to the model and a color-related question is asked, the model will have a high probability of predicting "red" due to the strong correlation between the textual color-based question and the textual answer. Models are more likely to find such correlations between the same modality which, in this case, is the textual modality. Such a model can be considered \emph{blind} as the model is unable to utilize the visual information. 
    
    VQA models typically exhibit an association of textual questions with textual answers while ignoring the visual input \cite{zhang2016yin, goyal2017making}. This tendency is termed linguistic bias which usually occurs due to dataset \emph{imbalance}. \citet{goyal2017making} used counterexamples in their dataset to overcome linguistic bias. Redistributing the train, validation, and test splits has also been a popular strategy in mitigating bias \cite{dancette2021beyondVQACE}. 

    Recent studies have shed light on other forms of bias, posing challenges for modern VQA architectures. \citet{hirota2022genderRacialBiasVQA} highlighted gender and racial bias in VQA datasets, emphasizing the need to address these issues to develop \emph{safe} VQA models. Models trained on a particular often exhibit their own styles in answering the question that might result in cross-dataset mismatch \cite{chao2018crossDataset}. The design focus should be towards creating datasets that are inclusive to everyone, regardless of their background.

    \subsubsection{Model Reasoning}

    With the advancements in model architecture and increased model accuracy in difficult VQA tasks, one might argue whether VQA models are actually capable of logical reasoning over multimodal data or simply encapsulate and extend the patterns found in the training data. Do models have the capabilities to \emph{understand} the question-answering paradigm on a visual input? One may not expect human-like modeling of the world but with the increased generalization capacities, one can expect close to human-like reasoning. Several datasets \cite{johnson2017clevr, pandhre2017shapes, zhang2019raven} challenged the rationale of VQA models and their reasoning capabilities. Researchers continue to work on improving the logical consistency of the models and their ability to reason over complex problems.

    \subsubsection{Model Explainability}
    A VQA model should be able to explain or refer to particular data on how it came to a particular conclusion. A model is termed explainable if during an inference the user can perfectly understand the logical sequence of the input data being processed to the answer output. Large unified architectures like BLIP \cite{li2022blip}, Unified-VLP \cite{lu2022unified}, etc., and multi-modal LLMs like GPT-4 \cite{openai2023gpt4}, KOSMOS-1 \cite{huang2023languageKosmos1}, took both the accuracy and generalization capabilities of VQA models to new heights at the cost of model explainability. Several works \cite{li2018tellExplainableVQA, goyal2016towardsExlainableVQA} addressed model explainability in VQA while proposing explainable architectures. However, state-of-the-art models \cite{wang2022ofa, wang2022imageBeit, bao2022vlmo} are changing the explainability landscape of VQA as the models are viewed as a black box or a network of black-box modules. 

    \subsection{Zero-shot VQA (ZS-VQA)}

    Generalization is an important concept in the field of machine learning and is key to the development of Artificial General Intelligence (AGI) \cite{goertzel2007artificial}. Every machine learning model is expected to exhibit some form of generalization as the models are tested on a different split of the dataset while testing. While the particular samples in the test split might be different, both the training and testing splits usually come from the same data distribution. 

    \subsubsection{ZS-VQA using Out-of-distribution Data}
    
    Let us consider an example where a machine learning model is expected to predict based on integer values within the input range $[10,30]$. During training, the model has seen all the discrete sample values excluding $13$ and $15$. If these particular values are used while testing, we can say that the testing samples are \emph{different} as they were not present in the training set. Nevertheless, their distribution is the same as that of the training set. On the contrary, if a large value like $100$ is fed to the model while testing, we can conclude the model is being tested on \emph{out-of-distribution} data.

    Zero-shot VQA is a problem setting for VQA where the performance of the model is evaluated on data that is beyond the training data of the model. Going beyond the training data can be defined in multiple ways, one of them is using a test split which comes from a different distribution than the training split \cite{teney2016zero}. Formally, for the VQA model trained on samples from the joint probability distribution $\mathcal{D}_{V, Q}$, we define the zero-shot setting for VQA as the evaluation setting of the model on a different distribution $\mathcal{D'}_{V,Q}$.

    A VQA dataset has visual and textual components. An ideal zero-shot setting will have both visual and textual components for evaluation from a different distribution. \citet{teney2016zero} proposed an out-of-distribution (OOD) split on the question-answer set for the Visual7W dataset \cite{zhu2016visual7w}. The resplitting guarantees that there will be at least one unseen word in every question in the validation and test split. However, the visual concepts do not necessarily come from a different distribution i.e. it is not guaranteed that the model will be asked questions on unseen visual concepts. \citet{farazi2020known} extended the setting to handle novelty in both visual and semantic concepts.

    \subsubsection{ZS-VQA on VLP Models}
    \label{sec:ZSVQAVLPmodels}
    While resplitting a dataset can theoretically ensure mutually exclusive training and evaluation distribution for both modalities, it is practically difficult to achieve albeit at the cost of a reduced dataset. Notably, the zero-shot splits for Visual7W by \citet{teney2016zero} had only 25\% of the number of test images in the original split. Additionally, imposing the visual concepts to come from a different distribution further decreases the size of the test split. 

    Instead of creating out-of-distribution splits, we shall redefine the zero-shot setting for VQA by using models that weren't \emph{specifically} trained in VQA i.e. the model has been trained on other multimodal vision-language tasks like Image Captioning \cite{alayrac2022flamingo}, Image Conditioned Masked Language Modeling \cite{jin2021good}, etc. For instance, \citet{guo2023imagesFrozenZeroShot} used a frozen image encoder, a text encoder, and an LLM to achieve a zero-shot setting where the modules are \emph{frozen} i.e. they haven't been trained on the task of VQA.

    \subsubsection{Vision-Language Interfacing}

    The modular ZS-VQA architectures \cite{tiong2022plug, song2022clipFewShot} can simplify VQA as the task of interfacing between Vision and Language (VL). Images are seen as unstructured data and VQA can be deemed as an interface for extracting meaningful information from large volumes of unstructured data that connects with the textual modality. Analogous to image captioning, VQA can be considered a more specific interface due to using questions in the input. \citet{tiong2022plug} employs the subtask Image Captioning within its architecture that can be a bottleneck for more advanced ZS-VQA models. Researchers are exploring efficient ways of interfacing between vision and language beyond traditional VL tasks.

\section{VQA in the Bigger Picture}
\label{sec:vqaBiggerPicture}

We shall now explore how the domain of VQA can be perceived in the world of multi-modal problems. We try to define the boundaries of VQA and delve deeper into its subdomains.

\begin{figure}[ht]
    \includegraphics[width=\linewidth]{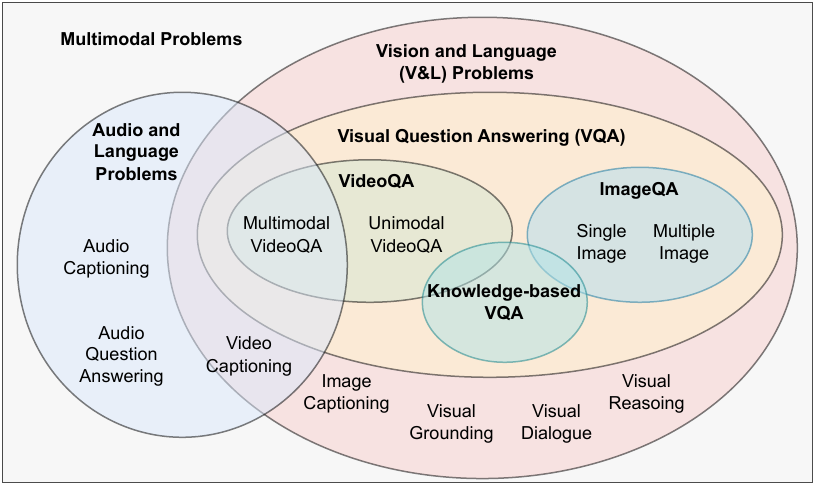}
    \caption{VQA among other multimodal problems}
    \label{fig:vqaWider}
\end{figure}

\subsection{Multimodal Question Answering (MQA)}
\label{sec:mqa}
In this literature, VQA often refers to the traditional setting where the model tries to generate an answer from a \emph{single image} and a \emph{single question} \cite{antol2015vqa, ren2015exploring}. However, as previously stated, VQA can be generalized to any form of visual input that is not limited to a single image. For instance - one of the recently evolving sub-domains of VQA, VideoQA \cite{xu2017videoCaptionQA, zhong2022video} takes a video as the visual input. Following fig-\ref{fig:vqaTasks}, the video can be either unimodal or multimodal. The key challenge of multimodal VideoQA is the processing of auditory data by a VQA system as it introduces a new modality to our previously established bimodal task. The introduction of auditory data also brings the opportunity to extend QA beyond bimodality.

Multimodal Question Answering (MQA) can be theorized as a superset of VQA where any form of multimodal input with a question is given to generate multimodal answers as the output. Let's define a multimodal question-answering model $\mathcal{M}$, such that, 
\begin{equation}
    \mathcal{M}:(X_{i_1}, X_{i_2}, ..., X_{i_m}, Q_k) \rightarrow (Y_{j_1}, Y_{j_2}, ..., Y_{j_n})
    \label{eq:multimodal}
\end{equation}
where both the input, $X$ , and output, $Y$, are multimodal from the modalities $\{i_1, i_2, .., i_m\} \in \mathbb{M}$ and $\{j_1, j_2, .., j_n\} \in \mathbb{N}$ given $\mathbb{M}, \mathbb{N}$ are sets of modalities. The unimodal questions $Q_k$ comes from the modality $k$. The associated model can be a discriminator, a generator, or a hybrid of both. The input and output modalities do not necessarily need to match e.g. the input modalities of VQA are both visual and textual while the output modality is textual only. 

MQA is not strictly equivalent to VQA as there exists a set of multimodal tasks that require no visual input thereby making them part exclusively part of MQA. A good example can be Audio/Spoken QA \cite{chuang2019speechbert} where QA is done on an auditory input only. Formally, MQA can be defined as a problem where a set of inputs from various modalities is given as the context to a free-form input question in order to generate a multi-modal answer. MQA is also a subset of multimodal tasks in general which are not limited to question answering e.g. captioning for images \cite{hossain2019comprehensiveImageCaptioning}, audio \cite{drossos2020clothoAudioCaptioning}, and video\cite{iashin2020multiVideoCaptioning} can be viewed as multimodal captioning.

There are few existing architectures that can work on MQA or most of the generalized multimodal problems. \citet{wang2023onePeace} proposed the One Peace architecture to work on any multimodal task using modality adapters. On the other hand, multimodal LLMs (\cite{openai2023gpt4, huang2023languageKosmos1, peng2023kosmos2} have been producing state-of-the-art performance on multimodal problems. Visual ChatGPT \cite{wu2023visual} works with visual and textual inputs to produce visual or textual outputs analogous to figure-\ref{fig:visDial}. \citet{maaz2023video} extends the setting of Visual ChatGPT to videos.

\subsection{Related Vision-Language (VL) Tasks}
\label{sec:RelatedVLtasks}
The task of VQA can be refined as a task that aims to combine visual and textual modalities and project back to textual modality. VQA faces challenges in both aligning and fusing the modalities \cite{baltruvsaitis2018multimodalMachineLearning}, bearing resemblance to other VL tasks. As discussed in section-\ref{sec:ZSVQAVLPmodels}, the pre-trained architectures for non-VQA tasks can be utilized for VQA in both traditional and zero-shot settings. Some of the VL tasks related to VQA shall be discussed in this section.

\subsubsection{Visual Captioning (VC)}
Visual captioning \cite{gan2017semanticVisualCaptioning} is undoubtedly the closest task to VQA and has been monumental in the construction of several VQA datasets \cite{ren2015exploring, xu2017videoCaptionQA}. VC has inspired several Joint Embedding VQA architectures \cite{malinowski2015ask, ren2015exploring} and has been a popular downstream task for vision language pertaining \cite{alayrac2022flamingo}. The primary goal of VC is to generate a syntactically and semantically correct textual description of the visual input. The associated models require recognition and understanding of visual elements and their relationships. VC can also be performed in knowledge-based settings, benefiting from knowledge retrieval methodologies. Image captioning is the most popular form of visual captioning while Video Captioning has been gaining popularity recently.

\subsubsection{Visual Commonsense Reasoning (VCR)}
VQA models need to exhibit some form of visual reasoning capabilities as many datasets \cite{johnson2017clevr, lu2021iconqa} are aimed at testing such capabilities. Often relying on complex and compositional questions, Visual Commonsense Reasoning (VCR) \cite{zellers2019recognitionVCR} can be viewed as an extension of VQA that highlights a key limitation of VQA systems -- the lack of common sense and reasoning capabilities. A commonsense-based question like "What is the color of the sun?" will often be incorrectly answered by a system specifically trained on VQA. The VCR dataset \cite{zellers2019recognitionVCR} has been instrumental in this domain followed by other generalized reasoning datasets like GQA \cite{hudson2019gqa}.

\subsubsection{Visual Grounding (VG)}
Visual Grounding (VG) \cite{zhu2016visual7w} is defined as the problem where visual elements relevant to the textual question have to be located i.e. the bounding boxes of the visual elements must be provided. The located element can be a person, action, object, etc. VG can also be considered as a specific setting of VQA. VG models must have a comprehensive understanding of visual attributes along with their location in the visual input. 

The Visual Toloka Challenge \cite{gao2023champion} is an interesting setting of VG that aims to ask a question about a visual element that is implicitly mentioned in the question. The simplicity can be achieved through logical congruence, e.g. an image of a human face with the question "What is used to smell?" will have the output bounding box surrounding the nose of the human.

\subsubsection{Inverse VQA (iVQA) and Visual Question Generation (VQG)}

The task of Inverse VQA (iVQA) \cite{liu2018ivqa} deals with generating a question for an image-answer pair that can be generalized to any visual input with a textual answer. A similar task of generating textual questions based on visual input is termed Visual Question Generation (VQG) \cite{mostafazadeh2016generating, mostafazadeh2016generating, guidingVisualQuestionGeneration}. The most straightforward strategy for VQG is to use a Visual Captioning model and incorporate automated techniques that convert textual captions to textual questions. However, the textual caption is usually generalized while the questions can have different paradigms. 

Retrieval models are often employed to use the top $k$ captions to produce multiple questions along with training end-to-end generative models \cite{mostafazadeh2016generating}. iVQA relies on similar approaches to answer generation but must consider the answer during QA generation. For VQG, the variations in questions along with their semantic and syntactic quality determine the quality of the model. iVQA emphasizes the relevancy of the answer and visual content with the generated question. 

Both iVQA and VQG can play pivotal roles in dataset creation and data augmentation along with many real-world applications like automated education systems. \citet{zeng2017leveraging} utilized video descriptions to generate synthetic QA pairs that are later used to train video QA models. Similar techniques are employed by \citet{changpinyo2022allVQAImageCaptions} relying on image captions to generate QA pairs and utilize the abundance of image-caption datasets to train large-scale VQA and Zero-Shot VQA models. VQG can be integrated as an important training module in the contemporary modular VQA architectures \cite{tiong2022plug} and to fine-tune VLP-based methods on downstream tasks \citet{wang2022ofa, chen2022pali}.

\subsubsection{Embodied and Interactive QA}
\label{sec:EmbodiedQA}
\citet{das2018embodied} introduced EmbodiedQA where an agent has to navigate through a 3D environment and answer related questions. The task shares similarities with navigation in a 360\textdegree\ video \cite{hu2017deep360} but incorporates reinforcement learning (RL) elements as the agent has to take actions learned from the environment alongside question answering. An emerging domain closely tied to embodiedQA is Interactive QA (IQA) \cite{gordon2018iqa} where an RL agent is required to navigate through an interactive scene to answer related questions. Such tasks require an RL agent's proficiency in numerous tasks such as detection, navigation, environment manipulation, question answering, and so on.

\subsubsection{Miscellenous Tasks}
Categorization of visual elements based on some relevant attributes describing e.g. categorizing clothes into t-shirts, pants, etc. is a form of product-based VL categorization \cite{zhuge2021kaleido}. Sentiment analysis has been popular in visual and textual domains but its multimodal VL setting \cite{ghosal2018contextual} has been less explored. The intersection of information retrieval (IR) with other modalities is cross-modal retrieval \cite{wang2016comprehensive, chen2022hivlp} which aims to retrieve any form of multimodal output based on a multimodal input. 

Visual Entailment \cite{xie2019visualEntailment} explores the semantic alignment of the visual input with the text. Tasks like Multimodal Machine Translation \cite{specia2016shared} aim to translate texts in visual inputs and have important use cases in realistic settings. VQA also interacts with Reinforcement Learning (RL) in tasks that require some form of action to be performed by the model. Embodied QA \cite{das2018embodied} explores such a task where an agent is randomly located in a virtual environment and asked a question for which it has to take multiple actions in the virtual environment to learn the actual answer. Following the discussion in sec-\ref{sec:VisualDialogChatbot}, Visual Dialog is also increasing in popularity with the rise of multimodal large language models (MLLMs) \cite{huang2023languageKosmos1, peng2023kosmos2}.

\subsection{Sub-domains of VQA}
\label{sec:modalitiesVQA}

As seen in section-\ref{sec:mqa}, the task of multimodal task of VQA can be extensively generalized. In this section, we shall explore different settings of VQA that are bound within the definition of VQA i.e. a visual input will be a \emph{strictly} required to answer the textual question. Usually, the subdomains of VQA occur due to variations in the visual modality.

\begin{figure}[ht]
    \centering
    \includegraphics[width=0.8\linewidth]{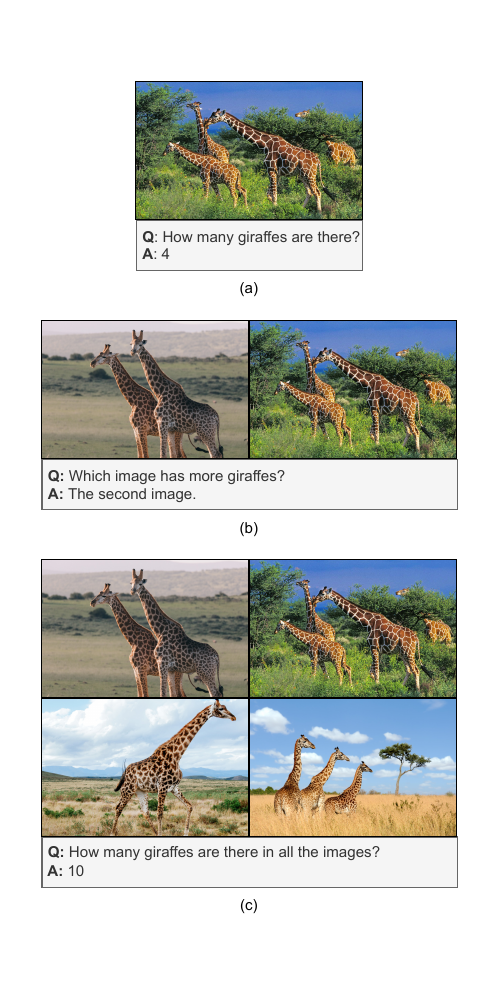}
    \vspace{-1cm}
    \caption{Comparison of VQA problems with (a) a single image, (b) a pair of images, and (c) a set of images. The problem can be specified as a counting-based problem.}
    \label{fig:imageCountingChallenge}
\end{figure}

\subsubsection{Single Image QA (SIQA)}
The traditional setting of VQA, Single Image QA, is defined as the textual answer generation or classification problem of a textual question on a \emph{single image}. SIQA has been addressed early in the VQA literature and our survey mostly covers datasets and techniques in this particular setting. Following eq-\ref{eq:vqa}, \ref{eq:multimodal}, for a VQA model, $\mathcal{M}$, an image $X_i$, a textual question $X_t$, and a textual answer $Y_t$, we can formally define SIQA as,

\begin{equation}
    \label{eq:tradVQA}
    \mathcal{M}:X_{i}, X_{t} \rightarrow Y_{t}
\end{equation}

\subsubsection{VideoQA}
VideoQA has been discussed extensively in sec-\ref{sec:videoQA}. Similar to eq-\ref{eq:tradVQA}, for a video input $X_{vid}$, the problem can be formally defined as,

\begin{equation}
    \label{eq:videoQA}
    \mathcal{M}:X_{vid}, X_{t} \rightarrow Y_{t}
\end{equation}

VideoQA can have further variations in modality based on the audio feed of the video. Some tasks require processing multimodal videos \cite{lei2018tvqa} i.e. videos with an audio feed while others don't.

\subsubsection{Change Detection QA (CDQA)}
In change detection \cite{shi2020change}, a pair of images is fed to a model that generates a binary image highlighting the areas where change occurs i.e. a pixel-wise binary classification. CDQA \cite{yuan2022change} aims to perform question-answering on such inputs and generate a textual answer instead. Given, a pair of images, $X_{i_1}, X_{i_2}$, and a question, $X_t$, the model is required to generate an appropriate answer, $Y_{t}$, related to the change in the images. Following eq-\ref{eq:multimodal}, CDQA can be defined as,

\begin{equation}
    \label{eq:cdqa}
    \mathcal{M}:X_{i_1}, X_{i_2}, X_{t} \rightarrow Y_{t}
\end{equation}

\subsubsection{Image Set QA (ISQA)}
A generalized form of CDQA where instead of a pair of images, a set of images along with a question is given as the input to a model to generate the textual answer as the output \cite{bansal2020visualImageSet}. ISQA can take various forms e.g. a question is asked about a specific image, the other images can serve as the context.
The set of images is less likely to exhibit a temporal relationship with each image analogous to a video. The problem can be formally defined as,

\begin{equation}
    \label{eq:isqa}
    \mathcal{M}:(X_{i_1},X_{i_2},...X_{i_n}, X_{t}) \rightarrow Y_{t}
\end{equation}

where, $M$ is our VQA model, $\{X_{i_1},X_{i_2},...X_{i_n}\} \in \mathbb{I}$ is a set of images, $X_{t}$ is the textual question, and $Y_{t}$ is the textual answer. ISQA can be reduced to VCQA which deals with a pair of images and a change-related question or can be reduced to any form of Image-pair VQA. 

\subsubsection{VQA 360\textdegree}
360\textdegree\ images are emerging as popular choices of visual input that expand beyond the conventional field of view found in standard images. Question answering on 360\textdegree\ images can be challenging as it requires the model to extract spatial information from visual content around the camera's optical center. Furthermore, a 360\textdegree\ dataset should not only include 360\textdegree\ images but also a diverse set of questions challenging the intrinsic properties of 360\textdegree\ images. \citet{chou2020visual360deg} introduces the task of VQA within the context of 360\textdegree\ images by presenting a natural 360\textdegree\ image VQA dataset and a novel model to perform multi-level spatial reasoning. 360\textdegree\ image VQA can be extended to 360\textdegree\ videos with audios \cite{yun2021pano}.

\subsubsection{Miscellaneous Modalities}
GIF-QA \cite{jang2017tgif} is analogous to VideoQA but processes a GIF instead of a video. GIFs do not have any audio feed and are analogous to image sequences. Single Image QA can have further variations like using images of graphs, charts, documents, infographics, etc \cite{methani2020plotqa, mathew2022infographicvqa, kembhavi2016diagramAI2D, masry2022chartqa}. QA on PDFs \cite{ding2023vqapdf} is an emerging topic that requires document processing and visual understanding capabilities equivalent to language and vision models respectively. SlidesQA \cite{tanaka2023slidevqa} proposes QA on slide decks which is similar to ISQA but requires an understanding of intra-slide and inter-slide relationships.

\section{Trends, Open Problems, and Future of VQA}
\label{sec:trendsOpenProblemsFuture}

The rapidly evolving domain of VQA has myriad open challenges and problems that future researchers should explore. The potential applications of VQA can also be excellent opportunities for engineers and entrepreneurs. VQA has seen plenty of revolutions with the advent of deep-learning-based architecture \cite{malinowski2015ask, ren2015exploring}, transformer architecture \cite{kim2021vilt, li2019visualbert, tan2019lxmert}, and currently, LLMs and generative AI \cite{guo2023imagesFrozenZeroShot, huang2023languageKosmos1}. The following section explores the trends of today's VQA systems while highlighting unexplored problems along with future research directions.

\subsection{Recent Trends}

\label{sec:trends}
 The current biggest trend in VQA is the use of Generative AI due to its recent rise in popularity. VideoQA \cite{xu2017videoCaptionQA} has also seen substantial work in recent years \cite{zhong2022video}. Zero-Shot VQA has also seen significant growth with the introduction of modular Zero-Shot architecture \cite{tiong2022plug}, graph-based methods \cite{chen2021zero}, and multimodal LLMs \cite{openai2023gpt4}. 

    \subsubsection{Dataset Trends}
    \label{sec:DatasetTrends}
The large VLP architectures \cite{li2022blip, wang2022ofa} rely on automatically mined datasets for VL tasks. Due to the higher availability of image-text pairs compared to image-QA triplets, VLP mostly relies on large captioning datasets like Conceptual Captions \cite{sharma2018conceptual}. \citet{changpinyo2022allVQAImageCaptions} established that image captioning data is sufficient to train the VLP models and provided a framework for automatic generation of VQA data at volume from captioning data. VLP often relies on a mixture of large-scale datasets \cite{chen2022pali} and the associated models are usually benefitted by scaling using a higher volume of data.

    As seen in sec-\ref{sec:vlpChallenges}, VLP datasets are shifting VQA models from the boundaries of substantially smaller VQA training data to generalized domains. Table-\ref{tab:tradDatasetQuant} highlights the decline of traditional datasets in VQA over the years. Current research works are directed toward establishing difficult evaluation benchmarks with a focus on generalization, zero-shot setting, and reasoning capabilities. The trends of VQA datasets can be attributed to the enhanced capabilities of VQA models in the past few years.
    
    \subsubsection{Cross-lingual and Multi-lingual datasets}
    
    Most of the VQA literature worked with English QA pairs and disregarded the performance of models in other languages. However, recent works have addressed this issue by proposing the VQA in cross-lingual and multi-lingual settings. \citet{pfeiffer2021xgqa} extended the standard GQA \cite{hudson2019gqa} dataset to 7 new languages. \citet{changpinyo2022towardsMAXM} proposed a QA translation framework on the multilingual image captioning dataset Crossmodal-3600 \cite{thapliyal2022crossmodal}. \citet{liu2022delving} addressed the performance gap between English and other languages by implementing simple modifications to the multi-lingual training setup. Further analysis of question types and languages highlighted a zero-shot performance gap and difficulties in answering certain question types in certain languages.

    \subsubsection{Foundational Models}
    \label{sec:foundationalModels}
    AI models are often trained on a variety of tasks with a large amount of data and are referred to as foundational models. The unimodal Large Language Model (LLM) BERT \cite{devlin2018bert} is an example of a foundational model in the textual modality and one of the first entries in the domain. State-of-the-art VQA models such as OFA \cite{wang2022ofa}, Pali \cite{chen2022pali}, BEiT-3 \cite{wang2022imageBeit}, etc. are VLP architectures having multiple pre-training objectives and can be considered as foundational models. 

    The key advantage of foundational models is the adaptability of a pre-trained model to a variety of \emph{downstream} tasks as seen in sec-\ref{sec:vlp}. This property ensures impressive performance \cite{wang2022ofa, wang2022imageBeit} that is not limited to VQA but extends to similar vision-language domains as seen in section-\ref{sec:RelatedVLtasks}. Recently,\citet{wang2023onePeace} extended the modalities further by proposing a generalized foundational model framework for multiple modalities using modality adapters and a fusion encoder. Modality adapters are modules for task-specific finetuning and have been primarily used in VQA as Vision Adapters \cite{chen2022visionAdapter}. 
    
    \subsubsection{Generative AI}
    \label{sec:genVQA}
     The generative approach in AI deals with models comprehending the patterns from the training data to generate new data. The Generative Pre-trained Transformer (GPT) \cite{radford2018improving} is one of the most popular generative textual models. In contrast, the discriminative approach is generally associated with non-probabilistic classifiers that are iteratively trained to distinguish between output classes e.g. a neural network-based classifier.  
    
     VQA is often modeled as a generative task as seen in \ref{sec:answerGeneration}. Recent trends show GPT-based LLMs \cite{brown2020language, openai2023gpt4} being adopted as modules in VQA systems primarily for Zero-Shot VQA (ZS-VQA) \cite{guo2023imagesFrozenZeroShot}. Furthermore, multimodal LLMs discussed in the next subsection are generative models showing promising results in traditional VQA, ZS-VQA, and Visual dialogue as seen in \ref{sec:VisualDialogChatbot}.
     
     \subsubsection{Multimodal LLMs (MLLMs)}

     Researchers are aiming to extend the capabilities of LLMs to other modalities and coined the term Multimodal LLM to define these LLMs. The successor to the popular ChatGPT model, GPT-4 \cite{openai2023gpt4}, has been marketed as multimodal i.e. it will be able to process inputs from modalities other than language and produce output in those modalities. However, the scope of modalities beyond languages is assumed to be limited to vision only. The Kosmos-1 \cite{huang2023languageKosmos1} and Kosmos-2 \cite{peng2023kosmos2} are recent advancements in multimodal LLMs capable of showing satisfactory performance in a variety of unimodal and multi-modal tasks. 

    In recent years, several learning paradigms have evolved to train MLLMs with varying architectures \cite{yin2023survey}. Initially, MLLMs relied on the pre-training and fine-tuning analogous to the Vision Language Pre-training (VLP) for VQA. Afterward, they relied on prompt engineering and instruction tuning, two widely adopted techniques in modern LLM literature. A set of MLLMs extended GPT-based backbones to visual modalities \cite{zhang2023pmc, liu2023visual, maaz2023video}. The bridge the gap between the modalities LLMs often relied on separate adapters, e.g. Llama Adapter \cite{zhang2023llama, gao2023llama}, used as learnable interfaces.

\subsection{Open Problems}

There are lots of potential areas in VQA that haven't been explored yet. Both the visual and linguistic domains can have variations derived from the generalized problem definition of multimodal question answering. 
In this section, we shall explore various novel problem statements in VQA.


\subsubsection{Non-English VLP Datasets and Models}
Linguistically, VQA has experienced a variety of questions ranging from simple binary questions \cite{zhang2016yin} to complex reasoning-based questions \cite{johnson2017clevr, zhang2019raven}. However, most of these models are natively trained on English texts as there is a lack of non-English VLP datasets. An interesting VQA setting that is often overlooked is the cross-lingual setting \cite{pfeiffer2021xgqa}. However, there is a lack of standardized work on VQA for languages except English and Chinese. There is potential for creating non-English VLP and VQA datasets.

\subsubsection{Visual Robustness Evaluation}
VQA has been widely adopted in different domains but lacks significant work in domain evaluation. Few works exist \cite{gupta2022grit} on the overall generalization and vision-language robustness of VQA models. Integration of unimodal and multimodal LLMs in VQA systems also introduced new problems like hallucinations \cite{li2023evaluatingHallucination} for which there is no VQA framework. Although VQA is not restricted to generative AI, object hallucination has been a common phenomenon in vision-language classifiers \cite{rohrbach2018objectHallucination}. Evaluation of VQA models has always been a challenging task and we still have a class of evaluation-related open problems that aims for developing reliable VQA systems.

\subsubsection{Multi-Image Counting QA}
While the traditional single image and single question setting has been widely adopted in many domains, other settings like image-pair and image-set question answering saw little to no work \cite{bansal2020visualImageSet, yuan2022change}. Counting-based questions as seen in \cite{acharya2019tallyqa} are limited to a single image while multi-image counting is still an open problem. Fig-\ref{fig:imageCountingChallenge} proposes the task of Image Set QA specified at counting-based questions only.

\subsubsection{Green Computing in VQA}
Environmental sustainability is a key issue that is addressed by green computing \cite{kurp2008green}. \citet{ahmad2021artificial} evaluated AI systems in sustainable energy industries while advocating for a greener AI system. Generative AI and LLMs integrated in VQA models are at the frontier of modern AI systems. Researchers have yet to establish significant works for more efficient and greener VQA systems that can ensure sustainability in the long run.


\subsection{What's next for VQA?}
\label{sec:futureWork}
The future of Visual Question Answering (VQA) holds significant promise as researchers and developers continue to innovate in this domain. The integration of VQA into real-world applications, such as healthcare, autonomous vehicles, and e-commerce, will become more prevalent, demonstrating its practical utility. The incorporation of generative models, such as GPT-4 \cite{openai2023gpt4}, into VQA systems, will also contribute to improved performance and adaptability. Furthermore, addressing challenges related to biases, fairness, and interpretability in VQA will be crucial for ethical and responsible deployment. Apart from the enhanced performance, VQA can expect to see broader applicability, and a growing emphasis on ethical considerations, which collectively hold the potential to revolutionize the way humans and machines interact with visual information.

\section{Conclusions}
\label{sec:Conclusion}
Our work went through the datasets and methods sketched in the setting of traditional VQA surveys and then delved deeper into the modern techniques in the context of vision language pre-training (VLP). However, we couldn't expand the discussion on two emerging subdomains of VQA - Zero Shot VQA (ZS-VQA) and VideoQA. Both of these domains are popular fields of interest for current researchers and should be prioritized by future researchers as well. Additional discussion on the intricacies of transformer-based architecture should also be beneficial in introducing architectural novelty. Nevertheless, we believe that the provided directions will be fruitful in sculpting the domain of VQA in the coming years.









\bibliographystyle{elsarticle-num-names} 
\bibliography{ref.bib}

\end{document}

%% file: TabSurveySpec.tex
\begin{table}[ht]
    \label{tab:surveySpecial}
    \centering
    \begin{tabularx}{0.5\textwidth}{lll}
    \toprule
         \textbf{Name} & \textbf{Year} & \textbf{Topic} \\
         \toprule
         \citet{barra2021visual} & 2021 &  Applications in VQA   \\
         \midrule
         \citet{zhang2019information} & 2019 & \multirow{2}{*}{Fusion Techniques}\\
         \citet{lu2023multi} & 2023 & \\
         \midrule
         \citet{yuan2021language} & 2021 & Language Bias \\
         \midrule
         \citet{yusuf2022analysis} & 2022 & Graph Convolutional Networks\\ \midrule
         \citet{zhong2022video} & 2022 & Video QA \\ \midrule
         \citet{lin2023medical} & 2023 & Medical VQA\\
         \midrule
         \citet{kafle2019challenges} & 2019 & \multirow{2}{*}{Vision and Language Research}\\
         \citet{mogadala2021trends} & 2021 & \\ \midrule
         \citet{gan2022visionVLPSurvey} & 2022 & \multirow{2}{*}{Vision Language Pre-training}\\
         \citet{chen2023vlpSurvey} & 2023 \\        
         \midrule
         \citet{baltruvsaitis2018multimodalMachineLearning} & 2018 & Multimodal Machine Learning\\
         \midrule
         \citet{fu2018recent} & 2018 & Zero-shot Recognition \\
         \citet{chen2021knowledge} & 2021 & Zero-shot Learning \\
         \midrule
         \citet{hossain2019comprehensiveImageCaptioning} & 2019 & Image Captioning \\ \bottomrule
    \end{tabularx}
    \caption{Topics covered by prominent specialized surveys related to VQA}
    
\end{table}

%% file: TabSurveyGen.tex
\begin{landscape}
\begin{table}[htpb]
\centering

\begin{tabularx}{1.2685\textwidth}{llm{6.65cm}m{10cm}}
    \toprule
    \textbf{Name} & \textbf{Year} &\textbf{Challenges \& Open Problems} & \textbf{Contributions} \\
    \midrule
    \citet{wu2017visualSurvey} &2017& Question Constraints, Visual and Textual \newline
    Understanding, External Knowledge,\newline
    Preference for Computer Vision-based methods & Comprehensive, the most cited VQA survey \newline
                       Categorization and generalization of early datasets and methods \newline
                       Discusses emerging works like Structure Scene Text Annotation \cite{krishna2017visual}
                       \\ \midrule
    \citet{kafle2017visual} &2017& Method superiority, Dataset Bias, \newline
    Attention in VQA, Open-Ended (OE), \newline
    and Multiple Choice (MC) Evaluation
    & Emphasis on problem formulation related to Vision and Language tasks \newline
    Qualitative comparison of methods and evaluation metrics \newline
    Elaborative discussion on research challenges with recommendations    \\ \midrule
    \citet{gupta2017survey} &2017& Answer Type Prediction Models, \newline
    Hybrid models, Answer Generation
    & 
    Simple and straightforward survey providing an introductory view\newline
    Focuses on a few important datasets and methods
    \\ \midrule
    \citet{teney2017visual} &2017& Dataset Bias, Zero-Shot VQA, External \newline
    Knowledge, Modular Architectures
    & 
    Reviews fundamental techniques with phase-by-phase generalization \newline
    Emphasizes attention-based and memory-augmented architectures \newline
    Highlights advanced methodologies and domain trends
    \\ \midrule
    \citet{hassantabar2018visual} &2018& Complex reasoning, short-term memory \newline
    and counting-based questions 
    & Introductory survey similar to \cite{gupta2017survey} focusing on few datasets and models
    \\ \midrule 
    \citet{manmadhan2020visual} &2020& R-CNN \cite{girshick2014rich} based Image Featurization, \newline
    Out of Vocabulary words, Transformer-based \newline
    Architectures, Sentence-based Embeddings
    & 
    Guide for newcomers expositing fundamental concepts \newline
    Highlights computer vision subtasks to solve VQA \newline
    Phase-wise comparison of methods in different VQA architectures
    \\ \midrule
    \citet{sharma2021survey} &2021& OE and MC Evaluation, Dataset Bias, \newline
    real VQA Image Featurization, \newline
    Conversational, and Scene Text Questions
    & 
    Compares traditional VQA models to scene text VQA models \newline
    Detailed result analysis on 13 prominent VQA datasets \newline
    Introduces a few open challenges in the domain
    \\ \midrule
    \citet{srivastava2021visual} &2021& Incorporating CV and NLP strategies, \newline
    Real-life Datasets  
    & Highlights major breakthroughs in VQA \newline
    Discussions and analysis based on architectural paradigms \\ \bottomrule

\end{tabularx}
\caption{Overview of existing generalized VQA surveys}
\label{tab:surveyGen}
  \end{table}
\end{landscape}

%% file: TabDsTradQuant.tex
\begin{landscape}
\begin{table}[ht]
\centering

\begin{tabular}{|c|c|c|cc|ccc|}
\hline
\multirow{2}{*}{\textbf{Name}} & \multirow{2}{*}{\textbf{Year}} & \multirow{2}{*}{\textbf{\begin{tabular}[c]{@{}c@{}}Task\\ Category\end{tabular}}} & \multicolumn{2}{c|}{\textbf{Number}}                           & \multicolumn{3}{c|}{\textbf{Description}}                                                                                                                                                                                                                                                                                                                                                         \\ \cline{4-8} 
                               &                                &                                                                                   & \multicolumn{1}{c|}{\textbf{\# Images}} & \textbf{\#Questions} & \multicolumn{1}{c|}{\textbf{Image Description}}                                                                    & \multicolumn{1}{c|}{\textbf{Question Description}}                                                                                                     & \textbf{Answer Description}                                                                                         \\ \hline
DAQUAR \cite{malinowski2014multi}                         & 2014                           & \multirow{2}{*}{OE}                                                               & \multicolumn{1}{c|}{1,449}              & 12,468               & \multicolumn{1}{c|}{\begin{tabular}[c]{@{}c@{}}Natural Images from\\ NYU-Depth V2 \cite{silberman2012indoor}\end{tabular}}                    & \multicolumn{1}{c|}{\begin{tabular}[c]{@{}c@{}}Synthetic template-based\\ and human annotated\\ Mostly object related\end{tabular}}                    & \begin{tabular}[c]{@{}c@{}}Limited to 37 or 894\\ answer classes\end{tabular}                                       \\ \cline{1-2} \cline{4-8} 
COCO-QA \cite{ren2015exploring}                      & \multirow{5}{*}{2015}          &                                                                                   & \multicolumn{1}{c|}{117,684}            & 117,684              & \multicolumn{1}{c|}{\multirow{4}{*}{\begin{tabular}[c]{@{}c@{}}Natural Images from\\ MS-COCO \cite{lin2014microsoft}\end{tabular}}}        & \multicolumn{1}{c|}{\begin{tabular}[c]{@{}c@{}}Generated using description \\ to QA algorithm\end{tabular}}                                            & \begin{tabular}[c]{@{}c@{}}One-word, class-based\\ answers\end{tabular}                                             \\ \cline{1-1} \cline{3-5} \cline{7-8} 
Visual Madlibs \cite{yu2015visual}                &                                & \begin{tabular}[c]{@{}c@{}}OE\\ MC\\ FITB\end{tabular}                             & \multicolumn{1}{c|}{10,738}             & 360,001              & \multicolumn{1}{c|}{}                                                                                              & \multicolumn{1}{c|}{\begin{tabular}[c]{@{}c@{}}Automatically generated \\ from COCO image captions \cite{chen2015microsoft}\end{tabular}}                                       & \begin{tabular}[c]{@{}c@{}}Each question is answered \\ by 3 workers on Amazon\\ Mechanical Turk (AMT)\end{tabular} \\ \cline{1-1} \cline{3-5} \cline{7-8} 
FM-IQA \cite{gao2015you}                        &                                & OE                                                                                & \multicolumn{1}{c|}{158,392}            & 316,193              & \multicolumn{1}{c|}{}                                                                                              & \multicolumn{2}{c|}{Human annotations from Baidu's online crowdsourcing}                                                                                                                                                                                                     \\ \cline{1-1} \cline{3-5} \cline{7-8} 
VQA real \cite{antol2015vqa}              &                                & \multirow{2}{*}{\begin{tabular}[c]{@{}c@{}}OE\\ MC\end{tabular}}                  & \multicolumn{1}{c|}{204,721}            & 614,163              & \multicolumn{1}{c|}{}                                                                                              & \multicolumn{1}{c|}{\multirow{2}{*}{\begin{tabular}[c]{@{}c@{}}Human annotations by\\ AMT workers with at least\\ 3 questions per image\end{tabular}}} & \multirow{2}{*}{\begin{tabular}[c]{@{}c@{}}Each question is answered\\ by 10 AMT workers\end{tabular}}              \\ \cline{1-1} \cline{4-6}
VQA abstract \cite{antol2015vqa}            &                                &                                                                                   & \multicolumn{1}{c|}{50,000}             & 150,000              & \multicolumn{1}{c|}{\begin{tabular}[c]{@{}c@{}}Cartoon-like Synthetic \\ Clipart Images \end{tabular}}         & \multicolumn{1}{c|}{}                                                                                                                                  &                                                                                                                     \\ \hline
Visual7W \cite{zhu2016visual7w}                      & 2016                           & \begin{tabular}[c]{@{}c@{}}MC\\ VG\end{tabular}                                   & \multicolumn{1}{c|}{47,300}             & 327,939              & \multicolumn{1}{c|}{\begin{tabular}[c]{@{}c@{}}Natural Images from\\ MS-COCO\end{tabular}}                         & \multicolumn{2}{c|}{\begin{tabular}[c]{@{}c@{}}Subset of Visual Genome QA pairs\footnotemark[1] with additional\\ annotations on VG, MC, etc by AMT workers\end{tabular}}                                                                                                                   \\ \hline
Visual Genome \cite{krishna2017visual}                 & \multirow{3}{*}{2017}                           & \begin{tabular}[c]{@{}c@{}}OE\\ VG\end{tabular}                                   & \multicolumn{1}{c|}{108,077}            & 1,773,258            & \multicolumn{1}{c|}{\begin{tabular}[c]{@{}c@{}}Natural Images from\\ MS-COCO and \\ YFCC100M \cite{thomee2016yfcc100m}\end{tabular}}         & \multicolumn{2}{c|}{\begin{tabular}[c]{@{}c@{}}Free-from and Region-based QA pairs based on \\ six "W"s\footnotemark[2]  annotated by AMT workers\end{tabular}}                                                                                                                                \\ \cline{1-1} \cline{3-8} 
VQA v2 \cite{goyal2017making}               &                            & \begin{tabular}[c]{@{}c@{}}OE\\ MC\end{tabular}                                   & \multicolumn{1}{c|}{204,721}            & 1,105,904            & \multicolumn{1}{c|}{\begin{tabular}[c]{@{}c@{}}Natural Images from\\ MS-COCO\end{tabular}}                         & \multicolumn{2}{c|}{\begin{tabular}[c]{@{}c@{}}Same as VQA with additional QA pairs \\ related to complementary images\end{tabular}}                                                                                                                                         \\ \cline{1-1} \cline{3-8} 
TDIUC \cite{kafle2017analysis}                          &                            & OE                                                                                & \multicolumn{1}{c|}{167,437}            & 1,654,167            & \multicolumn{1}{c|}{\begin{tabular}[c]{@{}c@{}}Natural   Images from \\ MS-COCO and \\ Visual Genome\end{tabular}} & \multicolumn{2}{c|}{\begin{tabular}[c]{@{}c@{}}12 types of questions both automatically and\\ manually generated with 12 volunteers\end{tabular}}                                                                                                                            \\ \hline
\end{tabular}

\caption{
Comparison of task category, images, and QA pairs of traditional VQA datasets. OE - Open Ended, MC - Multiple Choice, FITB - Fill In The Blanks, VG - Visual Grounding}
\label{tab:tradDatasetQuant}
\end{table}
\footnotetext[1]{The Visual Genome is a large-scale crowdsourced project. Although the corresponding paper was published after Visual7W, the Visual Genome dataset had been publicly available prior to its official publication.}
\footnotetext[2]{The 6 "W"s are \emph{what, who, what, why, when, where, how}. The additional "W" introduced in Visual7W is \emph{which}.}
\end{landscape}

%% file: TabDsTradCont.tex
\begin{landscape}
\begin{table}[ht]
\centering

\begin{tabular}{lll}
\toprule
\multirow{2}{*}{\textbf{Name}} & \multirow{2}{*}{\textbf{Contributions}}                                                                                                                                                                                                               & \multirow{2}{*}{\textbf{Limitations}}                                                                                                                                                                                                                                             \\
                               &                                                                                                                                                                                                                                                       &                                                                                                                                                                                                                                                                                   \\ \midrule
DAQUAR \cite{malinowski2014multi}                         & \begin{tabular}[c]{@{}l@{}}First VQA benchmark to attempt Visual Turing Test \cite{malinowski2014towards}\\ Various question categories with extensible templates\end{tabular}                                                                                & \begin{tabular}[c]{@{}l@{}}Insufficient data to train large models\\ Limited to indoor scenes with unfavorable lighting conditions\\ Questions are restricted to templates and answers are limited to classes\\ Complicated model evaluation due to multiple metrics\end{tabular} \\ \midrule
COCO-QA \cite{ren2015exploring}                       & \begin{tabular}[c]{@{}l@{}}Larger dataset with standardized image source \cite{lin2014microsoft}\\ QA algorithm is extensible to other image captioning datasets \cite{ordonez2011im2text, young2014image}\\ Easier evaluation due to formulation as classification problem\end{tabular} & \begin{tabular}[c]{@{}l@{}}Unnatural and grammatically inaccurate questions\\ Limited question diversity\\ Answers are limited to a single word only\end{tabular}                                                                                                                 \\ \midrule
Visual Madlibs \cite{yu2015visual}                & \begin{tabular}[c]{@{}l@{}}Proposes the novel task of fill-in-the-blanks (FITBS) with multiple choices\\ Diversified questions prompts\end{tabular}                                                                                                   & \begin{tabular}[c]{@{}l@{}}Insufficient answers for open-ended evaluation\\ FITBs based on declarative sentences are easily answerable\end{tabular}                                                                                                                               \\ \midrule
FM-IQA  \cite{gao2015you}                       & \begin{tabular}[c]{@{}l@{}}Multilingual Dataset (English and Chinese)\\ Free-form questions with diversified answer choices\\ Rigorous quality assurance for Chinese QA pairs\end{tabular}                                                            & \begin{tabular}[c]{@{}l@{}}Visual turing test-based manual evaluation is unscalable\\ English QA pairs may not be accurate due to automated translation\end{tabular}                                                                                                              \\ \midrule
VQA \cite{antol2015vqa}                   & \begin{tabular}[c]{@{}l@{}}Benchmark for free-form VQA used for evaluating many models\\ Diversified dataset with realistic and synthetic images\\ High answer-to-question ratio with automatic evaluation\end{tabular}                               & \begin{tabular}[c]{@{}l@{}}Unbalanced dataset resulting in questions answerable without images\\ Lack of reasoning-based and complex questions\\ Subjective questions without a single correct answer\end{tabular}                                                                \\ \midrule
Visual7W \cite{zhu2016visual7w}                       & \begin{tabular}[c]{@{}l@{}}Introduces the task of visual grounding\\ QA diversity corresponding to multiple standard vision tasks\end{tabular}                                                                                                        & \begin{tabular}[c]{@{}l@{}}Lacks binary (yes/no) questions\\ Wide performance gap between humans and AI\end{tabular}                                                                                                                                                              \\ \midrule
Visual Genome \cite{krishna2017visual}                 & \begin{tabular}[c]{@{}l@{}}Largest free-form dataset based on QA pairs\\ Diversified QA pairs on multiple image regions\\ Attribute and Relationship-based QA pairs using Scene Graphs\end{tabular}     & \begin{tabular}[c]{@{}l@{}}Difficult to evaluate long answers\\ Inherently too large resulting in preference for its subset \cite{zhu2016visual7w}\end{tabular}                                                                                                                     \\ \hline
VQA v2 \cite{goyal2017making}               & \begin{tabular}[c]{@{}l@{}}Introduces counter-examples to create a balanced dataset\\ Reduction of Language bias as seen in VQA \cite{antol2015vqa}\\ Counter-examples can be used as a modality for model explainability\end{tabular}                                    & \begin{tabular}[c]{@{}l@{}}Lacks questions on general knowledge\\ Insufficient reasoning-based questions especially on synthetic data\\ Question category biases\footnotemark might result in poor real-world performance \cite{kafle2017analysis} \end{tabular}                                    \\ \midrule
TDIUC \cite{kafle2017analysis}                         & \begin{tabular}[c]{@{}l@{}}Wide category of questions with a balanced distribution\\ Absurd/meaningless questions for image-based reasoning\\ Evaluation strategy counters question-category biases\end{tabular}                                      & \begin{tabular}[c]{@{}l@{}}Similar questions belonging to a certain category, especially colors\\ The majority (around 40\%) are binary questions on object presence\\ Manual annotations come from a small sample space\end{tabular}                                             \\ \bottomrule
\end{tabular}
\label{tab:tradDatasetCont}
\caption{Contributions and limitations of traditional VQA datasets}
\end{table}
\footnotetext{An abundance of a certain category of questions like "Is/Are" questions will result in models being trained better in that particular question category.}
\end{landscape}

%% file: TabDsKB.tex
\begin{landscape}
\begin{table}[p]
\centering

\begin{tabular}{|l|c|cc|ccc|l|}
\hline
\multicolumn{1}{|c|}{\textbf{Name}} & \textbf{Year}                              & \multicolumn{1}{c|}{\textbf{\#Img}} & \textbf{\#Ques} & \multicolumn{1}{c|}{\textbf{Image Description}}                                                                                                 & \multicolumn{1}{c|}{\textbf{QA Description}}                                                                                                                        & \textbf{KB/KG}                                                                     & \multicolumn{1}{c|}{\textbf{Contributions}}                                                                                                                                                                                                    \\ \hline
KB-VQA \cite{wang2015explicit}             & 2015                                       & \multicolumn{1}{c|}{700}          & 3-5/img      & \multicolumn{1}{c|}{\begin{tabular}[c]{@{}c@{}}Natural images from MS \\ COCO \cite{lin2014microsoft} val set covering\\ 150 objects and 100 scenes\end{tabular}} & \multicolumn{1}{c|}{\begin{tabular}[c]{@{}c@{}}Generated by 5 human annotators \\ based on 23 templates\end{tabular}}                                               & DBpedia \cite{auer2007dbpedia}  & \begin{tabular}[c]{@{}l@{}}Introduces KB test setting of VQA models\\ Multi-reasoning on image and KB\end{tabular}                                                                                                                        \\ \hline
FVQA \cite{wang2017fvqa}                                & \multirow{2}{*}{2018}                      & \multicolumn{1}{c|}{2190}         & 5826         & \multicolumn{1}{c|}{\begin{tabular}[c]{@{}c@{}}Natural images from MS \\ COCO val set and\\ImageNet \cite{deng2009imagenet} test set\end{tabular}}               & \multicolumn{1}{c|}{\begin{tabular}[c]{@{}c@{}}Collected from 38 annotators \\encompassing 32 types of\\questions\end{tabular}}                                  & \begin{tabular}[c]{@{}c@{}}DBpedia\\ ConceptNet \cite{liu2004conceptnet}\\ WebChild \cite{tandon2014acquiringWebchild}\end{tabular}            & \begin{tabular}[c]{@{}l@{}}Larger dataset with multiple knowledge bases\\ Supporting facts enable answering complex \\ questions that requires deep reasoning\end{tabular}                                                                     \\ \cline{1-1} \cline{3-8} 
R-VQA \cite{lu2018rvqa}                              &                                            & \multicolumn{2}{c|}{335k}                        & \multicolumn{3}{c|}{\begin{tabular}[c]{@{}c@{}}Instances of natural images, QA pairs, and relation facts from Visual Genome \cite{krishna2017visual} \\ Facts are filtered using a ranking algorithm and evaluated by humans\end{tabular}}                                                                                                                                                                                         & \begin{tabular}[c]{@{}l@{}}Ensures retrieval of relevant concepts\\ Utilization of semantic knowledge in images\end{tabular}                                                                                                                   \\ \hline
KVQA \cite{shah2019kvqa}                               & \multirow{2}{*}{2019}                      & \multicolumn{1}{c|}{24k}          & 183k         & \multicolumn{1}{c|}{\begin{tabular}[c]{@{}c@{}}Natural images of 18k people \\ from Wikidata \cite{vrandevcic2014wikidata}\end{tabular}}    & \multicolumn{1}{c|}{\begin{tabular}[c]{@{}c@{}}Templated questions\\ annotated by humans\end{tabular}}                                                              & \begin{tabular}[c]{@{}c@{}}Support\\ set from\\ Wikidata\end{tabular}             & \begin{tabular}[c]{@{}l@{}}Introduces KB-VQA on named entities\\ Reasoning on KGs to answer KB questions\\ Largest KG-based VQA dataset\end{tabular}                                                                                           \\ \cline{1-1} \cline{3-8} 

OK-VQA \cite{marino2019okVQA}                     &                                            & \multicolumn{1}{c|}{14,031}       & 14,055       & \multicolumn{1}{c|}{\begin{tabular}[c]{@{}c@{}}Natural images randomly\\ sampled from MS COCO\end{tabular}}                                     & \multicolumn{1}{c|}{\begin{tabular}[c]{@{}c@{}}Generated by AMT workers\\ with 5 answer labels and then\\ filtered to KB questions only\end{tabular}}     & \multirow{4}{*}{\begin{tabular}[c]{@{}c@{}}None\\ (Open\\ Knowledge)\end{tabular}} & \begin{tabular}[c]{@{}l@{}}Knowledge-base independent setting\\ Large-scale and difficult dataset\\ Diverse knowledge categories\\ Rigorous filtering ensures high-quality QA\end{tabular}                    \\ \cline{1-6} \cline{8-8} 

OK-VQA$_{S3}$ \cite{jain2021select}                          & \multicolumn{1}{l|}{\multirow{2}{*}{2021}} & \multicolumn{2}{c|}{2640}                        & \multicolumn{2}{c|}{Reannotated subset of OK-VQA}                                                                                                                                                                                                                                                           &                                                                                    & \multirow{2}{*}{\begin{tabular}[c]{@{}l@{}}Answerable only by consulting a KB/KG\\ Prevents guessing from answer distribution\\ Enforces information retrieval\\ Naturally explainable approach\end{tabular}} \\ \cline{1-1} \cline{3-6}
S3VQA \cite{jain2021select}                              & \multicolumn{1}{l|}{}                      & \multicolumn{2}{c|}{6765}                        & \multicolumn{1}{c|}{\begin{tabular}[c]{@{}c@{}}Natural images from \\ OpenImages collection \cite{kuznetsova2020open} \end{tabular}}                                      & \multicolumn{1}{c|}{\begin{tabular}[c]{@{}c@{}}Automatically generated on Wiki\\ pages using T5 model \cite{raffel2020exploring} \\ Exactly 1 answer/question\end{tabular}}          &                                                                                    &                                                                                                                                                                                                               \\ \cline{1-6} \cline{8-8} 
A-OKVQA                             & \multirow{2}{*}{2022}                      & \multicolumn{1}{c|}{23,692}       & 24,903       & \multicolumn{1}{c|}{\begin{tabular}[c]{@{}c@{}}Natural images from \\ MS-COCO\end{tabular}}                                                     & \multicolumn{1}{c|}{\begin{tabular}[c]{@{}c@{}}QA and rationale annotated \\ by 437 AMT workers\end{tabular}}                                             &                                                                                    & \begin{tabular}[c]{@{}l@{}}Rationale enhances reasoning, information\\ retrieval and model explainability\end{tabular}                                                                                        \\ \cline{1-1} \cline{3-8} 
ViQuAE                              &                                            & \multicolumn{1}{c|}{3.7k}         & 3.3k         & \multicolumn{1}{c|}{\begin{tabular}[c]{@{}c@{}}Natural images from\\ Wikimedia Commons \cite{commons2012wikimedia}\end{tabular}}                                           & \multicolumn{1}{c|}{\begin{tabular}[c]{@{}c@{}}Automatic annotations on\\ TriviaQA  \cite{joshi2017triviaqa} from KILT \cite{petroni2020kilt} \\ Manually rephrased and filtered\end{tabular}}       & Wikipedia                   & \begin{tabular}[c]{@{}l@{}}Covers a wide range of entities for VQA on\\ named entities extending KVQA\\ Diverse entity types\end{tabular}                                                                     \\ \hline
FVQA 2.0                            & 2023                                       & \multicolumn{2}{c|}{9899*}                       & \multicolumn{1}{c|}{\begin{tabular}[c]{@{}c@{}}Natural images from\\ MS-COCO and ImageNet\end{tabular}}                                          & \multicolumn{1}{c|}{\begin{tabular}[c]{@{}c@{}}Automatically generated using\\ question templates from KG \\ triplets and filtered manually\end{tabular}} & \begin{tabular}[c]{@{}c@{}}Dbpedia\\ ConceptNet\\ WebChild\end{tabular}            & \begin{tabular}[c]{@{}l@{}}Reduces language bias due to question\\ patters and answer distribution\\ Improves robustness through augmentation\end{tabular}                                                    \\ \hline

\end{tabular}
\caption{Knowledge-based VQA datasets: KB - Knowledge Base, KG - Knowledge Graph, AMT - Amazon Mechanical Turk}
\end{table}
\end{landscape}

%% file: TabDsRB.tex
\begin{table*}[ht]

\centering
\begin{tabular}{|l|l|l|l|}
\hline
\textbf{Name}   & \textbf{Year}         & \textbf{Description}                                                                                  & \textbf{Contributions}                                                                                                                                                                                          \\ \hline
SHAPES \cite{andreas2016neural}          & 2016                  & \multirow{3}{*}{\begin{tabular}[c]{@{}l@{}}Reasoning-based \\ synthetic dataset\end{tabular}}        & Evaluate understanding of spatial relations and complex questions                                                                                                                                               \\ \cline{1-2} \cline{4-4} 
CLEVR \cite{johnson2017clevr}  & 2017                  &                                                                                                      & Compositional questions to test complex visual and logical reasoning                                                                                                                                            \\ \cline{1-2} \cline{4-4} 
IconQA \cite{lu2021iconqa}        & 2021                  &                                                                                                      & Incorporates diagram understanding and basic knowledge                                                                                                                                                          \\ \hline
\textbf{VCR}\cite{zellers2019recognitionVCR}  & \multirow{3}{*}{2019} & \begin{tabular}[c]{@{}l@{}}Large-scale reasoning- \\based dataset from \\movie  scenes\end{tabular}  & \begin{tabular}[c]{@{}l@{}}Introduces new task of producing rationale behind an answer\\ Generalized benchmark for evaluation of common-sense reasoning\\ of VQA systems\end{tabular}                           \\ \cline{1-1} \cline{3-4} 
GQA \cite{hudson2019gqa}    &                       & \begin{tabular}[c]{@{}l@{}} Large-scale reasoning-\\based dataset from \\Visual Genome \cite{krishna2017visual} \end{tabular} & \begin{tabular}[c]{@{}l@{}}Controlling the answer distribution to address language bias\\ Multi-step reasoning-based and compositional questions combine \\ bias reduction and reasoning challenges\end{tabular} \\ \cline{1-1} \cline{3-4} 
RAVEN \cite{zhang2019raven}          &                       & \begin{tabular}[c]{@{}l@{}}RPM \cite{raven1938raven} problem-\\ based reasoning dataset\end{tabular}  & \begin{tabular}[c]{@{}l@{}}Prioritizes visual reasoning over visual recognition\\ Addresses compositional reasoning and visual memory\end{tabular}                                     \\ \hline
WHOOPS! \cite{bitton2023breakingWHOOPS}         & 2023                  & \begin{tabular}[c]{@{}l@{}}AI generated reasoning-\\based dataset\end{tabular}                             & Challenges the reasoning on unconventional images                                                                                                                                                               \\ \hline
Yin-Yang \cite{zhang2016yin}       & 2016                  & \begin{tabular}[c]{@{}l@{}}Reannotates VQA\\ abstract \cite{antol2015vqa} for binary\\ classification\end{tabular}       & \begin{tabular}[c]{@{}l@{}}Address language bias allowing to focus on complex semantics\\ and gain a better image understanding\end{tabular}                                                                    \\ \hline
VQA-CP \cite{agrawal2018don} & 2018                  & \multirow{2}{*}{\begin{tabular}[c]{@{}l@{}}Redistribution of VQA\\  datasets \cite{antol2015vqa, goyal2017making}\end{tabular}}          & Reduces textual bias through different answer distribution                                                                                                                                                      \\ \cline{1-2} \cline{4-4}

VQA-CE \cite{dancette2021beyondVQACE}         & \multirow{2}{*}{2021} &                                                                                                      & \begin{tabular}[c]{@{}l@{}}Detection of multimodal shortcuts in VQA datasets\\ Evaluation of robustness based on the aforementioned shortcuts\end{tabular}                                                      \\ \cline{1-1} \cline{3-4} 
ZS-F-VQA \cite{chen2021zero}       &                       & \begin{tabular}[c]{@{}l@{}}Redistribution of F-VQA\\ dataset \cite{wang2017fvqa}\end{tabular}                            & Evaluation of VQA models in Zero-shot settings                                                                                                                                                                  \\ \hline                                                                                                         
HowMany-QA \cite{trott2017interpretableHowManyQA}     & 2018                  & \begin{tabular}[c]{@{}l@{}}Subset of VQA v2 \cite{goyal2017making}\\ and Visual Genome\end{tabular}                         & \begin{tabular}[c]{@{}l@{}}Challenges VQA systems using questions that require counting\\ different objects in an image\end{tabular}                                                                            \\ \hline
TallyQA \cite{acharya2019tallyqa}       & 2019                  & \begin{tabular}[c]{@{}l@{}}Manually annotated and\\ imported dataset\end{tabular}                    & Includes complex and novel counting-based questions                                                                                                                                                             \\ \hline

\end{tabular}
\caption{Reasoning and Bias Reduction Datasets in VQA}
\end{table*}

%% file: TabDsFT.tex
\begin{table*}[htp]

\caption{Figure and Text-based Datasets in VQA}
\vspace{0.25cm}
\centering
\begin{threeparttable}

\begin{tabular}{|l|l|l|l|l|l|}
\hline
\textbf{Name}                                                                      & \textbf{Year}         & \textbf{\#Img} & \textbf{\#Ques} & \textbf{Descrition}                                                                                                                   & \textbf{Contributions}                                                                                                                                                                                              \\ \hline
AI2D \cite{kembhavi2016diagramAI2D}                                                                           & 2016                  & 5k           & 15k          & \begin{tabular}[c]{@{}l@{}}Multiple choice (MC)\\ manually annotated QA \\ on scientific diagrams\\ scrapped from Google\end{tabular} & \begin{tabular}[c]{@{}l@{}}Task involving diagram structure, elements,\\ and relationships using parse graphs\\ Complex visual input with higher levels of\\ element relationship increases task difficulty\end{tabular} \\ \hline
FigureQA   \cite{kahou2017figureqa}                                                                       & \multirow{2}{*}{2017} & 100k         & 1M           & \begin{tabular}[c]{@{}l@{}}Template-based QA on\\ digital charts from 5\\ diagram classes\end{tabular}                                & \begin{tabular}[c]{@{}l@{}}Logical linking between multiple plot elements\\  of different types of graphs\\ Additional data for secondary objectives\tnote{a}\end{tabular}                                                             \\ \cline{1-1} \cline{3-6} 
TQA \cite{kembhavi2017youTextbookVQA}                                                                               &                       & 1k          & 26k          & \begin{tabular}[c]{@{}l@{}}Images from middle \\ school science lessons\\ with MCQs given in the\\  lesson\end{tabular}                & \begin{tabular}[c]{@{}l@{}}Variations in questions based on text, diagrams\\ or both with high difficulty levels\\ Questions with additional textual context \\ evaluate high-level reasoning  \end{tabular}          \\ \hline
DVQA \cite{kafle2018dvqa}                                                                              & 2018                  & 3M           & 3M           & \begin{tabular}[c]{@{}l@{}}Template-based QA on\\ digital bar charts\end{tabular}                                                     & \begin{tabular}[c]{@{}l@{}}Structural understanding, retrieval of data, and\\ reasoning on bar charts\end{tabular}                                                                                                  \\ \hline
TextVQA \cite{singh2019towardsTextVQA}                                                                   & \multirow{3}{*}{2019} & 28k          & 45k          & \multirow{2}{*}{\begin{tabular}[c]{@{}l@{}}Manually annotated,\\ images of texts in real\\ scenes from natural\\ image datasets\end{tabular}} & \begin{tabular}[c]{@{}l@{}}3-stage crowdsourcing pipeline ensuring high\\ quality images with texts\\ Variety in questions for a particular image\end{tabular}                                                           \\ \cline{1-1} \cline{3-4} \cline{6-6} 
ST-VQA \cite{biten2019scene}                                                                  &                       & 23k          & 31.8k        &                                                                                                                                               & \begin{tabular}[c]{@{}l@{}}Addresses language bias due to strong answer\\ dependency on texts in scenes\\ Lower evaluation ambiguity\end{tabular}                                                                        \\ \cline{1-1} \cline{3-6}

OCR-VQA  \cite{mishra2019ocr}                                                                          &                       & 207k         & 1M           & \begin{tabular}[c]{@{}l@{}}Template-based QA\\ on book cover images\end{tabular}                                                   & \begin{tabular}[c]{@{}l@{}}Intersection of Optical Character Recognition\\ (OCR) and VQA tasks\end{tabular}                                                                                            \\ \hline
PlotQA   \cite{methani2020plotqa}                                                                          & 2020                  & 224k         & 28.9M        & \begin{tabular}[c]{@{}l@{}}Semi-automated QA on\\ real-world plots\end{tabular}                                                               & \begin{tabular}[c]{@{}l@{}}Data label variation with real-world QA pairs\\ Questions with out-of-vocabulary words\end{tabular}                                                                                           \\ \hline

Leaf-QA  \cite{chaudhry2020leaf}                                                                          & 2020                  & 250k         & 2M           & \begin{tabular}[c]{@{}l@{}}Digital images of charts \\ generated from public\\ data sources with \\ template-based QA\tnote{b}\end{tabular}   & \begin{tabular}[c]{@{}l@{}}Real-world diagrams on multiple categories\\ Structural and relational QA pairs\\ Out-of-vocabulary answers\\ GRE-based questions increase complexity\end{tabular}                                \\ \hline
DocVQA \cite{mathew2021docvqa}                                                                            & \multirow{2}{*}{2021} & 12k          & 50k          & \begin{tabular}[c]{@{}l@{}}Manually annotated,\\ images of documents\\ from UCSF Industry\\ Documents Library \tnote{1}\end{tabular}           & \begin{tabular}[c]{@{}l@{}}Increased tasks complexity due to the inclu-\\ sion of document elements like tables, charts,\\ forms, etc.\end{tabular}                                                       \\ \cline{1-1} \cline{3-6} 
InfographicVQA \cite{mathew2022infographicvqa}  \cite{mathew2022infographicvqa}                                                                 &                       & 5.4k         & 30k          & \begin{tabular}[c]{@{}l@{}}Digital images of info-\\ graphics scrapped from\\ the internet with manual\\ QA annotations\end{tabular}  & \begin{tabular}[c]{@{}l@{}}Annotation verification to ensure high quality\\ Emphasizes questions requiring basic\\ reasoning and arithmetic\end{tabular}     \\ \hline

ChartQA \cite{masry2022chartqa}          & 2022                  & 21.9k\tnote{c}       & 32.7k\tnote{d}        & \begin{tabular}[c]{@{}l@{}}Hybrid annotation on\\ digital charts crawled \\ from online sources\end{tabular}                                  & \begin{tabular}[c]{@{}l@{}}Reasoning at both visual and logical levels\\ Hybrid annotation allows training flexibility\\ and comparison of annotation quality\end{tabular}                                               \\ \hline

SlideVQA   \cite{tanaka2023slidevqa}                                                                       & 2023                  & 2.6k         & 14.5k        & \begin{tabular}[c]{@{}l@{}}Manual annotation on\\ slide decks of 20 slides\\ from slideshare\tnote{2}\end{tabular}                            & \begin{tabular}[c]{@{}l@{}}Sequential format of slides introduces new \\ challenges for current VQA models\\ Single/multi-hop and numerical reasoning\end{tabular} 

\\ \hline

\end{tabular}

\begin{tablenotes}
\vspace{1cm}
\item[a] Additional data including numerical values used to chart generation, bounding box annotations of plot elements, etc. can be used for objectives like Visual Grounding.
\item[b] Multiple paraphrases of the question templates were generated using Google Translate and one of them was randomly selected. 
\item[c] ChartQA-H (human annotated) has 4.8k images and ChartQA-M (automatically annotated) has 17.1k images.
\item[d] ChartQA-H has 9.6k questions and ChartQA-M has 23.1k questions.
\\
\item[1] https://www.industrydocuments.ucsf.edu/
\item[2] https://www.slideshare.net/
\end{tablenotes}

\end{threeparttable}

\end{table*}

%% file: TabModTrad.tex
\begin{table*}[ht]
\centering

\begin{tabular}{|l|cccc|l|}
\hline
\multirow{2}{*}{\textbf{Name}} & \multicolumn{4}{c|}{\textbf{Architecture}}                                                                                                                                                           & \multicolumn{1}{c|}{\multirow{2}{*}{\textbf{Comments}}} \\ \cline{2-5}
                               & \multicolumn{1}{c|}{\textbf{\begin{tabular}[c]{@{}c@{}}Image\\ Encoder\end{tabular}}} & \multicolumn{1}{c|}{\textbf{\begin{tabular}[c]{@{}c@{}}Text\\ Encoder\end{tabular}}} & \multicolumn{1}{c|}{\textbf{\begin{tabular}[c]{@{}c@{}}Fusion\\ Strategy\end{tabular}}} & \textbf{\begin{tabular}[c]{@{}c@{}}Answer\\ Generator\end{tabular}} & \multicolumn{1}{c|}{}                                   \\ \hline
Neural-Image-QA \cite{malinowski2015ask}     & \multicolumn{1}{c|}{GoogLeNet}                                                        & \multicolumn{3}{c|}{LSTM}                                                                                                                                                                                                                            & First Deep Learning Approach                            \\ \hline
VIS+LSTM \cite{ren2015exploring}              & \multicolumn{1}{c|}{VGGNet}                                                           & \multicolumn{2}{c|}{LSTM}                                                                                                                                                      & Softmax                                                             & COCO-QA Baseline                                        \\ \hline
mQA \cite{gao2015you}                           & \multicolumn{1}{c|}{GoogLeNet}                                                        & \multicolumn{1}{c|}{LSTM}                                                            & \multicolumn{1}{c|}{EWA}                                                                & LSTM, Softmax                                                       & F-VQA Baseline                                          \\ \hline
LSTM Q+I \cite{antol2015vqa}                       & \multicolumn{1}{c|}{VGGNet}                                                           & \multicolumn{1}{c|}{LSTM}                                                            & \multicolumn{1}{c|}{EWM}                                                                & \multirow{4}{*}{Softmax}                                            & VQA Baseline                                            \\ \cline{1-4} \cline{6-6} 
ABC-CNN \cite{chen2015abc}               & \multicolumn{1}{c|}{VGGNet}                                                           & \multicolumn{1}{c|}{LSTM}                                                            & \multicolumn{1}{c|}{EWA}                                                                &                                                                     & Introduces Attention                                    \\ \cline{1-4} \cline{6-6} 
iBOWIMG  \cite{zhou2015simple}                       & \multicolumn{1}{c|}{GoogLeNet}                                                        & \multicolumn{1}{c|}{BoW}                                                             & \multicolumn{1}{c|}{VC}                                                                 &                                                                     &                                                         \\ \cline{1-4} \cline{6-6} 
Full-CNN \cite{ma2016learning}                       & \multicolumn{1}{c|}{VGGNet}                                                           & \multicolumn{2}{c|}{CNN}                                                                                                                                                      &                                                                     &                                                         \\ \hline
LSTM-Att \cite{zhu2016visual7w}                       & \multicolumn{1}{c|}{VGGNet}                                                           & \multicolumn{1}{c|}{LSTM}                                                            & \multicolumn{1}{c|}{EWM}                                                                & LSTM, Softmax                                                       & Visual7W Baseline                                       \\ \hline
SMem-VQA \cite{xu2016ask}              & \multicolumn{1}{c|}{GoogLeNet}                                                        & \multicolumn{1}{c|}{BoW}                                                             & \multicolumn{1}{c|}{EWA}                                                                & \multirow{10}{*}{Softmax}                                                             &                                                         \\ \cline{1-4} \cline{6-6} 
DPPNet \cite{noh2016image}                & \multicolumn{1}{c|}{VGGNet}                                                           & \multicolumn{1}{c|}{GRU}                                                             & \multicolumn{1}{c|}{DPL}                                                               &                                                              & DPL - Dynamic Parameter Layer                                \\ \cline{1-4} \cline{6-6} 
Word + Region \cite{shih2016look}                  & \multicolumn{1}{c|}{VGGNet}                                                           & \multicolumn{1}{c|}{BoW}                                                             & \multicolumn{1}{c|}{VC}                                                                 &                                                              &                                                         \\ \cline{1-4} \cline{6-6} 
SAN \cite{yang2016stacked}                   & \multicolumn{1}{c|}{VGGNet}                                                           & \multicolumn{1}{c|}{CNN/LSTM}                                                        & \multicolumn{1}{c|}{EWA}                                                                &    & Uses multiple attention layers                                                         \\ \cline{1-4} \cline{6-6} 
MRN \cite{kim2016multimodal}                            & \multicolumn{1}{c|}{VGGNet, ResNet}                                                   & \multicolumn{1}{c|}{GRU}                                                             & \multicolumn{1}{c|}{EWA}                                                                &                                                              &                                                         \\ \cline{1-4} \cline{6-6} 
DAN \cite{nam2017dual}                   & \multicolumn{1}{c|}{VGGNet, ResNet}                                                   & \multicolumn{1}{c|}{BLSTM}                                                           & \multicolumn{1}{c|}{EWM}                                                                &                                                              &                                                         \\ \cline{1-4} \cline{6-6} 
MCB \cite{fukui2016multimodalVQAVisualGrounding}                   & \multicolumn{1}{c|}{ResNet}                                                           & \multicolumn{1}{c|}{LSTM}                                                            & \multicolumn{1}{c|}{BL}                                                                 &                                                              & Introduces Bilinear Pooling                             \\ \cline{1-4} \cline{6-6} 
MLB \cite{kim2016hadamard}                   & \multicolumn{1}{c|}{ResNet}                                                           & \multicolumn{1}{c|}{GRU}                                                             & \multicolumn{1}{c|}{BL}                                                                 &                                                              &                                                         \\ \cline{1-4} \cline{6-6} 
HieCoAtt \cite{lu2016hierarchical}              & \multicolumn{1}{c|}{VGGNet, ResNet}                                                   & \multicolumn{1}{c|}{LSTM}                                                            & \multicolumn{1}{c|}{EWA}                                                                &                                                              &      Introduces Co-attention                                                   \\ \cline{1-4} \cline{6-6} 
DMN+ \cite{xiong2016dynamic}                  & \multicolumn{1}{c|}{VGGNet}                                                           & \multicolumn{2}{c|}{BGRU}                                                                                                                                                      &                                                              &                                                         \\ \hline
Attr-CNN+LSTM \cite{wu2017image}                  & \multicolumn{1}{c|}{VGGNet}                                                           & \multicolumn{1}{c|}{LSTM}                                                            & \multicolumn{1}{c|}{LSTM}                                                               & LSTM                                                                &                                                         \\ \hline

MFB \cite{yu2017multi}                            & \multicolumn{1}{c|}{ResNet}                                                           & \multicolumn{1}{c|}{LSTM}                                                            & \multicolumn{1}{c|}{BL}                                                                 & \multirow{4}{*}{Softmax}                                                             &                                                         \\ \cline{1-4} \cline{6-6} 
MLAN \cite{yu2017multi}                           & \multicolumn{1}{c|}{ResNet}                                                           & \multicolumn{1}{c|}{GRU}                                                             & \multicolumn{1}{c|}{EWA}                                                                &                                                              &                                                         \\ \cline{1-4} \cline{6-6} 
MUTAN \cite{ben2017mutan}                          & \multicolumn{1}{c|}{ResNet}                                                           & \multicolumn{1}{c|}{GRU}                                                             & \multicolumn{1}{c|}{BL}                                                                 &                                                              &                                                         \\ \cline{1-4} \cline{6-6} 
SAAA \cite{kazemi2017show}                           & \multicolumn{1}{c|}{ResNet}                                                           & \multicolumn{1}{c|}{LSTM}                                                            & \multicolumn{1}{c|}{VC}                                                                 &                                                              &                                                         \\ \hline
Up-Down \cite{anderson2018bottom}                        & \multicolumn{1}{c|}{FR-CNN, ResNet}                                                   & \multicolumn{1}{c|}{GRU}                                                             & \multicolumn{1}{c|}{VC}                                                                 & Sigmoid                                                             &                                                         \\ \hline
MFH \cite{yu2018beyond}                            & \multicolumn{1}{c|}{ResNet}                                                           & \multicolumn{1}{c|}{LSTM}                                                            & \multicolumn{1}{c|}{BL}                                                                 & Softmax                                                             &                                                         \\ \hline
DCN \cite{nguyen2018improved}                           & \multicolumn{1}{c|}{Resnet}                                                           & \multicolumn{1}{c|}{BLSTM}                                                           & \multicolumn{1}{c|}{VC}                                                                 & Sigmoid                                                             &                                                         \\ \hline
Tips-Trick \cite{teney2018tips}                     & \multicolumn{1}{c|}{FR-CNN, ResNet}                                                   & \multicolumn{1}{c|}{GRU}                                                             & \multicolumn{1}{c|}{EWM}                                                                & Sigmoid  & Sigmoid outperforms Softmax                                                         \\ \hline
BAN \cite{kim2018bilinear}                            & \multicolumn{1}{c|}{FR-CNN}                                                           & \multicolumn{1}{c|}{GRU}                                                             & \multicolumn{1}{c|}{BL}                                                                 & Softmax                                                             &                                                         \\ \hline
MCAN \cite{yu2019deep}                          & \multicolumn{1}{c|}{FR-CNN}                                                           & \multicolumn{1}{c|}{LSTM}                                                            & \multicolumn{1}{c|}{EWA}                                                                & Softmax                                                             &                                                         \\ \hline
\end{tabular}

\caption{Architectural overview of traditional VQA methods preceding the transformer and VLP era. Some of the common techniques incorporated by these methods are Joint Embedding, Attention, Modular Network, and Bilinear Pooling Fusion.}
\label{tab:earlyVLModels}
\end{table*}

%% file: TabResOld.tex
\begin{table*}[p]
\centering

\begin{tabular}{|l|cccc|c|}
\hline
\multirow{2}{*}{\textbf{Model Name}} & \multicolumn{4}{c|}{\textbf{Test-Dev}}                                                                                                    & \textbf{Test-Std} \\ \cline{2-6} 
                                      & \multicolumn{1}{c|}{\textbf{Y\textbackslash{}N}} & \multicolumn{1}{c|}{\textbf{Num}} & \multicolumn{1}{c|}{\textbf{Other}} & \textbf{All} & \textbf{All}      \\ \hline
LSTM Q+I \cite{ren2015exploring}                              & \multicolumn{1}{c|}{80.5}                        & \multicolumn{1}{c|}{36.8}         & \multicolumn{1}{c|}{43}             & 57.8         & 58.2              \\ \hline
SMem {[}2-Hop{]} \cite{xu2016ask}                      & \multicolumn{1}{c|}{80.9}                        & \multicolumn{1}{c|}{37.3}         & \multicolumn{1}{c|}{43.1}           & 58           & 58.2              \\ \hline
SAN \cite{yang2016stacked}                                   & \multicolumn{1}{c|}{79.3}                        & \multicolumn{1}{c|}{36.6}         & \multicolumn{1}{c|}{46.1}           & 58.7         & 58.9              \\ \hline
FDA \cite{ilievski2016focused}                                    & \multicolumn{1}{c|}{81.1}                        & \multicolumn{1}{c|}{36.2}         & \multicolumn{1}{c|}{45.8}           & 59.2         & 59.5              \\ \hline
DMN+ \cite{xiong2016dynamic}                                   & \multicolumn{1}{c|}{80.5}                        & \multicolumn{1}{c|}{36.8}         & \multicolumn{1}{c|}{48.3}           & 60.3         & 60.4              \\ \hline
HierCoAtt \cite{lu2016hierarchical}                              & \multicolumn{1}{c|}{79.7}                        & \multicolumn{1}{c|}{38.7}         & \multicolumn{1}{c|}{51.7}           & 61.8         & 62.1              \\ \hline
DPPnet \cite{noh2016image}                                 & \multicolumn{1}{c|}{80.71}                       & \multicolumn{1}{c|}{37.24}        & \multicolumn{1}{c|}{41.69}          & 57.22        & 57.36             \\ \hline
MRN \cite{kim2016multimodal}                                    & \multicolumn{1}{c|}{82.28}                       & \multicolumn{1}{c|}{38.82}        & \multicolumn{1}{c|}{49.25}          & 61.68        & 61.84             \\ \hline
Deep Q+I \cite{lu2015deeper}                               & \multicolumn{1}{c|}{80.87}                       & \multicolumn{1}{c|}{36.46}        & \multicolumn{1}{c|}{43.4}           & 58.02        & 58.16             \\ \hline
iBOWIMG \cite{zhou2015simple}                                & \multicolumn{1}{c|}{76.5}                        & \multicolumn{1}{c|}{35}           & \multicolumn{1}{c|}{42.6}           & 55.7         & 55.9              \\ \hline
ACK \cite{wu2016ask}                                    & \multicolumn{1}{c|}{79.2}                        & \multicolumn{1}{c|}{36.1}         & \multicolumn{1}{c|}{40.1}           & 55.7         & 56                \\ \hline
MCB {[}Ensemble-7{]} \cite{fukui2016multimodalVQAVisualGrounding}                   & \multicolumn{1}{c|}{83.4}                        & \multicolumn{1}{c|}{39.8}         & \multicolumn{1}{c|}{58.5}           & 66.7         & 66.5              \\ \hline
MLB {[}Ensemble-7{]} \cite{kim2016hadamard}                   & \multicolumn{1}{c|}{84.57}                       & \multicolumn{1}{c|}{39.21}        & \multicolumn{1}{c|}{57.81}          & 66.77        & 66.89             \\ \hline
Dual-MFA \cite{lu2018co}                               & \multicolumn{1}{c|}{83.59}                       & \multicolumn{1}{c|}{40.18}        & \multicolumn{1}{c|}{56.84}          & 66.01        & 66.09             \\ \hline
VQA-Machine \cite{wang2017vqa}                            & \multicolumn{1}{c|}{81.5}                        & \multicolumn{1}{c|}{38.4}         & \multicolumn{1}{c|}{53}             & 63.1         & 63.3              \\ \hline
Neural-Image QA \cite{malinowski2015ask}                        & \multicolumn{1}{c|}{78.4}                        & \multicolumn{1}{c|}{36.4}         & \multicolumn{1}{c|}{46.3}           & 58.4         & 58.4              \\ \hline
NMN \cite{andreas2016neural}                                    & \multicolumn{1}{c|}{81.2}                        & \multicolumn{1}{c|}{38}           & \multicolumn{1}{c|}{44}             & 58.6         & 58.7              \\ \hline
DMN \cite{kumar2016ask}                                    & \multicolumn{1}{c|}{81}                          & \multicolumn{1}{c|}{38.4}         & \multicolumn{1}{c|}{45.2}           & 59.2         & 59.4              \\ \hline
MUTAN {[}Ensemble-5{]} \cite{ben2017mutan}                 & \multicolumn{1}{c|}{85.14}                       & \multicolumn{1}{c|}{39.81}        & \multicolumn{1}{c|}{58.52}          & 67.42        & 67.36             \\ \hline
QGHC \cite{gao2018question}                                   & \multicolumn{1}{c|}{83.54}                       & \multicolumn{1}{c|}{38.06}        & \multicolumn{1}{c|}{57.1}           & 65.89        & 65.9              \\ \hline
D-NMN \cite{andreas2016learning}                                  & \multicolumn{1}{c|}{81.1}                        & \multicolumn{1}{c|}{38.6}         & \multicolumn{1}{c|}{45.5}           & 59.4         & 59.4              \\ \hline
\end{tabular}
\caption{Performance Analysis of traditional VQA models in the pre-transformer era. The models are evaluated on the test-dev and test-std split of the VQA dataset \cite{antol2015vqa}.}
\label{tab:performanceVQATrad}
\end{table*}

%% file: TabResNew.tex
\begin{table}[ht]

\centering
\begin{tabular}{|l|l|l|}
\hline
\textbf{Model Name} & \textbf{Test-Dev} & \textbf{Test-Std} \\ \hline
VisualBERT \cite{li2019visualbert}          & 70.80             & 71.00             \\ \hline
ViLBERT \cite{lu2019vilbert}             & 71.79             & 72.22             \\ \hline
LXMERT \cite{tan2019lxmert}              & 72.42             & 72.54             \\ \hline
OSCAR \cite{li2020oscar}               & 73.16             & 73.44             \\ \hline
UNITER \cite{chen2020uniter}              & 73.82             & 74.02             \\ \hline
PixelBERT \cite{huang2020pixel}           & 74.45             & 74.55             \\ \hline
VILLA \cite{gan2020large}               & 74.69             & 74.87             \\ \hline
UNIMO \cite{li2020unimo}               & 75.06             & 75.27             \\ \hline
ALBEF \cite{li2021alignALBEF}               & 75.84             & 76.04             \\ \hline
VinVL \cite{zhang2021vinvl}               & 76.52             & 76.60             \\ \hline
METER \cite{dou2022empirical}               & 77.68             & 77.64             \\ \hline
BLIP \cite{li2022blip}                & 78.25             & 78.32             \\ \hline
GIT \cite{wang2022git}                 & 78.56             & 78.81             \\ \hline
SimVLM \cite{wang2021simvlm}              & 80.03             & 80.34             \\ \hline
Florence \cite{yuan2021florence}            & 80.16             & 80.36             \\ \hline
mPlug \cite{li2022mplug}               & 81.27             & 81.26             \\ \hline
GIT-2 \cite{wang2022git}               & 81.74             & 81.92             \\ \hline
OFA \cite{wang2022ofa}                 & 82.00             & 82.00             \\ \hline
Flamingo \cite{alayrac2022flamingo}            & 82.00             & 82.10             \\ \hline
CoCa \cite{yu2022coca}                & 82.30             & 82.30             \\ \hline
BLIP-2 \cite{li2023blip2}              & 82.30             & 82.30             \\ \hline
One-Peace \cite{wang2023onePeace}           & 82.60             & 82.50             \\ \hline
VLMO \cite{bao2022vlmo}                & 82.88             & 82.78             \\ \hline
BeiT-3 \cite{wang2022imageBeit}              & 84.20             & 84.00             \\ \hline
PaLI \cite{chen2022pali}                & 84.30             & 84.30             \\ \hline
\end{tabular}
\caption{Performance Analysis of VLP architectures in VQA. The models are evaluated on the test-dev and test-std split of the VQAv2 dataset \cite{goyal2017making}.}
\label{tab:performanceVLP}
\end{table}

%% file: VQA.bbl
\begin{thebibliography}{322}
\expandafter\ifx\csname natexlab\endcsname\relax\def\natexlab#1{#1}\fi
\providecommand{\url}[1]{\texttt{#1}}
\providecommand{\href}[2]{#2}
\providecommand{\path}[1]{#1}
\providecommand{\DOIprefix}{doi:}
\providecommand{\ArXivprefix}{arXiv:}
\providecommand{\URLprefix}{URL: }
\providecommand{\Pubmedprefix}{pmid:}
\providecommand{\doi}[1]{\href{http://dx.doi.org/#1}{\path{#1}}}
\providecommand{\Pubmed}[1]{\href{pmid:#1}{\path{#1}}}
\providecommand{\bibinfo}[2]{#2}
\ifx\xfnm\relax \def\xfnm[#1]{\unskip,\space#1}\fi
\bibitem[{Antol et~al.(2015)Antol, Agrawal, Lu, Mitchell, Batra, Zitnick, and Parikh}]{antol2015vqa}
\bibinfo{author}{S.~Antol}, \bibinfo{author}{A.~Agrawal}, \bibinfo{author}{J.~Lu}, \bibinfo{author}{M.~Mitchell}, \bibinfo{author}{D.~Batra}, \bibinfo{author}{C.~L. Zitnick}, \bibinfo{author}{D.~Parikh},
\newblock \bibinfo{title}{Vqa: Visual question answering},
\newblock in: \bibinfo{booktitle}{Proceedings of the IEEE international conference on computer vision}, \bibinfo{year}{2015}, pp. \bibinfo{pages}{2425--2433}.
\bibitem[{Bansal et~al.(2020)Bansal, Zhang, and Chellappa}]{bansal2020visualImageSet}
\bibinfo{author}{A.~Bansal}, \bibinfo{author}{Y.~Zhang}, \bibinfo{author}{R.~Chellappa},
\newblock \bibinfo{title}{Visual question answering on image sets},
\newblock in: \bibinfo{booktitle}{Computer Vision--ECCV 2020: 16th European Conference, Glasgow, UK, August 23--28, 2020, Proceedings, Part XXI 16}, \bibinfo{organization}{Springer}, \bibinfo{year}{2020}, pp. \bibinfo{pages}{51--67}.
\bibitem[{Xu et~al.(2017)Xu, Zhao, Xiao, Wu, Zhang, He, and Zhuang}]{xu2017videoCaptionQA}
\bibinfo{author}{D.~Xu}, \bibinfo{author}{Z.~Zhao}, \bibinfo{author}{J.~Xiao}, \bibinfo{author}{F.~Wu}, \bibinfo{author}{H.~Zhang}, \bibinfo{author}{X.~He}, \bibinfo{author}{Y.~Zhuang},
\newblock \bibinfo{title}{Video question answering via gradually refined attention over appearance and motion},
\newblock in: \bibinfo{booktitle}{Proceedings of the 25th ACM international conference on Multimedia}, \bibinfo{year}{2017}, pp. \bibinfo{pages}{1645--1653}.
\bibitem[{Zhong et~al.(2022)Zhong, Xiao, Ji, Li, Deng, and Chua}]{zhong2022video}
\bibinfo{author}{Y.~Zhong}, \bibinfo{author}{J.~Xiao}, \bibinfo{author}{W.~Ji}, \bibinfo{author}{Y.~Li}, \bibinfo{author}{W.~Deng}, \bibinfo{author}{T.-S. Chua},
\newblock \bibinfo{title}{Video question answering: Datasets, algorithms and challenges},
\newblock \bibinfo{journal}{arXiv preprint arXiv:2203.01225}  (\bibinfo{year}{2022}).
\bibitem[{Lei et~al.(2018)Lei, Yu, Bansal, and Berg}]{lei2018tvqa}
\bibinfo{author}{J.~Lei}, \bibinfo{author}{L.~Yu}, \bibinfo{author}{M.~Bansal}, \bibinfo{author}{T.~L. Berg},
\newblock \bibinfo{title}{Tvqa: Localized, compositional video question answering},
\newblock \bibinfo{journal}{arXiv preprint arXiv:1809.01696}  (\bibinfo{year}{2018}).
\bibitem[{Mezaris et~al.(2003)Mezaris, Kompatsiaris, and Strintzis}]{mezaris2003ontology}
\bibinfo{author}{V.~Mezaris}, \bibinfo{author}{I.~Kompatsiaris}, \bibinfo{author}{M.~G. Strintzis},
\newblock \bibinfo{title}{An ontology approach to object-based image retrieval},
\newblock in: \bibinfo{booktitle}{Proceedings 2003 International Conference on Image Processing (Cat. No. 03CH37429)}, volume~\bibinfo{volume}{2}, \bibinfo{organization}{IEEE}, \bibinfo{year}{2003}, pp. \bibinfo{pages}{II--511}.
\bibitem[{Zellers et~al.(2019)Zellers, Bisk, Farhadi, and Choi}]{zellers2019recognitionVCR}
\bibinfo{author}{R.~Zellers}, \bibinfo{author}{Y.~Bisk}, \bibinfo{author}{A.~Farhadi}, \bibinfo{author}{Y.~Choi},
\newblock \bibinfo{title}{From recognition to cognition: Visual commonsense reasoning},
\newblock in: \bibinfo{booktitle}{Proceedings of the IEEE/CVF conference on computer vision and pattern recognition}, \bibinfo{year}{2019}, pp. \bibinfo{pages}{6720--6731}.
\bibitem[{Hossain et~al.(2019)Hossain, Sohel, Shiratuddin, and Laga}]{hossain2019comprehensiveImageCaptioning}
\bibinfo{author}{M.~Z. Hossain}, \bibinfo{author}{F.~Sohel}, \bibinfo{author}{M.~F. Shiratuddin}, \bibinfo{author}{H.~Laga},
\newblock \bibinfo{title}{A comprehensive survey of deep learning for image captioning},
\newblock \bibinfo{journal}{ACM Computing Surveys (CsUR)} \bibinfo{volume}{51} (\bibinfo{year}{2019}) \bibinfo{pages}{1--36}.
\bibitem[{Das et~al.(2017)Das, Kottur, Gupta, Singh, Yadav, Moura, Parikh, and Batra}]{das2017visualDialog}
\bibinfo{author}{A.~Das}, \bibinfo{author}{S.~Kottur}, \bibinfo{author}{K.~Gupta}, \bibinfo{author}{A.~Singh}, \bibinfo{author}{D.~Yadav}, \bibinfo{author}{J.~M. Moura}, \bibinfo{author}{D.~Parikh}, \bibinfo{author}{D.~Batra},
\newblock \bibinfo{title}{Visual dialog},
\newblock in: \bibinfo{booktitle}{Proceedings of the IEEE conference on computer vision and pattern recognition}, \bibinfo{year}{2017}, pp. \bibinfo{pages}{326--335}.
\bibitem[{Goyal et~al.(2017)Goyal, Khot, Summers-Stay, Batra, and Parikh}]{goyal2017making}
\bibinfo{author}{Y.~Goyal}, \bibinfo{author}{T.~Khot}, \bibinfo{author}{D.~Summers-Stay}, \bibinfo{author}{D.~Batra}, \bibinfo{author}{D.~Parikh},
\newblock \bibinfo{title}{Making the v in vqa matter: Elevating the role of image understanding in visual question answering},
\newblock in: \bibinfo{booktitle}{Proceedings of the IEEE conference on computer vision and pattern recognition}, \bibinfo{year}{2017}, pp. \bibinfo{pages}{6904--6913}.
\bibitem[{Johnson et~al.(2017)Johnson, Hariharan, Van Der~Maaten, Fei-Fei, Lawrence~Zitnick, and Girshick}]{johnson2017clevr}
\bibinfo{author}{J.~Johnson}, \bibinfo{author}{B.~Hariharan}, \bibinfo{author}{L.~Van Der~Maaten}, \bibinfo{author}{L.~Fei-Fei}, \bibinfo{author}{C.~Lawrence~Zitnick}, \bibinfo{author}{R.~Girshick},
\newblock \bibinfo{title}{Clevr: A diagnostic dataset for compositional language and elementary visual reasoning},
\newblock in: \bibinfo{booktitle}{Proceedings of the IEEE conference on computer vision and pattern recognition}, \bibinfo{year}{2017}, pp. \bibinfo{pages}{2901--2910}.
\bibitem[{Hudson and Manning(2019)}]{hudson2019gqa}
\bibinfo{author}{D.~A. Hudson}, \bibinfo{author}{C.~D. Manning},
\newblock \bibinfo{title}{Gqa: A new dataset for real-world visual reasoning and compositional question answering},
\newblock in: \bibinfo{booktitle}{Proceedings of the IEEE/CVF conference on computer vision and pattern recognition}, \bibinfo{year}{2019}, pp. \bibinfo{pages}{6700--6709}.
\bibitem[{Zhang et~al.(2019)Zhang, Gao, Jia, Zhu, and Zhu}]{zhang2019raven}
\bibinfo{author}{C.~Zhang}, \bibinfo{author}{F.~Gao}, \bibinfo{author}{B.~Jia}, \bibinfo{author}{Y.~Zhu}, \bibinfo{author}{S.-C. Zhu},
\newblock \bibinfo{title}{Raven: A dataset for relational and analogical visual reasoning},
\newblock in: \bibinfo{booktitle}{Proceedings of the IEEE/CVF conference on computer vision and pattern recognition}, \bibinfo{year}{2019}, pp. \bibinfo{pages}{5317--5327}.
\bibitem[{Wang et~al.(2015)Wang, Wu, Shen, Hengel, and Dick}]{wang2015explicit}
\bibinfo{author}{P.~Wang}, \bibinfo{author}{Q.~Wu}, \bibinfo{author}{C.~Shen}, \bibinfo{author}{A.~v.~d. Hengel}, \bibinfo{author}{A.~Dick},
\newblock \bibinfo{title}{Explicit knowledge-based reasoning for visual question answering},
\newblock \bibinfo{journal}{arXiv preprint arXiv:1511.02570}  (\bibinfo{year}{2015}).
\bibitem[{Wang et~al.(2017)Wang, Wu, Shen, Dick, and Van Den~Hengel}]{wang2017fvqa}
\bibinfo{author}{P.~Wang}, \bibinfo{author}{Q.~Wu}, \bibinfo{author}{C.~Shen}, \bibinfo{author}{A.~Dick}, \bibinfo{author}{A.~Van Den~Hengel},
\newblock \bibinfo{title}{Fvqa: Fact-based visual question answering},
\newblock \bibinfo{journal}{IEEE transactions on pattern analysis and machine intelligence} \bibinfo{volume}{40} (\bibinfo{year}{2017}) \bibinfo{pages}{2413--2427}.
\bibitem[{Schwenk et~al.(2022)Schwenk, Khandelwal, Clark, Marino, and Mottaghi}]{schwenk2022okvqa}
\bibinfo{author}{D.~Schwenk}, \bibinfo{author}{A.~Khandelwal}, \bibinfo{author}{C.~Clark}, \bibinfo{author}{K.~Marino}, \bibinfo{author}{R.~Mottaghi},
\newblock \bibinfo{title}{A-okvqa: A benchmark for visual question answering using world knowledge},
\newblock in: \bibinfo{booktitle}{European Conference on Computer Vision}, \bibinfo{organization}{Springer}, \bibinfo{year}{2022}, pp. \bibinfo{pages}{146--162}.
\bibitem[{Methani et~al.(2020)Methani, Ganguly, Khapra, and Kumar}]{methani2020plotqa}
\bibinfo{author}{N.~Methani}, \bibinfo{author}{P.~Ganguly}, \bibinfo{author}{M.~M. Khapra}, \bibinfo{author}{P.~Kumar},
\newblock \bibinfo{title}{Plotqa: Reasoning over scientific plots},
\newblock in: \bibinfo{booktitle}{Proceedings of the IEEE/CVF Winter Conference on Applications of Computer Vision}, \bibinfo{year}{2020}, pp. \bibinfo{pages}{1527--1536}.
\bibitem[{Mishra et~al.(2019)Mishra, Shekhar, Singh, and Chakraborty}]{mishra2019ocr}
\bibinfo{author}{A.~Mishra}, \bibinfo{author}{S.~Shekhar}, \bibinfo{author}{A.~K. Singh}, \bibinfo{author}{A.~Chakraborty},
\newblock \bibinfo{title}{Ocr-vqa: Visual question answering by reading text in images},
\newblock in: \bibinfo{booktitle}{2019 international conference on document analysis and recognition (ICDAR)}, \bibinfo{organization}{IEEE}, \bibinfo{year}{2019}, pp. \bibinfo{pages}{947--952}.
\bibitem[{Biten et~al.(2019)Biten, Tito, Mafla, Gomez, Rusinol, Valveny, Jawahar, and Karatzas}]{biten2019scene}
\bibinfo{author}{A.~F. Biten}, \bibinfo{author}{R.~Tito}, \bibinfo{author}{A.~Mafla}, \bibinfo{author}{L.~Gomez}, \bibinfo{author}{M.~Rusinol}, \bibinfo{author}{E.~Valveny}, \bibinfo{author}{C.~Jawahar}, \bibinfo{author}{D.~Karatzas},
\newblock \bibinfo{title}{Scene text visual question answering},
\newblock in: \bibinfo{booktitle}{Proceedings of the IEEE/CVF international conference on computer vision}, \bibinfo{year}{2019}, pp. \bibinfo{pages}{4291--4301}.
\bibitem[{Lin et~al.(2023)Lin, Zhang, Tao, Shi, Haffari, Wu, He, and Ge}]{lin2023medical}
\bibinfo{author}{Z.~Lin}, \bibinfo{author}{D.~Zhang}, \bibinfo{author}{Q.~Tao}, \bibinfo{author}{D.~Shi}, \bibinfo{author}{G.~Haffari}, \bibinfo{author}{Q.~Wu}, \bibinfo{author}{M.~He}, \bibinfo{author}{Z.~Ge},
\newblock \bibinfo{title}{Medical visual question answering: A survey},
\newblock \bibinfo{journal}{Artificial Intelligence in Medicine}  (\bibinfo{year}{2023}) \bibinfo{pages}{102611}.
\bibitem[{Mathew et~al.(2021)Mathew, Karatzas, and Jawahar}]{mathew2021docvqa}
\bibinfo{author}{M.~Mathew}, \bibinfo{author}{D.~Karatzas}, \bibinfo{author}{C.~Jawahar},
\newblock \bibinfo{title}{Docvqa: A dataset for vqa on document images},
\newblock in: \bibinfo{booktitle}{Proceedings of the IEEE/CVF winter conference on applications of computer vision}, \bibinfo{year}{2021}, pp. \bibinfo{pages}{2200--2209}.
\bibitem[{Masry et~al.(2022)Masry, Long, Tan, Joty, and Hoque}]{masry2022chartqa}
\bibinfo{author}{A.~Masry}, \bibinfo{author}{D.~X. Long}, \bibinfo{author}{J.~Q. Tan}, \bibinfo{author}{S.~Joty}, \bibinfo{author}{E.~Hoque},
\newblock \bibinfo{title}{Chartqa: A benchmark for question answering about charts with visual and logical reasoning},
\newblock \bibinfo{journal}{arXiv preprint arXiv:2203.10244}  (\bibinfo{year}{2022}).
\bibitem[{Mathew et~al.(2022)Mathew, Bagal, Tito, Karatzas, Valveny, and Jawahar}]{mathew2022infographicvqa}
\bibinfo{author}{M.~Mathew}, \bibinfo{author}{V.~Bagal}, \bibinfo{author}{R.~Tito}, \bibinfo{author}{D.~Karatzas}, \bibinfo{author}{E.~Valveny}, \bibinfo{author}{C.~Jawahar},
\newblock \bibinfo{title}{Infographicvqa},
\newblock in: \bibinfo{booktitle}{Proceedings of the IEEE/CVF Winter Conference on Applications of Computer Vision}, \bibinfo{year}{2022}, pp. \bibinfo{pages}{1697--1706}.
\bibitem[{Ren et~al.(2015)Ren, Kiros, and Zemel}]{ren2015exploring}
\bibinfo{author}{M.~Ren}, \bibinfo{author}{R.~Kiros}, \bibinfo{author}{R.~Zemel},
\newblock \bibinfo{title}{Exploring models and data for image question answering},
\newblock \bibinfo{journal}{Advances in neural information processing systems} \bibinfo{volume}{28} (\bibinfo{year}{2015}).
\bibitem[{Malinowski et~al.(2015)Malinowski, Rohrbach, and Fritz}]{malinowski2015ask}
\bibinfo{author}{M.~Malinowski}, \bibinfo{author}{M.~Rohrbach}, \bibinfo{author}{M.~Fritz},
\newblock \bibinfo{title}{Ask your neurons: A neural-based approach to answering questions about images},
\newblock in: \bibinfo{booktitle}{Proceedings of the IEEE international conference on computer vision}, \bibinfo{year}{2015}, pp. \bibinfo{pages}{1--9}.
\bibitem[{Gan et~al.(2022)Gan, Li, Li, Wang, Liu, Gao et~al.}]{gan2022visionVLPSurvey}
\bibinfo{author}{Z.~Gan}, \bibinfo{author}{L.~Li}, \bibinfo{author}{C.~Li}, \bibinfo{author}{L.~Wang}, \bibinfo{author}{Z.~Liu}, \bibinfo{author}{J.~Gao}, et~al.,
\newblock \bibinfo{title}{Vision-language pre-training: Basics, recent advances, and future trends},
\newblock \bibinfo{journal}{Foundations and Trends{\textregistered} in Computer Graphics and Vision} \bibinfo{volume}{14} (\bibinfo{year}{2022}) \bibinfo{pages}{163--352}.
\bibitem[{Chen et~al.(2023)Chen, Zhang, Han, Chen, Shi, Xu, and Xu}]{chen2023vlpSurvey}
\bibinfo{author}{F.-L. Chen}, \bibinfo{author}{D.-Z. Zhang}, \bibinfo{author}{M.-L. Han}, \bibinfo{author}{X.-Y. Chen}, \bibinfo{author}{J.~Shi}, \bibinfo{author}{S.~Xu}, \bibinfo{author}{B.~Xu},
\newblock \bibinfo{title}{Vlp: A survey on vision-language pre-training},
\newblock \bibinfo{journal}{Machine Intelligence Research} \bibinfo{volume}{20} (\bibinfo{year}{2023}) \bibinfo{pages}{38--56}.
\bibitem[{Vaswani et~al.(2017)Vaswani, Shazeer, Parmar, Uszkoreit, Jones, Gomez, Kaiser, and Polosukhin}]{vaswani2017attention}
\bibinfo{author}{A.~Vaswani}, \bibinfo{author}{N.~Shazeer}, \bibinfo{author}{N.~Parmar}, \bibinfo{author}{J.~Uszkoreit}, \bibinfo{author}{L.~Jones}, \bibinfo{author}{A.~N. Gomez}, \bibinfo{author}{{\L}.~Kaiser}, \bibinfo{author}{I.~Polosukhin},
\newblock \bibinfo{title}{Attention is all you need},
\newblock \bibinfo{journal}{Advances in neural information processing systems} \bibinfo{volume}{30} (\bibinfo{year}{2017}).
\bibitem[{Li et~al.(2019)Li, Yatskar, Yin, Hsieh, and Chang}]{li2019visualbert}
\bibinfo{author}{L.~H. Li}, \bibinfo{author}{M.~Yatskar}, \bibinfo{author}{D.~Yin}, \bibinfo{author}{C.-J. Hsieh}, \bibinfo{author}{K.-W. Chang},
\newblock \bibinfo{title}{Visualbert: A simple and performant baseline for vision and language},
\newblock \bibinfo{journal}{arXiv preprint arXiv:1908.03557}  (\bibinfo{year}{2019}).
\bibitem[{Li et~al.(2020{\natexlab{a}})Li, Duan, Fang, Gong, and Jiang}]{li2020unicoder}
\bibinfo{author}{G.~Li}, \bibinfo{author}{N.~Duan}, \bibinfo{author}{Y.~Fang}, \bibinfo{author}{M.~Gong}, \bibinfo{author}{D.~Jiang},
\newblock \bibinfo{title}{Unicoder-vl: A universal encoder for vision and language by cross-modal pre-training},
\newblock in: \bibinfo{booktitle}{Proceedings of the AAAI conference on artificial intelligence}, volume~\bibinfo{volume}{34}, \bibinfo{year}{2020}{\natexlab{a}}, pp. \bibinfo{pages}{11336--11344}.
\bibitem[{Li et~al.(2020{\natexlab{b}})Li, Yin, Li, Zhang, Hu, Zhang, Wang, Hu, Dong, Wei et~al.}]{li2020oscar}
\bibinfo{author}{X.~Li}, \bibinfo{author}{X.~Yin}, \bibinfo{author}{C.~Li}, \bibinfo{author}{P.~Zhang}, \bibinfo{author}{X.~Hu}, \bibinfo{author}{L.~Zhang}, \bibinfo{author}{L.~Wang}, \bibinfo{author}{H.~Hu}, \bibinfo{author}{L.~Dong}, \bibinfo{author}{F.~Wei}, et~al.,
\newblock \bibinfo{title}{Oscar: Object-semantics aligned pre-training for vision-language tasks},
\newblock in: \bibinfo{booktitle}{Computer Vision--ECCV 2020: 16th European Conference, Glasgow, UK, August 23--28, 2020, Proceedings, Part XXX 16}, \bibinfo{organization}{Springer}, \bibinfo{year}{2020}{\natexlab{b}}, pp. \bibinfo{pages}{121--137}.
\bibitem[{Kafle and Kanan(2017)}]{kafle2017visual}
\bibinfo{author}{K.~Kafle}, \bibinfo{author}{C.~Kanan},
\newblock \bibinfo{title}{Visual question answering: Datasets, algorithms, and future challenges},
\newblock \bibinfo{journal}{Computer Vision and Image Understanding} \bibinfo{volume}{163} (\bibinfo{year}{2017}) \bibinfo{pages}{3--20}.
\bibitem[{Wu et~al.(2017)Wu, Teney, Wang, Shen, Dick, and Van Den~Hengel}]{wu2017visualSurvey}
\bibinfo{author}{Q.~Wu}, \bibinfo{author}{D.~Teney}, \bibinfo{author}{P.~Wang}, \bibinfo{author}{C.~Shen}, \bibinfo{author}{A.~Dick}, \bibinfo{author}{A.~Van Den~Hengel},
\newblock \bibinfo{title}{Visual question answering: A survey of methods and datasets},
\newblock \bibinfo{journal}{Computer Vision and Image Understanding} \bibinfo{volume}{163} (\bibinfo{year}{2017}) \bibinfo{pages}{21--40}.
\bibitem[{Bigham et~al.(2010)Bigham, Jayant, Ji, Little, Miller, Miller, Miller, Tatarowicz, White, White et~al.}]{bigham2010vizwiz}
\bibinfo{author}{J.~P. Bigham}, \bibinfo{author}{C.~Jayant}, \bibinfo{author}{H.~Ji}, \bibinfo{author}{G.~Little}, \bibinfo{author}{A.~Miller}, \bibinfo{author}{R.~C. Miller}, \bibinfo{author}{R.~Miller}, \bibinfo{author}{A.~Tatarowicz}, \bibinfo{author}{B.~White}, \bibinfo{author}{S.~White}, et~al.,
\newblock \bibinfo{title}{Vizwiz: nearly real-time answers to visual questions},
\newblock in: \bibinfo{booktitle}{Proceedings of the 23nd annual ACM symposium on User interface software and technology}, \bibinfo{year}{2010}, pp. \bibinfo{pages}{333--342}.
\bibitem[{Barra et~al.(2021)Barra, Bisogni, De~Marsico, and Ricciardi}]{barra2021visual}
\bibinfo{author}{S.~Barra}, \bibinfo{author}{C.~Bisogni}, \bibinfo{author}{M.~De~Marsico}, \bibinfo{author}{S.~Ricciardi},
\newblock \bibinfo{title}{Visual question answering: Which investigated applications?},
\newblock \bibinfo{journal}{Pattern Recognition Letters} \bibinfo{volume}{151} (\bibinfo{year}{2021}) \bibinfo{pages}{325--331}.
\bibitem[{Gurari et~al.(2018)Gurari, Li, Stangl, Guo, Lin, Grauman, Luo, and Bigham}]{gurari2018vizwiz}
\bibinfo{author}{D.~Gurari}, \bibinfo{author}{Q.~Li}, \bibinfo{author}{A.~J. Stangl}, \bibinfo{author}{A.~Guo}, \bibinfo{author}{C.~Lin}, \bibinfo{author}{K.~Grauman}, \bibinfo{author}{J.~Luo}, \bibinfo{author}{J.~P. Bigham},
\newblock \bibinfo{title}{Vizwiz grand challenge: Answering visual questions from blind people},
\newblock in: \bibinfo{booktitle}{Proceedings of the IEEE conference on computer vision and pattern recognition}, \bibinfo{year}{2018}, pp. \bibinfo{pages}{3608--3617}.
\bibitem[{Gurari et~al.(2019)Gurari, Li, Lin, Zhao, Guo, Stangl, and Bigham}]{gurari2019vizwiz}
\bibinfo{author}{D.~Gurari}, \bibinfo{author}{Q.~Li}, \bibinfo{author}{C.~Lin}, \bibinfo{author}{Y.~Zhao}, \bibinfo{author}{A.~Guo}, \bibinfo{author}{A.~Stangl}, \bibinfo{author}{J.~P. Bigham},
\newblock \bibinfo{title}{Vizwiz-priv: A dataset for recognizing the presence and purpose of private visual information in images taken by blind people},
\newblock in: \bibinfo{booktitle}{Proceedings of the IEEE/CVF Conference on Computer Vision and Pattern Recognition}, \bibinfo{year}{2019}, pp. \bibinfo{pages}{939--948}.
\bibitem[{Tseng et~al.(2022)Tseng, Bell, and Gurari}]{tseng2022vizwiz}
\bibinfo{author}{Y.-Y. Tseng}, \bibinfo{author}{A.~Bell}, \bibinfo{author}{D.~Gurari},
\newblock \bibinfo{title}{Vizwiz-fewshot: Locating objects in images taken by people with visual impairments},
\newblock in: \bibinfo{booktitle}{European Conference on Computer Vision}, \bibinfo{organization}{Springer}, \bibinfo{year}{2022}, pp. \bibinfo{pages}{575--591}.
\bibitem[{Burton et~al.(2012)Burton, Brady, Brewer, Neylan, Bigham, and Hurst}]{burton2012crowdsourcing}
\bibinfo{author}{M.~A. Burton}, \bibinfo{author}{E.~Brady}, \bibinfo{author}{R.~Brewer}, \bibinfo{author}{C.~Neylan}, \bibinfo{author}{J.~P. Bigham}, \bibinfo{author}{A.~Hurst},
\newblock \bibinfo{title}{Crowdsourcing subjective fashion advice using vizwiz: challenges and opportunities},
\newblock in: \bibinfo{booktitle}{Proceedings of the 14th international ACM SIGACCESS conference on Computers and accessibility}, \bibinfo{year}{2012}, pp. \bibinfo{pages}{135--142}.
\bibitem[{Brady et~al.(2013)Brady, Morris, Zhong, White, and Bigham}]{brady2013visual}
\bibinfo{author}{E.~Brady}, \bibinfo{author}{M.~R. Morris}, \bibinfo{author}{Y.~Zhong}, \bibinfo{author}{S.~White}, \bibinfo{author}{J.~P. Bigham},
\newblock \bibinfo{title}{Visual challenges in the everyday lives of blind people},
\newblock in: \bibinfo{booktitle}{Proceedings of the SIGCHI conference on human factors in computing systems}, \bibinfo{year}{2013}, pp. \bibinfo{pages}{2117--2126}.
\bibitem[{Lasecki et~al.(2013)Lasecki, Thiha, Zhong, Brady, and Bigham}]{lasecki2013answering}
\bibinfo{author}{W.~S. Lasecki}, \bibinfo{author}{P.~Thiha}, \bibinfo{author}{Y.~Zhong}, \bibinfo{author}{E.~Brady}, \bibinfo{author}{J.~P. Bigham},
\newblock \bibinfo{title}{Answering visual questions with conversational crowd assistants},
\newblock in: \bibinfo{booktitle}{Proceedings of the 15th International ACM SIGACCESS Conference on Computers and Accessibility}, \bibinfo{year}{2013}, pp. \bibinfo{pages}{1--8}.
\bibitem[{Gurari and Grauman(2017)}]{gurari2017crowdverge}
\bibinfo{author}{D.~Gurari}, \bibinfo{author}{K.~Grauman},
\newblock \bibinfo{title}{Crowdverge: Predicting if people will agree on the answer to a visual question},
\newblock in: \bibinfo{booktitle}{Proceedings of the 2017 CHI Conference on Human Factors in Computing Systems}, \bibinfo{year}{2017}, pp. \bibinfo{pages}{3511--3522}.
\bibitem[{OpenAI(2023)}]{openai2023gpt4}
\bibinfo{author}{OpenAI}, \bibinfo{title}{Gpt-4 technical report}, \bibinfo{year}{2023}. \href{http://arxiv.org/abs/2303.08774}{{\tt arXiv:2303.08774}}.
\bibitem[{Gurari et~al.(2020)Gurari, Zhao, Zhang, and Bhattacharya}]{gurari2020captioning}
\bibinfo{author}{D.~Gurari}, \bibinfo{author}{Y.~Zhao}, \bibinfo{author}{M.~Zhang}, \bibinfo{author}{N.~Bhattacharya},
\newblock \bibinfo{title}{Captioning images taken by people who are blind},
\newblock in: \bibinfo{booktitle}{Computer Vision--ECCV 2020: 16th European Conference, Glasgow, UK, August 23--28, 2020, Proceedings, Part XVII 16}, \bibinfo{organization}{Springer}, \bibinfo{year}{2020}, pp. \bibinfo{pages}{417--434}.
\bibitem[{Chen et~al.(2022)Chen, Anjum, and Gurari}]{chen2022grounding}
\bibinfo{author}{C.~Chen}, \bibinfo{author}{S.~Anjum}, \bibinfo{author}{D.~Gurari},
\newblock \bibinfo{title}{Grounding answers for visual questions asked by visually impaired people},
\newblock \bibinfo{journal}{arXiv preprint arXiv:2202.01993}  (\bibinfo{year}{2022}).
\bibitem[{Salyers et~al.(2017)Salyers, Bonfils, Luther, Firmin, White, Adams, and Rollins}]{salyers2017relationshipHealthcareBurnout}
\bibinfo{author}{M.~P. Salyers}, \bibinfo{author}{K.~A. Bonfils}, \bibinfo{author}{L.~Luther}, \bibinfo{author}{R.~L. Firmin}, \bibinfo{author}{D.~A. White}, \bibinfo{author}{E.~L. Adams}, \bibinfo{author}{A.~L. Rollins},
\newblock \bibinfo{title}{The relationship between professional burnout and quality and safety in healthcare: a meta-analysis},
\newblock \bibinfo{journal}{Journal of general internal medicine} \bibinfo{volume}{32} (\bibinfo{year}{2017}) \bibinfo{pages}{475--482}.
\bibitem[{He et~al.(2017)He, Xia, Yu, Jian, Meng, and Chen}]{educationalRobotSystem}
\bibinfo{author}{B.~He}, \bibinfo{author}{M.~Xia}, \bibinfo{author}{X.~Yu}, \bibinfo{author}{P.~Jian}, \bibinfo{author}{H.~Meng}, \bibinfo{author}{Z.~Chen},
\newblock \bibinfo{title}{An educational robot system of visual question answering for preschoolers},
\newblock in: \bibinfo{booktitle}{2017 2nd International Conference on Robotics and Automation Engineering (ICRAE)}, \bibinfo{year}{2017}, pp. \bibinfo{pages}{441--445}. \DOIprefix\doi{10.1109/ICRAE.2017.8291426}.
\bibitem[{Anwar et~al.(2019)Anwar, Bascou, Menekse, and Kardgar}]{anwar2019systematicEduRobotics}
\bibinfo{author}{S.~Anwar}, \bibinfo{author}{N.~A. Bascou}, \bibinfo{author}{M.~Menekse}, \bibinfo{author}{A.~Kardgar},
\newblock \bibinfo{title}{A systematic review of studies on educational robotics},
\newblock \bibinfo{journal}{Journal of Pre-College Engineering Education Research (J-PEER)} \bibinfo{volume}{9} (\bibinfo{year}{2019}) \bibinfo{pages}{2}.
\bibitem[{Sophia and Jacob(2021)}]{eduBot}
\bibinfo{author}{J.~Sophia}, \bibinfo{author}{T.~Jacob},
\newblock \bibinfo{title}{Edubot-a chatbot for education in covid-19 pandemic and vqabot comparison},
\newblock in: \bibinfo{booktitle}{2021 Second International Conference on Electronics and Sustainable Communication Systems (ICESC)}, \bibinfo{year}{2021}, pp. \bibinfo{pages}{1707--1714}. \DOIprefix\doi{10.1109/ICESC51422.2021.9532611}.
\bibitem[{Wu et~al.(2023)Wu, Yin, Qi, Wang, Tang, and Duan}]{wu2023visual}
\bibinfo{author}{C.~Wu}, \bibinfo{author}{S.~Yin}, \bibinfo{author}{W.~Qi}, \bibinfo{author}{X.~Wang}, \bibinfo{author}{Z.~Tang}, \bibinfo{author}{N.~Duan},
\newblock \bibinfo{title}{Visual chatgpt: Talking, drawing and editing with visual foundation models},
\newblock \bibinfo{journal}{arXiv preprint arXiv:2303.04671}  (\bibinfo{year}{2023}).
\bibitem[{Suresh et~al.(2018)Suresh, Nagaraj~Rao, and Srinivasa}]{gamificationVQA}
\bibinfo{author}{S.~Suresh}, \bibinfo{author}{V.~Nagaraj~Rao}, \bibinfo{author}{G.~Srinivasa},
\newblock \bibinfo{title}{Gamification of a visual question answer system},
\newblock in: \bibinfo{booktitle}{2018 IEEE Tenth International Conference on Technology for Education (T4E)}, \bibinfo{year}{2018}, pp. \bibinfo{pages}{41--44}. \DOIprefix\doi{10.1109/T4E.2018.00016}.
\bibitem[{Vedd et~al.(2022)Vedd, Wang, Rei, Miao, and Specia}]{guidingVisualQuestionGeneration}
\bibinfo{author}{N.~Vedd}, \bibinfo{author}{Z.~Wang}, \bibinfo{author}{M.~Rei}, \bibinfo{author}{Y.~Miao}, \bibinfo{author}{L.~Specia},
\newblock \bibinfo{title}{Guiding visual question generation},
\newblock in: \bibinfo{booktitle}{Proceedings of the 2022 Conference of the North American Chapter of the Association for Computational Linguistics: Human Language Technologies}, \bibinfo{publisher}{Association for Computational Linguistics}, \bibinfo{address}{Seattle, United States}, \bibinfo{year}{2022}, pp. \bibinfo{pages}{1640--1654}. \URLprefix \url{https://aclanthology.org/2022.naacl-main.118}. \DOIprefix\doi{10.18653/v1/2022.naacl-main.118}.
\bibitem[{Kembhavi et~al.(2016)Kembhavi, Salvato, Kolve, Seo, Hajishirzi, and Farhadi}]{kembhavi2016diagramAI2D}
\bibinfo{author}{A.~Kembhavi}, \bibinfo{author}{M.~Salvato}, \bibinfo{author}{E.~Kolve}, \bibinfo{author}{M.~Seo}, \bibinfo{author}{H.~Hajishirzi}, \bibinfo{author}{A.~Farhadi},
\newblock \bibinfo{title}{A diagram is worth a dozen images},
\newblock in: \bibinfo{booktitle}{Computer Vision--ECCV 2016: 14th European Conference, Amsterdam, The Netherlands, October 11--14, 2016, Proceedings, Part IV 14}, \bibinfo{organization}{Springer}, \bibinfo{year}{2016}, pp. \bibinfo{pages}{235--251}.
\bibitem[{Bongini et~al.(2020)Bongini, Becattini, Bagdanov, and Del~Bimbo}]{bongini2020visualCultural}
\bibinfo{author}{P.~Bongini}, \bibinfo{author}{F.~Becattini}, \bibinfo{author}{A.~D. Bagdanov}, \bibinfo{author}{A.~Del~Bimbo},
\newblock \bibinfo{title}{Visual question answering for cultural heritage},
\newblock in: \bibinfo{booktitle}{IOP Conference Series: Materials Science and Engineering}, volume \bibinfo{volume}{949}, \bibinfo{organization}{IOP Publishing}, \bibinfo{year}{2020}, p. \bibinfo{pages}{012074}.
\bibitem[{Kembhavi et~al.(2017)Kembhavi, Seo, Schwenk, Choi, Farhadi, and Hajishirzi}]{kembhavi2017youTextbookVQA}
\bibinfo{author}{A.~Kembhavi}, \bibinfo{author}{M.~Seo}, \bibinfo{author}{D.~Schwenk}, \bibinfo{author}{J.~Choi}, \bibinfo{author}{A.~Farhadi}, \bibinfo{author}{H.~Hajishirzi},
\newblock \bibinfo{title}{Are you smarter than a sixth grader? textbook question answering for multimodal machine comprehension},
\newblock in: \bibinfo{booktitle}{Proceedings of the IEEE Conference on Computer Vision and Pattern recognition}, \bibinfo{year}{2017}, pp. \bibinfo{pages}{4999--5007}.
\bibitem[{Ding et~al.(2023)Ding, Luo, Chung, and Han}]{ding2023vqapdf}
\bibinfo{author}{Y.~Ding}, \bibinfo{author}{S.~Luo}, \bibinfo{author}{H.~Chung}, \bibinfo{author}{S.~C. Han},
\newblock \bibinfo{title}{Vqa: A new dataset for real-world vqa on pdf documents},
\newblock \bibinfo{journal}{arXiv preprint arXiv:2304.06447}  (\bibinfo{year}{2023}).
\bibitem[{Tanaka et~al.(2023)Tanaka, Nishida, Nishida, Hasegawa, Saito, and Saito}]{tanaka2023slidevqa}
\bibinfo{author}{R.~Tanaka}, \bibinfo{author}{K.~Nishida}, \bibinfo{author}{K.~Nishida}, \bibinfo{author}{T.~Hasegawa}, \bibinfo{author}{I.~Saito}, \bibinfo{author}{K.~Saito},
\newblock \bibinfo{title}{Slidevqa: A dataset for document visual question answering on multiple images},
\newblock \bibinfo{journal}{arXiv preprint arXiv:2301.04883}  (\bibinfo{year}{2023}).
\bibitem[{Bommasani et~al.(2021)Bommasani, Hudson, Adeli, Altman, Arora, von Arx, Bernstein, Bohg, Bosselut, Brunskill et~al.}]{bommasani2021opportunitiesRisksFoundationalModels}
\bibinfo{author}{R.~Bommasani}, \bibinfo{author}{D.~A. Hudson}, \bibinfo{author}{E.~Adeli}, \bibinfo{author}{R.~Altman}, \bibinfo{author}{S.~Arora}, \bibinfo{author}{S.~von Arx}, \bibinfo{author}{M.~S. Bernstein}, \bibinfo{author}{J.~Bohg}, \bibinfo{author}{A.~Bosselut}, \bibinfo{author}{E.~Brunskill}, et~al.,
\newblock \bibinfo{title}{On the opportunities and risks of foundation models},
\newblock \bibinfo{journal}{arXiv preprint arXiv:2108.07258}  (\bibinfo{year}{2021}).
\bibitem[{Radford et~al.(2018)Radford, Narasimhan, Salimans, Sutskever et~al.}]{radford2018improving}
\bibinfo{author}{A.~Radford}, \bibinfo{author}{K.~Narasimhan}, \bibinfo{author}{T.~Salimans}, \bibinfo{author}{I.~Sutskever}, et~al.,
\newblock \bibinfo{title}{Improving language understanding by generative pre-training},
\newblock \bibinfo{journal}{Technical Report, OpenAI}  (\bibinfo{year}{2018}).
\bibitem[{Brown et~al.(2020)Brown, Mann, Ryder, Subbiah, Kaplan, Dhariwal, Neelakantan, Shyam, Sastry, Askell et~al.}]{brown2020language}
\bibinfo{author}{T.~Brown}, \bibinfo{author}{B.~Mann}, \bibinfo{author}{N.~Ryder}, \bibinfo{author}{M.~Subbiah}, \bibinfo{author}{J.~D. Kaplan}, \bibinfo{author}{P.~Dhariwal}, \bibinfo{author}{A.~Neelakantan}, \bibinfo{author}{P.~Shyam}, \bibinfo{author}{G.~Sastry}, \bibinfo{author}{A.~Askell}, et~al.,
\newblock \bibinfo{title}{Language models are few-shot learners},
\newblock \bibinfo{journal}{Advances in neural information processing systems} \bibinfo{volume}{33} (\bibinfo{year}{2020}) \bibinfo{pages}{1877--1901}.
\bibitem[{Toor et~al.(2019)Toor, Wechsler, and Nappi}]{toor2019biometric}
\bibinfo{author}{A.~S. Toor}, \bibinfo{author}{H.~Wechsler}, \bibinfo{author}{M.~Nappi},
\newblock \bibinfo{title}{Biometric surveillance using visual question answering},
\newblock \bibinfo{journal}{Pattern Recognition Letters} \bibinfo{volume}{126} (\bibinfo{year}{2019}) \bibinfo{pages}{111--118}.
\bibitem[{Sarkar and Rahnemoonfar(2021)}]{sarkar2021vqaAidPostDisasterDamageAssessment}
\bibinfo{author}{A.~Sarkar}, \bibinfo{author}{M.~Rahnemoonfar},
\newblock \bibinfo{title}{Vqa-aid: Visual question answering for post-disaster damage assessment and analysis},
\newblock in: \bibinfo{booktitle}{2021 IEEE International Geoscience and Remote Sensing Symposium IGARSS}, \bibinfo{organization}{IEEE}, \bibinfo{year}{2021}, pp. \bibinfo{pages}{8660--8663}.
\bibitem[{Sarkar et~al.(2023)Sarkar, Chowdhury, Murphy, Gangopadhyay, and Rahnemoonfar}]{sarkar2023samPostDisasterDamageAssessment}
\bibinfo{author}{A.~Sarkar}, \bibinfo{author}{T.~Chowdhury}, \bibinfo{author}{R.~Murphy}, \bibinfo{author}{A.~Gangopadhyay}, \bibinfo{author}{M.~Rahnemoonfar},
\newblock \bibinfo{title}{Sam-vqa: Supervised attention-based visual question answering model for post-disaster damage assessment on remote sensing imagery},
\newblock \bibinfo{journal}{IEEE Transactions on Geoscience and Remote Sensing}  (\bibinfo{year}{2023}).
\bibitem[{Jang et~al.(2017)Jang, Song, Yu, Kim, and Kim}]{jang2017tgif}
\bibinfo{author}{Y.~Jang}, \bibinfo{author}{Y.~Song}, \bibinfo{author}{Y.~Yu}, \bibinfo{author}{Y.~Kim}, \bibinfo{author}{G.~Kim},
\newblock \bibinfo{title}{Tgif-qa: Toward spatio-temporal reasoning in visual question answering},
\newblock in: \bibinfo{booktitle}{Proceedings of the IEEE conference on computer vision and pattern recognition}, \bibinfo{year}{2017}, pp. \bibinfo{pages}{2758--2766}.
\bibitem[{Chou et~al.(2020)Chou, Chao, Lai, Sun, and Yang}]{chou2020visual360deg}
\bibinfo{author}{S.-H. Chou}, \bibinfo{author}{W.-L. Chao}, \bibinfo{author}{W.-S. Lai}, \bibinfo{author}{M.~Sun}, \bibinfo{author}{M.-H. Yang},
\newblock \bibinfo{title}{Visual question answering on 360deg images},
\newblock in: \bibinfo{booktitle}{Proceedings of the IEEE/CVF winter conference on applications of computer vision}, \bibinfo{year}{2020}, pp. \bibinfo{pages}{1607--1616}.
\bibitem[{Das et~al.(2018)Das, Datta, Gkioxari, Lee, Parikh, and Batra}]{das2018embodied}
\bibinfo{author}{A.~Das}, \bibinfo{author}{S.~Datta}, \bibinfo{author}{G.~Gkioxari}, \bibinfo{author}{S.~Lee}, \bibinfo{author}{D.~Parikh}, \bibinfo{author}{D.~Batra},
\newblock \bibinfo{title}{Embodied question answering},
\newblock in: \bibinfo{booktitle}{Proceedings of the IEEE conference on computer vision and pattern recognition}, \bibinfo{year}{2018}, pp. \bibinfo{pages}{1--10}.
\bibitem[{Lin et~al.(2014)Lin, Maire, Belongie, Hays, Perona, Ramanan, Doll{\'a}r, and Zitnick}]{lin2014microsoft}
\bibinfo{author}{T.-Y. Lin}, \bibinfo{author}{M.~Maire}, \bibinfo{author}{S.~Belongie}, \bibinfo{author}{J.~Hays}, \bibinfo{author}{P.~Perona}, \bibinfo{author}{D.~Ramanan}, \bibinfo{author}{P.~Doll{\'a}r}, \bibinfo{author}{C.~L. Zitnick},
\newblock \bibinfo{title}{Microsoft coco: Common objects in context},
\newblock in: \bibinfo{booktitle}{European conference on computer vision}, \bibinfo{organization}{Springer}, \bibinfo{year}{2014}, pp. \bibinfo{pages}{740--755}.
\bibitem[{Xu et~al.(2016)Xu, Mei, Yao, and Rui}]{xu2016msr}
\bibinfo{author}{J.~Xu}, \bibinfo{author}{T.~Mei}, \bibinfo{author}{T.~Yao}, \bibinfo{author}{Y.~Rui},
\newblock \bibinfo{title}{Msr-vtt: A large video description dataset for bridging video and language},
\newblock in: \bibinfo{booktitle}{Proceedings of the IEEE conference on computer vision and pattern recognition}, \bibinfo{year}{2016}, pp. \bibinfo{pages}{5288--5296}.
\bibitem[{Mori et~al.(1999)Mori, Nishida, and Yamada}]{mori1999opticalOCR}
\bibinfo{author}{S.~Mori}, \bibinfo{author}{H.~Nishida}, \bibinfo{author}{H.~Yamada}, \bibinfo{title}{Optical character recognition}, \bibinfo{publisher}{John Wiley \& Sons, Inc.}, \bibinfo{year}{1999}.
\bibitem[{Suhr et~al.(2017)Suhr, Lewis, Yeh, and Artzi}]{suhr2017corpus}
\bibinfo{author}{A.~Suhr}, \bibinfo{author}{M.~Lewis}, \bibinfo{author}{J.~Yeh}, \bibinfo{author}{Y.~Artzi},
\newblock \bibinfo{title}{A corpus of natural language for visual reasoning},
\newblock in: \bibinfo{booktitle}{Proceedings of the 55th Annual Meeting of the Association for Computational Linguistics (Volume 2: Short Papers)}, \bibinfo{year}{2017}, pp. \bibinfo{pages}{217--223}.
\bibitem[{Shrestha et~al.(2020)Shrestha, Kafle, and Kanan}]{shrestha2020negativeLinguisticBias}
\bibinfo{author}{R.~Shrestha}, \bibinfo{author}{K.~Kafle}, \bibinfo{author}{C.~Kanan},
\newblock \bibinfo{title}{A negative case analysis of visual grounding methods for vqa},
\newblock \bibinfo{journal}{arXiv preprint arXiv:2004.05704}  (\bibinfo{year}{2020}).
\bibitem[{Hirota et~al.(2022)Hirota, Nakashima, and Garcia}]{hirota2022genderRacialBiasVQA}
\bibinfo{author}{Y.~Hirota}, \bibinfo{author}{Y.~Nakashima}, \bibinfo{author}{N.~Garcia},
\newblock \bibinfo{title}{Gender and racial bias in visual question answering datasets},
\newblock in: \bibinfo{booktitle}{Proceedings of the 2022 ACM Conference on Fairness, Accountability, and Transparency}, \bibinfo{year}{2022}, pp. \bibinfo{pages}{1280--1292}.
\bibitem[{Acharya et~al.(2019)Acharya, Kafle, and Kanan}]{acharya2019tallyqa}
\bibinfo{author}{M.~Acharya}, \bibinfo{author}{K.~Kafle}, \bibinfo{author}{C.~Kanan},
\newblock \bibinfo{title}{Tallyqa: Answering complex counting questions},
\newblock in: \bibinfo{booktitle}{Proceedings of the AAAI conference on artificial intelligence}, volume \bibinfo{volume}{33-01}, \bibinfo{year}{2019}, pp. \bibinfo{pages}{8076--8084}.
\bibitem[{Yuan et~al.(2022)Yuan, Mou, Xiong, and Zhu}]{yuan2022change}
\bibinfo{author}{Z.~Yuan}, \bibinfo{author}{L.~Mou}, \bibinfo{author}{Z.~Xiong}, \bibinfo{author}{X.~X. Zhu},
\newblock \bibinfo{title}{Change detection meets visual question answering},
\newblock \bibinfo{journal}{IEEE Transactions on Geoscience and Remote Sensing} \bibinfo{volume}{60} (\bibinfo{year}{2022}) \bibinfo{pages}{1--13}.
\bibitem[{Guo et~al.(2023)Guo, Li, Li, Tiong, Li, Tao, and Hoi}]{guo2023imagesFrozenZeroShot}
\bibinfo{author}{J.~Guo}, \bibinfo{author}{J.~Li}, \bibinfo{author}{D.~Li}, \bibinfo{author}{A.~M.~H. Tiong}, \bibinfo{author}{B.~Li}, \bibinfo{author}{D.~Tao}, \bibinfo{author}{S.~Hoi},
\newblock \bibinfo{title}{From images to textual prompts: Zero-shot visual question answering with frozen large language models},
\newblock in: \bibinfo{booktitle}{Proceedings of the IEEE/CVF Conference on Computer Vision and Pattern Recognition}, \bibinfo{year}{2023}, pp. \bibinfo{pages}{10867--10877}.
\bibitem[{Baltru{\v{s}}aitis et~al.(2018)Baltru{\v{s}}aitis, Ahuja, and Morency}]{baltruvsaitis2018multimodalMachineLearning}
\bibinfo{author}{T.~Baltru{\v{s}}aitis}, \bibinfo{author}{C.~Ahuja}, \bibinfo{author}{L.-P. Morency},
\newblock \bibinfo{title}{Multimodal machine learning: A survey and taxonomy},
\newblock \bibinfo{journal}{IEEE transactions on pattern analysis and machine intelligence} \bibinfo{volume}{41} (\bibinfo{year}{2018}) \bibinfo{pages}{423--443}.
\bibitem[{Zhang et~al.(2019)Zhang, Cao, and Wu}]{zhang2019information}
\bibinfo{author}{D.~Zhang}, \bibinfo{author}{R.~Cao}, \bibinfo{author}{S.~Wu},
\newblock \bibinfo{title}{Information fusion in visual question answering: A survey},
\newblock \bibinfo{journal}{Information Fusion} \bibinfo{volume}{52} (\bibinfo{year}{2019}) \bibinfo{pages}{268--280}.
\bibitem[{Lu et~al.(2023)Lu, Liu, Yin, Yin, Liu, and Zheng}]{lu2023multi}
\bibinfo{author}{S.~Lu}, \bibinfo{author}{M.~Liu}, \bibinfo{author}{L.~Yin}, \bibinfo{author}{Z.~Yin}, \bibinfo{author}{X.~Liu}, \bibinfo{author}{W.~Zheng},
\newblock \bibinfo{title}{The multi-modal fusion in visual question answering: a review of attention mechanisms},
\newblock \bibinfo{journal}{PeerJ Computer Science} \bibinfo{volume}{9} (\bibinfo{year}{2023}) \bibinfo{pages}{e1400}.
\bibitem[{Kafle et~al.(2019)Kafle, Shrestha, and Kanan}]{kafle2019challenges}
\bibinfo{author}{K.~Kafle}, \bibinfo{author}{R.~Shrestha}, \bibinfo{author}{C.~Kanan},
\newblock \bibinfo{title}{Challenges and prospects in vision and language research},
\newblock \bibinfo{journal}{Frontiers in Artificial Intelligence} \bibinfo{volume}{2} (\bibinfo{year}{2019}) \bibinfo{pages}{28}.
\bibitem[{Yuan(2021)}]{yuan2021language}
\bibinfo{author}{D.~Yuan},
\newblock \bibinfo{title}{Language bias in visual question answering: A survey and taxonomy},
\newblock \bibinfo{journal}{arXiv preprint arXiv:2111.08531}  (\bibinfo{year}{2021}).
\bibitem[{Yusuf et~al.(2022)Yusuf, Chong, and Xianling}]{yusuf2022analysis}
\bibinfo{author}{A.~A. Yusuf}, \bibinfo{author}{F.~Chong}, \bibinfo{author}{M.~Xianling},
\newblock \bibinfo{title}{An analysis of graph convolutional networks and recent datasets for visual question answering},
\newblock \bibinfo{journal}{Artificial Intelligence Review} \bibinfo{volume}{55} (\bibinfo{year}{2022}) \bibinfo{pages}{6277--6300}.
\bibitem[{Mogadala et~al.(2021)Mogadala, Kalimuthu, and Klakow}]{mogadala2021trends}
\bibinfo{author}{A.~Mogadala}, \bibinfo{author}{M.~Kalimuthu}, \bibinfo{author}{D.~Klakow},
\newblock \bibinfo{title}{Trends in integration of vision and language research: A survey of tasks, datasets, and methods},
\newblock \bibinfo{journal}{Journal of Artificial Intelligence Research} \bibinfo{volume}{71} (\bibinfo{year}{2021}) \bibinfo{pages}{1183--1317}.
\bibitem[{Fu et~al.(2018)Fu, Xiang, Jiang, Xue, Sigal, and Gong}]{fu2018recent}
\bibinfo{author}{Y.~Fu}, \bibinfo{author}{T.~Xiang}, \bibinfo{author}{Y.-G. Jiang}, \bibinfo{author}{X.~Xue}, \bibinfo{author}{L.~Sigal}, \bibinfo{author}{S.~Gong},
\newblock \bibinfo{title}{Recent advances in zero-shot recognition: Toward data-efficient understanding of visual content},
\newblock \bibinfo{journal}{IEEE Signal Processing Magazine} \bibinfo{volume}{35} (\bibinfo{year}{2018}) \bibinfo{pages}{112--125}.
\bibitem[{Chen et~al.(2021)Chen, Geng, Chen, Horrocks, Pan, and Chen}]{chen2021knowledge}
\bibinfo{author}{J.~Chen}, \bibinfo{author}{Y.~Geng}, \bibinfo{author}{Z.~Chen}, \bibinfo{author}{I.~Horrocks}, \bibinfo{author}{J.~Z. Pan}, \bibinfo{author}{H.~Chen},
\newblock \bibinfo{title}{Knowledge-aware zero-shot learning: Survey and perspective},
\newblock \bibinfo{journal}{arXiv preprint arXiv:2103.00070}  (\bibinfo{year}{2021}).
\bibitem[{Krishna et~al.(2017)Krishna, Zhu, Groth, Johnson, Hata, Kravitz, Chen, Kalantidis, Li, Shamma et~al.}]{krishna2017visual}
\bibinfo{author}{R.~Krishna}, \bibinfo{author}{Y.~Zhu}, \bibinfo{author}{O.~Groth}, \bibinfo{author}{J.~Johnson}, \bibinfo{author}{K.~Hata}, \bibinfo{author}{J.~Kravitz}, \bibinfo{author}{S.~Chen}, \bibinfo{author}{Y.~Kalantidis}, \bibinfo{author}{L.-J. Li}, \bibinfo{author}{D.~A. Shamma}, et~al.,
\newblock \bibinfo{title}{Visual genome: Connecting language and vision using crowdsourced dense image annotations},
\newblock \bibinfo{journal}{International journal of computer vision} \bibinfo{volume}{123} (\bibinfo{year}{2017}) \bibinfo{pages}{32--73}.
\bibitem[{Gupta(2017)}]{gupta2017survey}
\bibinfo{author}{A.~K. Gupta},
\newblock \bibinfo{title}{Survey of visual question answering: Datasets and techniques},
\newblock \bibinfo{journal}{arXiv preprint arXiv:1705.03865}  (\bibinfo{year}{2017}).
\bibitem[{Teney et~al.(2017)Teney, Wu, and van~den Hengel}]{teney2017visual}
\bibinfo{author}{D.~Teney}, \bibinfo{author}{Q.~Wu}, \bibinfo{author}{A.~van~den Hengel},
\newblock \bibinfo{title}{Visual question answering: A tutorial},
\newblock \bibinfo{journal}{IEEE Signal Processing Magazine} \bibinfo{volume}{34} (\bibinfo{year}{2017}) \bibinfo{pages}{63--75}.
\bibitem[{Hassantabar(2018)}]{hassantabar2018visual}
\bibinfo{author}{S.~Hassantabar}, \bibinfo{title}{Visual question answering: Datasets, methods, challenges and oppurtunities}, \bibinfo{year}{2018}.
\bibitem[{Manmadhan and Kovoor(2020)}]{manmadhan2020visual}
\bibinfo{author}{S.~Manmadhan}, \bibinfo{author}{B.~C. Kovoor},
\newblock \bibinfo{title}{Visual question answering: a state-of-the-art review},
\newblock \bibinfo{journal}{Artificial Intelligence Review} \bibinfo{volume}{53} (\bibinfo{year}{2020}) \bibinfo{pages}{5705--5745}.
\bibitem[{Girshick et~al.(2014)Girshick, Donahue, Darrell, and Malik}]{girshick2014rich}
\bibinfo{author}{R.~Girshick}, \bibinfo{author}{J.~Donahue}, \bibinfo{author}{T.~Darrell}, \bibinfo{author}{J.~Malik},
\newblock \bibinfo{title}{Rich feature hierarchies for accurate object detection and semantic segmentation},
\newblock in: \bibinfo{booktitle}{Proceedings of the IEEE conference on computer vision and pattern recognition}, \bibinfo{year}{2014}, pp. \bibinfo{pages}{580--587}.
\bibitem[{Sharma and Jalal(2021)}]{sharma2021survey}
\bibinfo{author}{H.~Sharma}, \bibinfo{author}{A.~S. Jalal},
\newblock \bibinfo{title}{A survey of methods, datasets and evaluation metrics for visual question answering},
\newblock \bibinfo{journal}{Image and Vision Computing} \bibinfo{volume}{116} (\bibinfo{year}{2021}) \bibinfo{pages}{104327}.
\bibitem[{Srivastava et~al.(2021)Srivastava, Murali, Dubey, and Mukherjee}]{srivastava2021visual}
\bibinfo{author}{Y.~Srivastava}, \bibinfo{author}{V.~Murali}, \bibinfo{author}{S.~R. Dubey}, \bibinfo{author}{S.~Mukherjee},
\newblock \bibinfo{title}{Visual question answering using deep learning: A survey and performance analysis},
\newblock in: \bibinfo{booktitle}{Computer Vision and Image Processing: 5th International Conference, CVIP 2020, Prayagraj, India, December 4-6, 2020, Revised Selected Papers, Part II 5}, \bibinfo{organization}{Springer}, \bibinfo{year}{2021}, pp. \bibinfo{pages}{75--86}.
\bibitem[{Malinowski and Fritz(2014)}]{malinowski2014multi}
\bibinfo{author}{M.~Malinowski}, \bibinfo{author}{M.~Fritz},
\newblock \bibinfo{title}{A multi-world approach to question answering about real-world scenes based on uncertain input},
\newblock \bibinfo{journal}{Advances in neural information processing systems} \bibinfo{volume}{27} (\bibinfo{year}{2014}).
\bibitem[{Pandhre and Sodhani(2017)}]{pandhre2017shapes}
\bibinfo{author}{S.~Pandhre}, \bibinfo{author}{S.~Sodhani},
\newblock \bibinfo{title}{Survey of recent advances in visual question answering},
\newblock \bibinfo{journal}{arXiv preprint arXiv:1709.08203}  (\bibinfo{year}{2017}).
\bibitem[{Agrawal et~al.(2018)Agrawal, Batra, Parikh, and Kembhavi}]{agrawal2018don}
\bibinfo{author}{A.~Agrawal}, \bibinfo{author}{D.~Batra}, \bibinfo{author}{D.~Parikh}, \bibinfo{author}{A.~Kembhavi},
\newblock \bibinfo{title}{Don't just assume; look and answer: Overcoming priors for visual question answering},
\newblock in: \bibinfo{booktitle}{Proceedings of the IEEE conference on computer vision and pattern recognition}, \bibinfo{year}{2018}, pp. \bibinfo{pages}{4971--4980}.
\bibitem[{Teney and Hengel(2016)}]{teney2016zero}
\bibinfo{author}{D.~Teney}, \bibinfo{author}{A.~v.~d. Hengel},
\newblock \bibinfo{title}{Zero-shot visual question answering},
\newblock \bibinfo{journal}{arXiv preprint arXiv:1611.05546}  (\bibinfo{year}{2016}).
\bibitem[{Hasan et~al.(2018)Hasan, Ling, Farri, Liu, M{\"u}ller, and Lungren}]{hasan2018overview}
\bibinfo{author}{S.~A. Hasan}, \bibinfo{author}{Y.~Ling}, \bibinfo{author}{O.~Farri}, \bibinfo{author}{J.~Liu}, \bibinfo{author}{H.~M{\"u}ller}, \bibinfo{author}{M.~Lungren},
\newblock \bibinfo{title}{Overview of imageclef 2018 medical domain visual question answering task},
\newblock \bibinfo{journal}{Proceedings of CLEF 2018 Working Notes}  (\bibinfo{year}{2018}).
\bibitem[{Malinowski and Fritz(2014)}]{malinowski2014towards}
\bibinfo{author}{M.~Malinowski}, \bibinfo{author}{M.~Fritz},
\newblock \bibinfo{title}{Towards a visual turing challenge},
\newblock \bibinfo{journal}{arXiv preprint arXiv:1410.8027}  (\bibinfo{year}{2014}).
\bibitem[{Chen et~al.(2015)Chen, Fang, Lin, Vedantam, Gupta, Doll{\'a}r, and Zitnick}]{chen2015microsoft}
\bibinfo{author}{X.~Chen}, \bibinfo{author}{H.~Fang}, \bibinfo{author}{T.-Y. Lin}, \bibinfo{author}{R.~Vedantam}, \bibinfo{author}{S.~Gupta}, \bibinfo{author}{P.~Doll{\'a}r}, \bibinfo{author}{C.~L. Zitnick},
\newblock \bibinfo{title}{Microsoft coco captions: Data collection and evaluation server},
\newblock \bibinfo{journal}{arXiv preprint arXiv:1504.00325}  (\bibinfo{year}{2015}).
\bibitem[{Zhang et~al.(2016)Zhang, Goyal, Summers-Stay, Batra, and Parikh}]{zhang2016yin}
\bibinfo{author}{P.~Zhang}, \bibinfo{author}{Y.~Goyal}, \bibinfo{author}{D.~Summers-Stay}, \bibinfo{author}{D.~Batra}, \bibinfo{author}{D.~Parikh},
\newblock \bibinfo{title}{Yin and yang: Balancing and answering binary visual questions},
\newblock in: \bibinfo{booktitle}{Proceedings of the IEEE conference on computer vision and pattern recognition}, \bibinfo{year}{2016}, pp. \bibinfo{pages}{5014--5022}.
\bibitem[{Zhu et~al.(2016)Zhu, Groth, Bernstein, and Fei-Fei}]{zhu2016visual7w}
\bibinfo{author}{Y.~Zhu}, \bibinfo{author}{O.~Groth}, \bibinfo{author}{M.~Bernstein}, \bibinfo{author}{L.~Fei-Fei},
\newblock \bibinfo{title}{Visual7w: Grounded question answering in images},
\newblock in: \bibinfo{booktitle}{Proceedings of the IEEE conference on computer vision and pattern recognition}, \bibinfo{year}{2016}, pp. \bibinfo{pages}{4995--5004}.
\bibitem[{Deng et~al.(2018)Deng, Wu, Wu, Hu, Lyu, and Tan}]{deng2018visualGrounding}
\bibinfo{author}{C.~Deng}, \bibinfo{author}{Q.~Wu}, \bibinfo{author}{Q.~Wu}, \bibinfo{author}{F.~Hu}, \bibinfo{author}{F.~Lyu}, \bibinfo{author}{M.~Tan},
\newblock \bibinfo{title}{Visual grounding via accumulated attention},
\newblock in: \bibinfo{booktitle}{Proceedings of the IEEE conference on computer vision and pattern recognition}, \bibinfo{year}{2018}, pp. \bibinfo{pages}{7746--7755}.
\bibitem[{Kuhn and Neveu(2003)}]{kuhn2003political}
\bibinfo{author}{R.~Kuhn}, \bibinfo{author}{E.~Neveu}, \bibinfo{title}{Political journalism: New challenges, new practices}, \bibinfo{publisher}{Routledge}, \bibinfo{year}{2003}.
\bibitem[{Yu et~al.(2015)Yu, Park, Berg, and Berg}]{yu2015visual}
\bibinfo{author}{L.~Yu}, \bibinfo{author}{E.~Park}, \bibinfo{author}{A.~C. Berg}, \bibinfo{author}{T.~L. Berg},
\newblock \bibinfo{title}{Visual madlibs: Fill in the blank image generation and question answering},
\newblock \bibinfo{journal}{arXiv preprint arXiv:1506.00278}  (\bibinfo{year}{2015}).
\bibitem[{Gao et~al.(2015)Gao, Mao, Zhou, Huang, Wang, and Xu}]{gao2015you}
\bibinfo{author}{H.~Gao}, \bibinfo{author}{J.~Mao}, \bibinfo{author}{J.~Zhou}, \bibinfo{author}{Z.~Huang}, \bibinfo{author}{L.~Wang}, \bibinfo{author}{W.~Xu},
\newblock \bibinfo{title}{Are you talking to a machine? dataset and methods for multilingual image question},
\newblock \bibinfo{journal}{Advances in neural information processing systems} \bibinfo{volume}{28} (\bibinfo{year}{2015}).
\bibitem[{Rafi et~al.(2022)Rafi, Islam, Labib, Hasan, Shah, and Ahmed}]{rafi2022deepBanglaVQA}
\bibinfo{author}{M.~H. Rafi}, \bibinfo{author}{S.~Islam}, \bibinfo{author}{S.~H.~I. Labib}, \bibinfo{author}{S.~S. Hasan}, \bibinfo{author}{F.~M. Shah}, \bibinfo{author}{S.~Ahmed},
\newblock \bibinfo{title}{A deep learning-based bengali visual question answering system},
\newblock in: \bibinfo{booktitle}{2022 25th International Conference on Computer and Information Technology (ICCIT)}, \bibinfo{organization}{IEEE}, \bibinfo{year}{2022}, pp. \bibinfo{pages}{114--119}.
\bibitem[{Chandrasekar et~al.(2022)Chandrasekar, Shimpi, and Naik}]{chandrasekar2022indicHindiVQA}
\bibinfo{author}{A.~Chandrasekar}, \bibinfo{author}{A.~Shimpi}, \bibinfo{author}{D.~Naik},
\newblock \bibinfo{title}{Indic visual question answering},
\newblock in: \bibinfo{booktitle}{2022 IEEE International Conference on Signal Processing and Communications (SPCOM)}, \bibinfo{organization}{IEEE}, \bibinfo{year}{2022}, pp. \bibinfo{pages}{1--5}.
\bibitem[{kamel et~al.(2023)kamel, Hassan, and Elrefaei}]{kamel2023vaqa}
\bibinfo{author}{S.~M. kamel}, \bibinfo{author}{S.~I. Hassan}, \bibinfo{author}{L.~Elrefaei},
\newblock \bibinfo{title}{Vaqa: Visual arabic question answering},
\newblock \bibinfo{journal}{Arabian Journal for Science and Engineering}  (\bibinfo{year}{2023}) \bibinfo{pages}{1--21}.
\bibitem[{Kafle and Kanan(2017)}]{kafle2017analysis}
\bibinfo{author}{K.~Kafle}, \bibinfo{author}{C.~Kanan},
\newblock \bibinfo{title}{An analysis of visual question answering algorithms},
\newblock in: \bibinfo{booktitle}{Proceedings of the IEEE international conference on computer vision}, \bibinfo{year}{2017}, pp. \bibinfo{pages}{1965--1973}.
\bibitem[{Marino et~al.(2019)Marino, Rastegari, Farhadi, and Mottaghi}]{marino2019okVQA}
\bibinfo{author}{K.~Marino}, \bibinfo{author}{M.~Rastegari}, \bibinfo{author}{A.~Farhadi}, \bibinfo{author}{R.~Mottaghi},
\newblock \bibinfo{title}{Ok-vqa: A visual question answering benchmark requiring external knowledge},
\newblock in: \bibinfo{booktitle}{Proceedings of the IEEE/cvf conference on computer vision and pattern recognition}, \bibinfo{year}{2019}, pp. \bibinfo{pages}{3195--3204}.
\bibitem[{Tiong et~al.(2022)Tiong, Li, Li, Savarese, and Hoi}]{tiong2022plug}
\bibinfo{author}{A.~M.~H. Tiong}, \bibinfo{author}{J.~Li}, \bibinfo{author}{B.~Li}, \bibinfo{author}{S.~Savarese}, \bibinfo{author}{S.~C. Hoi},
\newblock \bibinfo{title}{Plug-and-play vqa: Zero-shot vqa by conjoining large pretrained models with zero training},
\newblock \bibinfo{journal}{arXiv preprint arXiv:2210.08773}  (\bibinfo{year}{2022}).
\bibitem[{Peng et~al.(2023)Peng, Wang, Dong, Hao, Huang, Ma, and Wei}]{peng2023kosmos2}
\bibinfo{author}{Z.~Peng}, \bibinfo{author}{W.~Wang}, \bibinfo{author}{L.~Dong}, \bibinfo{author}{Y.~Hao}, \bibinfo{author}{S.~Huang}, \bibinfo{author}{S.~Ma}, \bibinfo{author}{F.~Wei},
\newblock \bibinfo{title}{Kosmos-2: Grounding multimodal large language models to the world},
\newblock \bibinfo{journal}{arXiv preprint arXiv:2306.14824}  (\bibinfo{year}{2023}).
\bibitem[{Auer et~al.(2007)Auer, Bizer, Kobilarov, Lehmann, Cyganiak, and Ives}]{auer2007dbpedia}
\bibinfo{author}{S.~Auer}, \bibinfo{author}{C.~Bizer}, \bibinfo{author}{G.~Kobilarov}, \bibinfo{author}{J.~Lehmann}, \bibinfo{author}{R.~Cyganiak}, \bibinfo{author}{Z.~Ives},
\newblock \bibinfo{title}{Dbpedia: A nucleus for a web of open data},
\newblock in: \bibinfo{booktitle}{international semantic web conference}, \bibinfo{organization}{Springer}, \bibinfo{year}{2007}, pp. \bibinfo{pages}{722--735}.
\bibitem[{Tandon et~al.(2014)Tandon, Melo, and Weikum}]{tandon2014acquiringWebchild}
\bibinfo{author}{N.~Tandon}, \bibinfo{author}{G.~Melo}, \bibinfo{author}{G.~Weikum},
\newblock \bibinfo{title}{Acquiring comparative commonsense knowledge from the web},
\newblock in: \bibinfo{booktitle}{Proceedings of the AAAI Conference on Artificial Intelligence}, volume~\bibinfo{volume}{28}, \bibinfo{year}{2014}, pp. \bibinfo{pages}{154–--162}.
\bibitem[{Liu and Singh(2004)}]{liu2004conceptnet}
\bibinfo{author}{H.~Liu}, \bibinfo{author}{P.~Singh},
\newblock \bibinfo{title}{Conceptnet—a practical commonsense reasoning tool-kit},
\newblock \bibinfo{journal}{BT technology journal} \bibinfo{volume}{22} (\bibinfo{year}{2004}) \bibinfo{pages}{211--226}.
\bibitem[{Lu et~al.(2018)Lu, Ji, Zhang, Duan, Zhou, and Wang}]{lu2018rvqa}
\bibinfo{author}{P.~Lu}, \bibinfo{author}{L.~Ji}, \bibinfo{author}{W.~Zhang}, \bibinfo{author}{N.~Duan}, \bibinfo{author}{M.~Zhou}, \bibinfo{author}{J.~Wang},
\newblock \bibinfo{title}{R-vqa: learning visual relation facts with semantic attention for visual question answering},
\newblock in: \bibinfo{booktitle}{Proceedings of the 24th ACM SIGKDD International Conference on Knowledge Discovery \& Data Mining}, \bibinfo{year}{2018}, pp. \bibinfo{pages}{1880--1889}.
\bibitem[{Lin et~al.(2023)Lin, Wang, and Byrne}]{lin2023fvqa2}
\bibinfo{author}{W.~Lin}, \bibinfo{author}{Z.~Wang}, \bibinfo{author}{B.~Byrne},
\newblock \bibinfo{title}{Fvqa 2.0: Introducing adversarial samples into fact-based visual question answering},
\newblock \bibinfo{journal}{arXiv preprint arXiv:2303.10699}  (\bibinfo{year}{2023}).
\bibitem[{Jain et~al.(2021)Jain, Kothyari, Kumar, Jyothi, Ramakrishnan, and Chakrabarti}]{jain2021select}
\bibinfo{author}{A.~Jain}, \bibinfo{author}{M.~Kothyari}, \bibinfo{author}{V.~Kumar}, \bibinfo{author}{P.~Jyothi}, \bibinfo{author}{G.~Ramakrishnan}, \bibinfo{author}{S.~Chakrabarti},
\newblock \bibinfo{title}{Select, substitute, search: A new benchmark for knowledge-augmented visual question answering},
\newblock in: \bibinfo{booktitle}{Proceedings of the 44th International ACM SIGIR Conference on Research and Development in Information Retrieval}, \bibinfo{year}{2021}, pp. \bibinfo{pages}{2491--2498}.
\bibitem[{Silberman et~al.(2012)Silberman, Hoiem, Kohli, and Fergus}]{silberman2012indoor}
\bibinfo{author}{N.~Silberman}, \bibinfo{author}{D.~Hoiem}, \bibinfo{author}{P.~Kohli}, \bibinfo{author}{R.~Fergus},
\newblock \bibinfo{title}{Indoor segmentation and support inference from rgbd images},
\newblock in: \bibinfo{booktitle}{European conference on computer vision}, \bibinfo{organization}{Springer}, \bibinfo{year}{2012}, pp. \bibinfo{pages}{746--760}.
\bibitem[{Thomee et~al.(2016)Thomee, Shamma, Friedland, Elizalde, Ni, Poland, Borth, and Li}]{thomee2016yfcc100m}
\bibinfo{author}{B.~Thomee}, \bibinfo{author}{D.~A. Shamma}, \bibinfo{author}{G.~Friedland}, \bibinfo{author}{B.~Elizalde}, \bibinfo{author}{K.~Ni}, \bibinfo{author}{D.~Poland}, \bibinfo{author}{D.~Borth}, \bibinfo{author}{L.-J. Li},
\newblock \bibinfo{title}{Yfcc100m: The new data in multimedia research},
\newblock \bibinfo{journal}{Communications of the ACM} \bibinfo{volume}{59} (\bibinfo{year}{2016}) \bibinfo{pages}{64--73}.
\bibitem[{Ordonez et~al.(2011)Ordonez, Kulkarni, and Berg}]{ordonez2011im2text}
\bibinfo{author}{V.~Ordonez}, \bibinfo{author}{G.~Kulkarni}, \bibinfo{author}{T.~Berg},
\newblock \bibinfo{title}{Im2text: Describing images using 1 million captioned photographs},
\newblock \bibinfo{journal}{Advances in neural information processing systems} \bibinfo{volume}{24} (\bibinfo{year}{2011}).
\bibitem[{Young et~al.(2014)Young, Lai, Hodosh, and Hockenmaier}]{young2014image}
\bibinfo{author}{P.~Young}, \bibinfo{author}{A.~Lai}, \bibinfo{author}{M.~Hodosh}, \bibinfo{author}{J.~Hockenmaier},
\newblock \bibinfo{title}{From image descriptions to visual denotations: New similarity metrics for semantic inference over event descriptions},
\newblock \bibinfo{journal}{Transactions of the Association for Computational Linguistics} \bibinfo{volume}{2} (\bibinfo{year}{2014}) \bibinfo{pages}{67--78}.
\bibitem[{Li et~al.(2022)Li, Li, Xiong, and Hoi}]{li2022blip}
\bibinfo{author}{J.~Li}, \bibinfo{author}{D.~Li}, \bibinfo{author}{C.~Xiong}, \bibinfo{author}{S.~Hoi},
\newblock \bibinfo{title}{Blip: Bootstrapping language-image pre-training for unified vision-language understanding and generation},
\newblock in: \bibinfo{booktitle}{International Conference on Machine Learning}, \bibinfo{organization}{PMLR}, \bibinfo{year}{2022}, pp. \bibinfo{pages}{12888--12900}.
\bibitem[{Chen et~al.(2022)Chen, Wang, Changpinyo, Piergiovanni, Padlewski, Salz, Goodman, Grycner, Mustafa, Beyer et~al.}]{chen2022pali}
\bibinfo{author}{X.~Chen}, \bibinfo{author}{X.~Wang}, \bibinfo{author}{S.~Changpinyo}, \bibinfo{author}{A.~Piergiovanni}, \bibinfo{author}{P.~Padlewski}, \bibinfo{author}{D.~Salz}, \bibinfo{author}{S.~Goodman}, \bibinfo{author}{A.~Grycner}, \bibinfo{author}{B.~Mustafa}, \bibinfo{author}{L.~Beyer}, et~al.,
\newblock \bibinfo{title}{Pali: A jointly-scaled multilingual language-image model},
\newblock \bibinfo{journal}{arXiv preprint arXiv:2209.06794}  (\bibinfo{year}{2022}).
\bibitem[{Song et~al.(2022)Song, Dong, Zhang, Liu, and Wei}]{song2022clipFewShot}
\bibinfo{author}{H.~Song}, \bibinfo{author}{L.~Dong}, \bibinfo{author}{W.-N. Zhang}, \bibinfo{author}{T.~Liu}, \bibinfo{author}{F.~Wei},
\newblock \bibinfo{title}{Clip models are few-shot learners: Empirical studies on vqa and visual entailment},
\newblock \bibinfo{journal}{arXiv preprint arXiv:2203.07190}  (\bibinfo{year}{2022}).
\bibitem[{Alayrac et~al.(2022)Alayrac, Donahue, Luc, Miech, Barr, Hasson, Lenc, Mensch, Millican, Reynolds et~al.}]{alayrac2022flamingo}
\bibinfo{author}{J.-B. Alayrac}, \bibinfo{author}{J.~Donahue}, \bibinfo{author}{P.~Luc}, \bibinfo{author}{A.~Miech}, \bibinfo{author}{I.~Barr}, \bibinfo{author}{Y.~Hasson}, \bibinfo{author}{K.~Lenc}, \bibinfo{author}{A.~Mensch}, \bibinfo{author}{K.~Millican}, \bibinfo{author}{M.~Reynolds}, et~al.,
\newblock \bibinfo{title}{Flamingo: a visual language model for few-shot learning},
\newblock \bibinfo{journal}{Advances in Neural Information Processing Systems} \bibinfo{volume}{35} (\bibinfo{year}{2022}) \bibinfo{pages}{23716--23736}.
\bibitem[{Huang et~al.(2023)Huang, Dong, Wang, Hao, Singhal, Ma, Lv, Cui, Mohammed, Liu et~al.}]{huang2023languageKosmos1}
\bibinfo{author}{S.~Huang}, \bibinfo{author}{L.~Dong}, \bibinfo{author}{W.~Wang}, \bibinfo{author}{Y.~Hao}, \bibinfo{author}{S.~Singhal}, \bibinfo{author}{S.~Ma}, \bibinfo{author}{T.~Lv}, \bibinfo{author}{L.~Cui}, \bibinfo{author}{O.~K. Mohammed}, \bibinfo{author}{Q.~Liu}, et~al.,
\newblock \bibinfo{title}{Language is not all you need: Aligning perception with language models},
\newblock \bibinfo{journal}{arXiv preprint arXiv:2302.14045}  (\bibinfo{year}{2023}).
\bibitem[{Shah et~al.(2019)Shah, Mishra, Yadati, and Talukdar}]{shah2019kvqa}
\bibinfo{author}{S.~Shah}, \bibinfo{author}{A.~Mishra}, \bibinfo{author}{N.~Yadati}, \bibinfo{author}{P.~P. Talukdar},
\newblock \bibinfo{title}{Kvqa: Knowledge-aware visual question answering},
\newblock in: \bibinfo{booktitle}{Proceedings of the AAAI Conference on Artificial Intelligence}, volume \bibinfo{volume}{33-01}, \bibinfo{year}{2019}, pp. \bibinfo{pages}{8876--8884}.
\bibitem[{Lerner et~al.(2022)Lerner, Ferret, Guinaudeau, Le~Borgne, Besan{\c{c}}on, Moreno, and Lov{\'o}n~Melgarejo}]{lerner2022viquae}
\bibinfo{author}{P.~Lerner}, \bibinfo{author}{O.~Ferret}, \bibinfo{author}{C.~Guinaudeau}, \bibinfo{author}{H.~Le~Borgne}, \bibinfo{author}{R.~Besan{\c{c}}on}, \bibinfo{author}{J.~G. Moreno}, \bibinfo{author}{J.~Lov{\'o}n~Melgarejo},
\newblock \bibinfo{title}{Viquae, a dataset for knowledge-based visual question answering about named entities},
\newblock in: \bibinfo{booktitle}{Proceedings of the 45th International ACM SIGIR Conference on Research and Development in Information Retrieval}, \bibinfo{year}{2022}, pp. \bibinfo{pages}{3108--3120}.
\bibitem[{Vrande{\v{c}}i{\'c} and Kr{\"o}tzsch(2014)}]{vrandevcic2014wikidata}
\bibinfo{author}{D.~Vrande{\v{c}}i{\'c}}, \bibinfo{author}{M.~Kr{\"o}tzsch},
\newblock \bibinfo{title}{Wikidata: a free collaborative knowledgebase},
\newblock \bibinfo{journal}{Communications of the ACM} \bibinfo{volume}{57} (\bibinfo{year}{2014}) \bibinfo{pages}{78--85}.
\bibitem[{Lu et~al.(2021)Lu, Qiu, Chen, Xia, Zhao, Zhang, Yu, Liang, and Zhu}]{lu2021iconqa}
\bibinfo{author}{P.~Lu}, \bibinfo{author}{L.~Qiu}, \bibinfo{author}{J.~Chen}, \bibinfo{author}{T.~Xia}, \bibinfo{author}{Y.~Zhao}, \bibinfo{author}{W.~Zhang}, \bibinfo{author}{Z.~Yu}, \bibinfo{author}{X.~Liang}, \bibinfo{author}{S.-C. Zhu},
\newblock \bibinfo{title}{Iconqa: A new benchmark for abstract diagram understanding and visual language reasoning},
\newblock \bibinfo{journal}{arXiv preprint arXiv:2110.13214}  (\bibinfo{year}{2021}).
\bibitem[{Dancette et~al.(2021)Dancette, Cadene, Teney, and Cord}]{dancette2021beyondVQACE}
\bibinfo{author}{C.~Dancette}, \bibinfo{author}{R.~Cadene}, \bibinfo{author}{D.~Teney}, \bibinfo{author}{M.~Cord},
\newblock \bibinfo{title}{Beyond question-based biases: Assessing multimodal shortcut learning in visual question answering},
\newblock in: \bibinfo{booktitle}{Proceedings of the IEEE/CVF International Conference on Computer Vision}, \bibinfo{year}{2021}, pp. \bibinfo{pages}{1574--1583}.
\bibitem[{Ma et~al.(2023)Ma, Wang, Kong, Wang, Liu, Pei, and Zhao}]{ma2023robustVQASurvey}
\bibinfo{author}{J.~Ma}, \bibinfo{author}{P.~Wang}, \bibinfo{author}{D.~Kong}, \bibinfo{author}{Z.~Wang}, \bibinfo{author}{J.~Liu}, \bibinfo{author}{H.~Pei}, \bibinfo{author}{J.~Zhao},
\newblock \bibinfo{title}{Robust visual question answering: Datasets, methods, and future challenges},
\newblock \bibinfo{journal}{arXiv preprint arXiv:2307.11471}  (\bibinfo{year}{2023}).
\bibitem[{Gao et~al.(2022)Gao, Wang, Shan, and Chen}]{gao2022cric}
\bibinfo{author}{D.~Gao}, \bibinfo{author}{R.~Wang}, \bibinfo{author}{S.~Shan}, \bibinfo{author}{X.~Chen},
\newblock \bibinfo{title}{Cric: A vqa dataset for compositional reasoning on vision and commonsense},
\newblock \bibinfo{journal}{IEEE Transactions on Pattern Analysis and Machine Intelligence} \bibinfo{volume}{45} (\bibinfo{year}{2022}) \bibinfo{pages}{5561--5578}.
\bibitem[{Andreas et~al.(2016)Andreas, Rohrbach, Darrell, and Klein}]{andreas2016neural}
\bibinfo{author}{J.~Andreas}, \bibinfo{author}{M.~Rohrbach}, \bibinfo{author}{T.~Darrell}, \bibinfo{author}{D.~Klein},
\newblock \bibinfo{title}{Neural module networks},
\newblock in: \bibinfo{booktitle}{Proceedings of the IEEE conference on computer vision and pattern recognition}, \bibinfo{year}{2016}, pp. \bibinfo{pages}{39--48}.
\bibitem[{Liu et~al.(2019)Liu, Liu, Bai, and Yuille}]{liu2019clevrRef}
\bibinfo{author}{R.~Liu}, \bibinfo{author}{C.~Liu}, \bibinfo{author}{Y.~Bai}, \bibinfo{author}{A.~L. Yuille},
\newblock \bibinfo{title}{Clevr-ref+: Diagnosing visual reasoning with referring expressions},
\newblock in: \bibinfo{booktitle}{Proceedings of the IEEE/CVF conference on computer vision and pattern recognition}, \bibinfo{year}{2019}, pp. \bibinfo{pages}{4185--4194}.
\bibitem[{Kottur et~al.(2019)Kottur, Moura, Parikh, Batra, and Rohrbach}]{kottur2019clevrDialog}
\bibinfo{author}{S.~Kottur}, \bibinfo{author}{J.~M. Moura}, \bibinfo{author}{D.~Parikh}, \bibinfo{author}{D.~Batra}, \bibinfo{author}{M.~Rohrbach},
\newblock \bibinfo{title}{Clevr-dialog: A diagnostic dataset for multi-round reasoning in visual dialog},
\newblock \bibinfo{journal}{arXiv preprint arXiv:1903.03166}  (\bibinfo{year}{2019}).
\bibitem[{Arras et~al.(2022)Arras, Osman, and Samek}]{arras2022clevrXAI}
\bibinfo{author}{L.~Arras}, \bibinfo{author}{A.~Osman}, \bibinfo{author}{W.~Samek},
\newblock \bibinfo{title}{Clevr-xai: A benchmark dataset for the ground truth evaluation of neural network explanations},
\newblock \bibinfo{journal}{Information Fusion} \bibinfo{volume}{81} (\bibinfo{year}{2022}) \bibinfo{pages}{14--40}.
\bibitem[{Salewski et~al.(2020)Salewski, Koepke, Lensch, and Akata}]{salewski2020clevrX}
\bibinfo{author}{L.~Salewski}, \bibinfo{author}{A.~S. Koepke}, \bibinfo{author}{H.~P. Lensch}, \bibinfo{author}{Z.~Akata},
\newblock \bibinfo{title}{Clevr-x: A visual reasoning dataset for natural language explanations},
\newblock in: \bibinfo{booktitle}{International Workshop on Extending Explainable AI Beyond Deep Models and Classifiers}, \bibinfo{organization}{Springer}, \bibinfo{year}{2020}, pp. \bibinfo{pages}{69--88}.
\bibitem[{Li et~al.(2023)Li, Wang, Stengel-Eskin, Kortylewski, Ma, Van~Durme, and Yuille}]{li2023superCLEVR}
\bibinfo{author}{Z.~Li}, \bibinfo{author}{X.~Wang}, \bibinfo{author}{E.~Stengel-Eskin}, \bibinfo{author}{A.~Kortylewski}, \bibinfo{author}{W.~Ma}, \bibinfo{author}{B.~Van~Durme}, \bibinfo{author}{A.~L. Yuille},
\newblock \bibinfo{title}{Super-clevr: A virtual benchmark to diagnose domain robustness in visual reasoning},
\newblock in: \bibinfo{booktitle}{Proceedings of the IEEE/CVF Conference on Computer Vision and Pattern Recognition}, \bibinfo{year}{2023}, pp. \bibinfo{pages}{14963--14973}.
\bibitem[{Bitton-Guetta et~al.(2023)Bitton-Guetta, Bitton, Hessel, Schmidt, Elovici, Stanovsky, and Schwartz}]{bitton2023breakingWHOOPS}
\bibinfo{author}{N.~Bitton-Guetta}, \bibinfo{author}{Y.~Bitton}, \bibinfo{author}{J.~Hessel}, \bibinfo{author}{L.~Schmidt}, \bibinfo{author}{Y.~Elovici}, \bibinfo{author}{G.~Stanovsky}, \bibinfo{author}{R.~Schwartz},
\newblock \bibinfo{title}{Breaking common sense: Whoops! a vision-and-language benchmark of synthetic and compositional images},
\newblock \bibinfo{journal}{arXiv preprint arXiv:2303.07274}  (\bibinfo{year}{2023}).
\bibitem[{Deng et~al.(2009)Deng, Dong, Socher, Li, Li, and Fei-Fei}]{deng2009imagenet}
\bibinfo{author}{J.~Deng}, \bibinfo{author}{W.~Dong}, \bibinfo{author}{R.~Socher}, \bibinfo{author}{L.-J. Li}, \bibinfo{author}{K.~Li}, \bibinfo{author}{L.~Fei-Fei},
\newblock \bibinfo{title}{Imagenet: A large-scale hierarchical image database},
\newblock in: \bibinfo{booktitle}{2009 IEEE conference on computer vision and pattern recognition}, \bibinfo{organization}{Ieee}, \bibinfo{year}{2009}, pp. \bibinfo{pages}{248--255}.
\bibitem[{Kuznetsova et~al.(2020)Kuznetsova, Rom, Alldrin, Uijlings, Krasin, Pont-Tuset, Kamali, Popov, Malloci, Kolesnikov et~al.}]{kuznetsova2020open}
\bibinfo{author}{A.~Kuznetsova}, \bibinfo{author}{H.~Rom}, \bibinfo{author}{N.~Alldrin}, \bibinfo{author}{J.~Uijlings}, \bibinfo{author}{I.~Krasin}, \bibinfo{author}{J.~Pont-Tuset}, \bibinfo{author}{S.~Kamali}, \bibinfo{author}{S.~Popov}, \bibinfo{author}{M.~Malloci}, \bibinfo{author}{A.~Kolesnikov}, et~al.,
\newblock \bibinfo{title}{The open images dataset v4: Unified image classification, object detection, and visual relationship detection at scale},
\newblock \bibinfo{journal}{International Journal of Computer Vision} \bibinfo{volume}{128} (\bibinfo{year}{2020}) \bibinfo{pages}{1956--1981}.
\bibitem[{Raffel et~al.(2020)Raffel, Shazeer, Roberts, Lee, Narang, Matena, Zhou, Li, and Liu}]{raffel2020exploring}
\bibinfo{author}{C.~Raffel}, \bibinfo{author}{N.~Shazeer}, \bibinfo{author}{A.~Roberts}, \bibinfo{author}{K.~Lee}, \bibinfo{author}{S.~Narang}, \bibinfo{author}{M.~Matena}, \bibinfo{author}{Y.~Zhou}, \bibinfo{author}{W.~Li}, \bibinfo{author}{P.~J. Liu},
\newblock \bibinfo{title}{Exploring the limits of transfer learning with a unified text-to-text transformer},
\newblock \bibinfo{journal}{The Journal of Machine Learning Research} \bibinfo{volume}{21} (\bibinfo{year}{2020}) \bibinfo{pages}{5485--5551}.
\bibitem[{Commons(2012)}]{commons2012wikimedia}
\bibinfo{author}{W.~Commons},
\newblock \bibinfo{title}{Wikimedia commons},
\newblock \bibinfo{journal}{Retrieved June} \bibinfo{volume}{2} (\bibinfo{year}{2012}).
\bibitem[{Joshi et~al.(2017)Joshi, Choi, Weld, and Zettlemoyer}]{joshi2017triviaqa}
\bibinfo{author}{M.~Joshi}, \bibinfo{author}{E.~Choi}, \bibinfo{author}{D.~S. Weld}, \bibinfo{author}{L.~Zettlemoyer},
\newblock \bibinfo{title}{Triviaqa: A large scale distantly supervised challenge dataset for reading comprehension},
\newblock \bibinfo{journal}{arXiv preprint arXiv:1705.03551}  (\bibinfo{year}{2017}).
\bibitem[{Petroni et~al.(2020)Petroni, Piktus, Fan, Lewis, Yazdani, De~Cao, Thorne, Jernite, Karpukhin, Maillard et~al.}]{petroni2020kilt}
\bibinfo{author}{F.~Petroni}, \bibinfo{author}{A.~Piktus}, \bibinfo{author}{A.~Fan}, \bibinfo{author}{P.~Lewis}, \bibinfo{author}{M.~Yazdani}, \bibinfo{author}{N.~De~Cao}, \bibinfo{author}{J.~Thorne}, \bibinfo{author}{Y.~Jernite}, \bibinfo{author}{V.~Karpukhin}, \bibinfo{author}{J.~Maillard}, et~al.,
\newblock \bibinfo{title}{Kilt: a benchmark for knowledge intensive language tasks},
\newblock \bibinfo{journal}{arXiv preprint arXiv:2009.02252}  (\bibinfo{year}{2020}).
\bibitem[{Raven and Court(1938)}]{raven1938raven}
\bibinfo{author}{J.~C. Raven}, \bibinfo{author}{J.~Court}, \bibinfo{title}{Raven's progressive matrices}, \bibinfo{publisher}{Western Psychological Services Los Angeles, CA}, \bibinfo{year}{1938}.
\bibitem[{Chen et~al.(2021)Chen, Chen, Geng, Pan, Yuan, and Chen}]{chen2021zero}
\bibinfo{author}{Z.~Chen}, \bibinfo{author}{J.~Chen}, \bibinfo{author}{Y.~Geng}, \bibinfo{author}{J.~Z. Pan}, \bibinfo{author}{Z.~Yuan}, \bibinfo{author}{H.~Chen},
\newblock \bibinfo{title}{Zero-shot visual question answering using knowledge graph},
\newblock in: \bibinfo{booktitle}{The Semantic Web--ISWC 2021: 20th International Semantic Web Conference, ISWC 2021, Virtual Event, October 24--28, 2021, Proceedings 20}, \bibinfo{organization}{Springer}, \bibinfo{year}{2021}, pp. \bibinfo{pages}{146--162}.
\bibitem[{Trott et~al.(2017)Trott, Xiong, and Socher}]{trott2017interpretableHowManyQA}
\bibinfo{author}{A.~Trott}, \bibinfo{author}{C.~Xiong}, \bibinfo{author}{R.~Socher},
\newblock \bibinfo{title}{Interpretable counting for visual question answering},
\newblock \bibinfo{journal}{arXiv preprint arXiv:1712.08697}  (\bibinfo{year}{2017}).
\bibitem[{Kahou et~al.(2017)Kahou, Michalski, Atkinson, K{\'a}d{\'a}r, Trischler, and Bengio}]{kahou2017figureqa}
\bibinfo{author}{S.~E. Kahou}, \bibinfo{author}{V.~Michalski}, \bibinfo{author}{A.~Atkinson}, \bibinfo{author}{{\'A}.~K{\'a}d{\'a}r}, \bibinfo{author}{A.~Trischler}, \bibinfo{author}{Y.~Bengio},
\newblock \bibinfo{title}{Figureqa: An annotated figure dataset for visual reasoning},
\newblock \bibinfo{journal}{arXiv preprint arXiv:1710.07300}  (\bibinfo{year}{2017}).
\bibitem[{Kafle et~al.(2018)Kafle, Price, Cohen, and Kanan}]{kafle2018dvqa}
\bibinfo{author}{K.~Kafle}, \bibinfo{author}{B.~Price}, \bibinfo{author}{S.~Cohen}, \bibinfo{author}{C.~Kanan},
\newblock \bibinfo{title}{Dvqa: Understanding data visualizations via question answering},
\newblock in: \bibinfo{booktitle}{Proceedings of the IEEE conference on computer vision and pattern recognition}, \bibinfo{year}{2018}, pp. \bibinfo{pages}{5648--5656}.
\bibitem[{Singh et~al.(2019)Singh, Natarajan, Shah, Jiang, Chen, Batra, Parikh, and Rohrbach}]{singh2019towardsTextVQA}
\bibinfo{author}{A.~Singh}, \bibinfo{author}{V.~Natarajan}, \bibinfo{author}{M.~Shah}, \bibinfo{author}{Y.~Jiang}, \bibinfo{author}{X.~Chen}, \bibinfo{author}{D.~Batra}, \bibinfo{author}{D.~Parikh}, \bibinfo{author}{M.~Rohrbach},
\newblock \bibinfo{title}{Towards vqa models that can read},
\newblock in: \bibinfo{booktitle}{Proceedings of the IEEE/CVF Conference on Computer Vision and Pattern Recognition}, \bibinfo{year}{2019}, pp. \bibinfo{pages}{8317--8326}.
\bibitem[{Chaudhry et~al.(2020)Chaudhry, Shekhar, Gupta, Maneriker, Bansal, and Joshi}]{chaudhry2020leaf}
\bibinfo{author}{R.~Chaudhry}, \bibinfo{author}{S.~Shekhar}, \bibinfo{author}{U.~Gupta}, \bibinfo{author}{P.~Maneriker}, \bibinfo{author}{P.~Bansal}, \bibinfo{author}{A.~Joshi},
\newblock \bibinfo{title}{Leaf-qa: Locate, encode \& attend for figure question answering},
\newblock in: \bibinfo{booktitle}{Proceedings of the IEEE/CVF Winter Conference on Applications of Computer Vision}, \bibinfo{year}{2020}, pp. \bibinfo{pages}{3512--3521}.
\bibitem[{Siegel et~al.(2016)Siegel, Horvitz, Levin, Divvala, and Farhadi}]{siegel2016figureseer}
\bibinfo{author}{N.~Siegel}, \bibinfo{author}{Z.~Horvitz}, \bibinfo{author}{R.~Levin}, \bibinfo{author}{S.~Divvala}, \bibinfo{author}{A.~Farhadi},
\newblock \bibinfo{title}{Figureseer: Parsing result-figures in research papers},
\newblock in: \bibinfo{booktitle}{Computer Vision--ECCV 2016: 14th European Conference, Amsterdam, The Netherlands, October 11--14, 2016, Proceedings, Part VII 14}, \bibinfo{organization}{Springer}, \bibinfo{year}{2016}, pp. \bibinfo{pages}{664--680}.
\bibitem[{Zeng et~al.(2021)Zeng, Zhang, Zhou, and Yang}]{zeng2021beyondOCRplus}
\bibinfo{author}{G.~Zeng}, \bibinfo{author}{Y.~Zhang}, \bibinfo{author}{Y.~Zhou}, \bibinfo{author}{X.~Yang},
\newblock \bibinfo{title}{Beyond ocr+ vqa: Involving ocr into the flow for robust and accurate textvqa},
\newblock in: \bibinfo{booktitle}{Proceedings of the 29th ACM International Conference on Multimedia}, \bibinfo{year}{2021}, pp. \bibinfo{pages}{376--385}.
\bibitem[{Tapaswi et~al.(2016)Tapaswi, Zhu, Stiefelhagen, Torralba, Urtasun, and Fidler}]{tapaswi2016movieqa}
\bibinfo{author}{M.~Tapaswi}, \bibinfo{author}{Y.~Zhu}, \bibinfo{author}{R.~Stiefelhagen}, \bibinfo{author}{A.~Torralba}, \bibinfo{author}{R.~Urtasun}, \bibinfo{author}{S.~Fidler},
\newblock \bibinfo{title}{Movieqa: Understanding stories in movies through question-answering},
\newblock in: \bibinfo{booktitle}{Proceedings of the IEEE conference on computer vision and pattern recognition}, \bibinfo{year}{2016}, pp. \bibinfo{pages}{4631--4640}.
\bibitem[{Yang et~al.(2022)Yang, Wang, Duan, Chen, Hou, Jin, and Zhu}]{yang2022avqa}
\bibinfo{author}{P.~Yang}, \bibinfo{author}{X.~Wang}, \bibinfo{author}{X.~Duan}, \bibinfo{author}{H.~Chen}, \bibinfo{author}{R.~Hou}, \bibinfo{author}{C.~Jin}, \bibinfo{author}{W.~Zhu},
\newblock \bibinfo{title}{Avqa: A dataset for audio-visual question answering on videos},
\newblock in: \bibinfo{booktitle}{Proceedings of the 30th ACM International Conference on Multimedia}, \bibinfo{year}{2022}, pp. \bibinfo{pages}{3480--3491}.
\bibitem[{Garcia et~al.(2020)Garcia, Otani, Chu, and Nakashima}]{garcia2020knowit}
\bibinfo{author}{N.~Garcia}, \bibinfo{author}{M.~Otani}, \bibinfo{author}{C.~Chu}, \bibinfo{author}{Y.~Nakashima},
\newblock \bibinfo{title}{Knowit vqa: Answering knowledge-based questions about videos},
\newblock in: \bibinfo{booktitle}{Proceedings of the AAAI conference on artificial intelligence}, volume~\bibinfo{volume}{34}, \bibinfo{year}{2020}, pp. \bibinfo{pages}{10826--10834}.
\bibitem[{Mun et~al.(2017)Mun, Hongsuck~Seo, Jung, and Han}]{mun2017marioqa}
\bibinfo{author}{J.~Mun}, \bibinfo{author}{P.~Hongsuck~Seo}, \bibinfo{author}{I.~Jung}, \bibinfo{author}{B.~Han},
\newblock \bibinfo{title}{Marioqa: Answering questions by watching gameplay videos},
\newblock in: \bibinfo{booktitle}{Proceedings of the IEEE International Conference on Computer Vision}, \bibinfo{year}{2017}, pp. \bibinfo{pages}{2867--2875}.
\bibitem[{Li et~al.(2016)Li, Song, Cao, Tetreault, Goldberg, Jaimes, and Luo}]{li2016tgif}
\bibinfo{author}{Y.~Li}, \bibinfo{author}{Y.~Song}, \bibinfo{author}{L.~Cao}, \bibinfo{author}{J.~Tetreault}, \bibinfo{author}{L.~Goldberg}, \bibinfo{author}{A.~Jaimes}, \bibinfo{author}{J.~Luo},
\newblock \bibinfo{title}{Tgif: A new dataset and benchmark on animated gif description},
\newblock in: \bibinfo{booktitle}{Proceedings of the IEEE Conference on Computer Vision and Pattern Recognition}, \bibinfo{year}{2016}, pp. \bibinfo{pages}{4641--4650}.
\bibitem[{Zhang et~al.(2023)Zhang, Wu, Zhao, Lin, Zhang, Wang, and Xie}]{zhang2023pmc}
\bibinfo{author}{X.~Zhang}, \bibinfo{author}{C.~Wu}, \bibinfo{author}{Z.~Zhao}, \bibinfo{author}{W.~Lin}, \bibinfo{author}{Y.~Zhang}, \bibinfo{author}{Y.~Wang}, \bibinfo{author}{W.~Xie},
\newblock \bibinfo{title}{Pmc-vqa: Visual instruction tuning for medical visual question answering},
\newblock \bibinfo{journal}{arXiv preprint arXiv:2305.10415}  (\bibinfo{year}{2023}).
\bibitem[{Yang et~al.(2021)Yang, Xia, Liu, Du, Yang, Pelillo, and Zhang}]{yang2021asymmetric}
\bibinfo{author}{K.~Yang}, \bibinfo{author}{G.-S. Xia}, \bibinfo{author}{Z.~Liu}, \bibinfo{author}{B.~Du}, \bibinfo{author}{W.~Yang}, \bibinfo{author}{M.~Pelillo}, \bibinfo{author}{L.~Zhang},
\newblock \bibinfo{title}{Asymmetric siamese networks for semantic change detection in aerial images},
\newblock \bibinfo{journal}{IEEE Transactions on Geoscience and Remote Sensing} \bibinfo{volume}{60} (\bibinfo{year}{2021}) \bibinfo{pages}{1--18}.
\bibitem[{Cortes and Vapnik(1995)}]{cortes1995supportSVM}
\bibinfo{author}{C.~Cortes}, \bibinfo{author}{V.~Vapnik},
\newblock \bibinfo{title}{Support-vector networks},
\newblock \bibinfo{journal}{Machine learning} \bibinfo{volume}{20} (\bibinfo{year}{1995}) \bibinfo{pages}{273--297}.
\bibitem[{Haar(1909)}]{haar1909theorie}
\bibinfo{author}{A.~Haar}, \bibinfo{title}{Zur theorie der orthogonalen funktionensysteme}, \bibinfo{publisher}{Georg-August-Universitat, Gottingen.}, \bibinfo{year}{1909}.
\bibitem[{Viola and Jones(2001)}]{viola2001rapidObjDetHaar}
\bibinfo{author}{P.~Viola}, \bibinfo{author}{M.~Jones},
\newblock \bibinfo{title}{Rapid object detection using a boosted cascade of simple features},
\newblock in: \bibinfo{booktitle}{Proceedings of the 2001 IEEE computer society conference on computer vision and pattern recognition. CVPR 2001}, volume~\bibinfo{volume}{1}, \bibinfo{organization}{Ieee}, \bibinfo{year}{2001}, pp. \bibinfo{pages}{I--I}.
\bibitem[{Dalal and Triggs(2005)}]{dalal2005histogramsHOG}
\bibinfo{author}{N.~Dalal}, \bibinfo{author}{B.~Triggs},
\newblock \bibinfo{title}{Histograms of oriented gradients for human detection},
\newblock in: \bibinfo{booktitle}{2005 IEEE computer society conference on computer vision and pattern recognition (CVPR'05)}, volume~\bibinfo{volume}{1}, \bibinfo{organization}{Ieee}, \bibinfo{year}{2005}, pp. \bibinfo{pages}{886--893}.
\bibitem[{Lowe(1999)}]{lowe1999object}
\bibinfo{author}{D.~G. Lowe},
\newblock \bibinfo{title}{Object recognition from local scale-invariant features},
\newblock in: \bibinfo{booktitle}{Proceedings of the seventh IEEE international conference on computer vision}, volume~\bibinfo{volume}{2}, \bibinfo{organization}{Ieee}, \bibinfo{year}{1999}, pp. \bibinfo{pages}{1150--1157}.
\bibitem[{Hong(1991)}]{hong1991algebraic}
\bibinfo{author}{Z.-Q. Hong},
\newblock \bibinfo{title}{Algebraic feature extraction of image for recognition},
\newblock \bibinfo{journal}{Pattern recognition} \bibinfo{volume}{24} (\bibinfo{year}{1991}) \bibinfo{pages}{211--219}.
\bibitem[{Hyvarinen et~al.(1998)Hyvarinen, Oja, Hoyer, and Hurri}]{hyvarinen1998image}
\bibinfo{author}{A.~Hyvarinen}, \bibinfo{author}{E.~Oja}, \bibinfo{author}{P.~Hoyer}, \bibinfo{author}{J.~Hurri},
\newblock \bibinfo{title}{Image feature extraction by sparse coding and independent component analysis},
\newblock in: \bibinfo{booktitle}{Proceedings. Fourteenth International Conference on Pattern Recognition (Cat. No. 98EX170)}, volume~\bibinfo{volume}{2}, \bibinfo{organization}{IEEE}, \bibinfo{year}{1998}, pp. \bibinfo{pages}{1268--1273}.
\bibitem[{Fukushima and Miyake(1982)}]{fukushima1982neocognitron}
\bibinfo{author}{K.~Fukushima}, \bibinfo{author}{S.~Miyake},
\newblock \bibinfo{title}{Neocognitron: A self-organizing neural network model for a mechanism of visual pattern recognition},
\newblock in: \bibinfo{booktitle}{Competition and cooperation in neural nets}, \bibinfo{publisher}{Springer}, \bibinfo{year}{1982}, pp. \bibinfo{pages}{267--285}.
\bibitem[{Ciregan et~al.(2012)Ciregan, Meier, and Schmidhuber}]{ciregan2012multi}
\bibinfo{author}{D.~Ciregan}, \bibinfo{author}{U.~Meier}, \bibinfo{author}{J.~Schmidhuber},
\newblock \bibinfo{title}{Multi-column deep neural networks for image classification},
\newblock in: \bibinfo{booktitle}{2012 IEEE conference on computer vision and pattern recognition}, \bibinfo{organization}{IEEE}, \bibinfo{year}{2012}, pp. \bibinfo{pages}{3642--3649}.
\bibitem[{Pomerleau(1988)}]{pomerleau1988alvinn}
\bibinfo{author}{D.~A. Pomerleau},
\newblock \bibinfo{title}{Alvinn: An autonomous land vehicle in a neural network},
\newblock \bibinfo{journal}{Advances in neural information processing systems} \bibinfo{volume}{1} (\bibinfo{year}{1988}).
\bibitem[{Sarlashkar et~al.(1998)Sarlashkar, Bodruzzaman, and Malkani}]{sarlashkar1998feature}
\bibinfo{author}{A.~Sarlashkar}, \bibinfo{author}{M.~Bodruzzaman}, \bibinfo{author}{M.~Malkani},
\newblock \bibinfo{title}{Feature extraction using wavelet transform for neural network based image classification},
\newblock in: \bibinfo{booktitle}{Proceedings of Thirtieth Southeastern Symposium on System Theory}, \bibinfo{organization}{IEEE}, \bibinfo{year}{1998}, pp. \bibinfo{pages}{412--416}.
\bibitem[{Lerner et~al.(1999)Lerner, Guterman, Aladjem, and Dinstein}]{lerner1999comparative}
\bibinfo{author}{B.~Lerner}, \bibinfo{author}{H.~Guterman}, \bibinfo{author}{M.~Aladjem}, \bibinfo{author}{I.~h. Dinstein},
\newblock \bibinfo{title}{A comparative study of neural network based feature extraction paradigms},
\newblock \bibinfo{journal}{Pattern Recognition Letters} \bibinfo{volume}{20} (\bibinfo{year}{1999}) \bibinfo{pages}{7--14}.
\bibitem[{Krizhevsky et~al.(2012)Krizhevsky, Sutskever, and Hinton}]{krizhevsky2012imagenetAlexNet}
\bibinfo{author}{A.~Krizhevsky}, \bibinfo{author}{I.~Sutskever}, \bibinfo{author}{G.~E. Hinton},
\newblock \bibinfo{title}{Imagenet classification with deep convolutional neural networks},
\newblock \bibinfo{journal}{Advances in neural information processing systems} \bibinfo{volume}{25} (\bibinfo{year}{2012}).
\bibitem[{LeCun et~al.(1998)LeCun, Bottou, Bengio, and Haffner}]{lecun1998gradient}
\bibinfo{author}{Y.~LeCun}, \bibinfo{author}{L.~Bottou}, \bibinfo{author}{Y.~Bengio}, \bibinfo{author}{P.~Haffner},
\newblock \bibinfo{title}{Gradient-based learning applied to document recognition},
\newblock \bibinfo{journal}{Proceedings of the IEEE} \bibinfo{volume}{86} (\bibinfo{year}{1998}) \bibinfo{pages}{2278--2324}.
\bibitem[{Simonyan and Zisserman(2014)}]{simonyan2014very}
\bibinfo{author}{K.~Simonyan}, \bibinfo{author}{A.~Zisserman},
\newblock \bibinfo{title}{Very deep convolutional networks for large-scale image recognition},
\newblock \bibinfo{journal}{arXiv preprint arXiv:1409.1556}  (\bibinfo{year}{2014}).
\bibitem[{He et~al.(2016)He, Zhang, Ren, and Sun}]{he2016deep}
\bibinfo{author}{K.~He}, \bibinfo{author}{X.~Zhang}, \bibinfo{author}{S.~Ren}, \bibinfo{author}{J.~Sun},
\newblock \bibinfo{title}{Deep residual learning for image recognition},
\newblock in: \bibinfo{booktitle}{Proceedings of the IEEE conference on computer vision and pattern recognition}, \bibinfo{year}{2016}, pp. \bibinfo{pages}{770--778}.
\bibitem[{Szegedy et~al.(2015)Szegedy, Liu, Jia, Sermanet, Reed, Anguelov, Erhan, Vanhoucke, and Rabinovich}]{szegedy2015going}
\bibinfo{author}{C.~Szegedy}, \bibinfo{author}{W.~Liu}, \bibinfo{author}{Y.~Jia}, \bibinfo{author}{P.~Sermanet}, \bibinfo{author}{S.~Reed}, \bibinfo{author}{D.~Anguelov}, \bibinfo{author}{D.~Erhan}, \bibinfo{author}{V.~Vanhoucke}, \bibinfo{author}{A.~Rabinovich},
\newblock \bibinfo{title}{Going deeper with convolutions},
\newblock in: \bibinfo{booktitle}{Proceedings of the IEEE conference on computer vision and pattern recognition}, \bibinfo{year}{2015}, pp. \bibinfo{pages}{1--9}.
\bibitem[{Bozinovski and Fulgosi(1976)}]{bozinovski1976influenceTransfer}
\bibinfo{author}{S.~Bozinovski}, \bibinfo{author}{A.~Fulgosi},
\newblock \bibinfo{title}{The influence of pattern similarity and transfer learning upon training of a base perceptron b2},
\newblock in: \bibinfo{booktitle}{Proceedings of Symposium Informatica}, volume~\bibinfo{volume}{3}, \bibinfo{year}{1976}, pp. \bibinfo{pages}{121--126}.
\bibitem[{Girshick(2015)}]{girshick2015fast}
\bibinfo{author}{R.~Girshick},
\newblock \bibinfo{title}{Fast r-cnn},
\newblock in: \bibinfo{booktitle}{Proceedings of the IEEE international conference on computer vision}, \bibinfo{year}{2015}, pp. \bibinfo{pages}{1440--1448}.
\bibitem[{Ren et~al.(2015)Ren, He, Girshick, and Sun}]{ren2015faster}
\bibinfo{author}{S.~Ren}, \bibinfo{author}{K.~He}, \bibinfo{author}{R.~Girshick}, \bibinfo{author}{J.~Sun},
\newblock \bibinfo{title}{Faster r-cnn: Towards real-time object detection with region proposal networks},
\newblock \bibinfo{journal}{Advances in neural information processing systems} \bibinfo{volume}{28} (\bibinfo{year}{2015}).
\bibitem[{Kafle and Kanan(2016)}]{kafle2016answer}
\bibinfo{author}{K.~Kafle}, \bibinfo{author}{C.~Kanan},
\newblock \bibinfo{title}{Answer-type prediction for visual question answering},
\newblock in: \bibinfo{booktitle}{Proceedings of the IEEE conference on computer vision and pattern recognition}, \bibinfo{year}{2016}, pp. \bibinfo{pages}{4976--4984}.
\bibitem[{Dosovitskiy et~al.(2020)Dosovitskiy, Beyer, Kolesnikov, Weissenborn, Zhai, Unterthiner, Dehghani, Minderer, Heigold, Gelly et~al.}]{dosovitskiy2020image}
\bibinfo{author}{A.~Dosovitskiy}, \bibinfo{author}{L.~Beyer}, \bibinfo{author}{A.~Kolesnikov}, \bibinfo{author}{D.~Weissenborn}, \bibinfo{author}{X.~Zhai}, \bibinfo{author}{T.~Unterthiner}, \bibinfo{author}{M.~Dehghani}, \bibinfo{author}{M.~Minderer}, \bibinfo{author}{G.~Heigold}, \bibinfo{author}{S.~Gelly}, et~al.,
\newblock \bibinfo{title}{An image is worth 16x16 words: Transformers for image recognition at scale},
\newblock \bibinfo{journal}{arXiv preprint arXiv:2010.11929}  (\bibinfo{year}{2020}).
\bibitem[{Liu et~al.(2021)Liu, Lin, Cao, Hu, Wei, Zhang, Lin, and Guo}]{liu2021swin}
\bibinfo{author}{Z.~Liu}, \bibinfo{author}{Y.~Lin}, \bibinfo{author}{Y.~Cao}, \bibinfo{author}{H.~Hu}, \bibinfo{author}{Y.~Wei}, \bibinfo{author}{Z.~Zhang}, \bibinfo{author}{S.~Lin}, \bibinfo{author}{B.~Guo},
\newblock \bibinfo{title}{Swin transformer: Hierarchical vision transformer using shifted windows},
\newblock in: \bibinfo{booktitle}{Proceedings of the IEEE/CVF International Conference on Computer Vision}, \bibinfo{year}{2021}, pp. \bibinfo{pages}{10012--10022}.
\bibitem[{Liu et~al.(2022)Liu, Mao, Wu, Feichtenhofer, Darrell, and Xie}]{liu2022convnet}
\bibinfo{author}{Z.~Liu}, \bibinfo{author}{H.~Mao}, \bibinfo{author}{C.-Y. Wu}, \bibinfo{author}{C.~Feichtenhofer}, \bibinfo{author}{T.~Darrell}, \bibinfo{author}{S.~Xie},
\newblock \bibinfo{title}{A convnet for the 2020s},
\newblock \bibinfo{journal}{arXiv preprint arXiv:2201.03545}  (\bibinfo{year}{2022}).
\bibitem[{Hirota et~al.(2021)Hirota, Garcia, Otani, Chu, Nakashima, Taniguchi, and Onoye}]{hirota2021picture}
\bibinfo{author}{Y.~Hirota}, \bibinfo{author}{N.~Garcia}, \bibinfo{author}{M.~Otani}, \bibinfo{author}{C.~Chu}, \bibinfo{author}{Y.~Nakashima}, \bibinfo{author}{I.~Taniguchi}, \bibinfo{author}{T.~Onoye},
\newblock \bibinfo{title}{A picture may be worth a hundred words for visual question answering},
\newblock \bibinfo{journal}{arXiv preprint arXiv:2106.13445}  (\bibinfo{year}{2021}).
\bibitem[{Xue et~al.(2021)Xue, Huang, Liu, Peng, Fu, Li, and Luo}]{xue2021probing}
\bibinfo{author}{H.~Xue}, \bibinfo{author}{Y.~Huang}, \bibinfo{author}{B.~Liu}, \bibinfo{author}{H.~Peng}, \bibinfo{author}{J.~Fu}, \bibinfo{author}{H.~Li}, \bibinfo{author}{J.~Luo},
\newblock \bibinfo{title}{Probing inter-modality: Visual parsing with self-attention for vision-and-language pre-training},
\newblock \bibinfo{journal}{Advances in Neural Information Processing Systems} \bibinfo{volume}{34} (\bibinfo{year}{2021}).
\bibitem[{Luo et~al.(2022)Luo, Zhou, Sun, Wang, Cao, Wu, Huang, and Ji}]{luo2022towards}
\bibinfo{author}{G.~Luo}, \bibinfo{author}{Y.~Zhou}, \bibinfo{author}{X.~Sun}, \bibinfo{author}{Y.~Wang}, \bibinfo{author}{L.~Cao}, \bibinfo{author}{Y.~Wu}, \bibinfo{author}{F.~Huang}, \bibinfo{author}{R.~Ji},
\newblock \bibinfo{title}{Towards lightweight transformer via group-wise transformation for vision-and-language tasks},
\newblock \bibinfo{journal}{IEEE Transactions on Image Processing}  (\bibinfo{year}{2022}).
\bibitem[{Miller and Charles(1991)}]{miller1991contextual}
\bibinfo{author}{G.~A. Miller}, \bibinfo{author}{W.~G. Charles},
\newblock \bibinfo{title}{Contextual correlates of semantic similarity},
\newblock \bibinfo{journal}{Language and cognitive processes} \bibinfo{volume}{6} (\bibinfo{year}{1991}) \bibinfo{pages}{1--28}.
\bibitem[{Eckart and Young(1936)}]{eckart1936approximation}
\bibinfo{author}{C.~Eckart}, \bibinfo{author}{G.~Young},
\newblock \bibinfo{title}{The approximation of one matrix by another of lower rank},
\newblock \bibinfo{journal}{Psychometrika} \bibinfo{volume}{1} (\bibinfo{year}{1936}) \bibinfo{pages}{211--218}.
\bibitem[{Xu and Rudnicky(2000)}]{xu2000can}
\bibinfo{author}{W.~Xu}, \bibinfo{author}{A.~Rudnicky},
\newblock \bibinfo{title}{Can artificial neural networks learn language models?},
\newblock \bibinfo{journal}{Sixth International Conference on Spoken Language Processing}  (\bibinfo{year}{2000}) \bibinfo{pages}{202--205}.
\bibitem[{Bengio et~al.(2000)Bengio, Ducharme, and Vincent}]{bengio2000neural}
\bibinfo{author}{Y.~Bengio}, \bibinfo{author}{R.~Ducharme}, \bibinfo{author}{P.~Vincent},
\newblock \bibinfo{title}{A neural probabilistic language model},
\newblock \bibinfo{journal}{Advances in Neural Information Processing Systems} \bibinfo{volume}{13} (\bibinfo{year}{2000}).
\bibitem[{Rumelhart et~al.(1985)Rumelhart, Hinton, Williams et~al.}]{rumelhart1985learningRNN}
\bibinfo{author}{D.~E. Rumelhart}, \bibinfo{author}{G.~E. Hinton}, \bibinfo{author}{R.~J. Williams}, et~al., \bibinfo{title}{Learning internal representations by error propagation}, \bibinfo{year}{1985}.
\bibitem[{Mikolov et~al.(2013{\natexlab{a}})Mikolov, Yih, and Zweig}]{mikolov2013linguistic}
\bibinfo{author}{T.~Mikolov}, \bibinfo{author}{W.-t. Yih}, \bibinfo{author}{G.~Zweig},
\newblock \bibinfo{title}{Linguistic regularities in continuous space word representations},
\newblock in: \bibinfo{booktitle}{Proceedings of the 2013 conference of the north american chapter of the association for computational linguistics: Human language technologies}, \bibinfo{year}{2013}{\natexlab{a}}, pp. \bibinfo{pages}{746--751}.
\bibitem[{Mikolov et~al.(2013{\natexlab{b}})Mikolov, Chen, Corrado, and Dean}]{mikolov2013efficient}
\bibinfo{author}{T.~Mikolov}, \bibinfo{author}{K.~Chen}, \bibinfo{author}{G.~Corrado}, \bibinfo{author}{J.~Dean},
\newblock \bibinfo{title}{Efficient estimation of word representations in vector space},
\newblock \bibinfo{journal}{arXiv preprint arXiv:1301.3781}  (\bibinfo{year}{2013}{\natexlab{b}}).
\bibitem[{Chung et~al.(2014)Chung, Gulcehre, Cho, and Bengio}]{chung2014empirical}
\bibinfo{author}{J.~Chung}, \bibinfo{author}{C.~Gulcehre}, \bibinfo{author}{K.~Cho}, \bibinfo{author}{Y.~Bengio},
\newblock \bibinfo{title}{Empirical evaluation of gated recurrent neural networks on sequence modeling},
\newblock \bibinfo{journal}{arXiv preprint arXiv:1412.3555}  (\bibinfo{year}{2014}).
\bibitem[{Hochreiter and Schmidhuber(1997)}]{hochreiter1997longLSTM}
\bibinfo{author}{S.~Hochreiter}, \bibinfo{author}{J.~Schmidhuber},
\newblock \bibinfo{title}{Long short-term memory},
\newblock \bibinfo{journal}{Neural computation} \bibinfo{volume}{9} (\bibinfo{year}{1997}) \bibinfo{pages}{1735--1780}.
\bibitem[{Yang et~al.(2021)Yang, Garcia, Chu, Otani, Nakashima, and Takemura}]{yang2021comparative}
\bibinfo{author}{Z.~Yang}, \bibinfo{author}{N.~Garcia}, \bibinfo{author}{C.~Chu}, \bibinfo{author}{M.~Otani}, \bibinfo{author}{Y.~Nakashima}, \bibinfo{author}{H.~Takemura},
\newblock \bibinfo{title}{A comparative study of language transformers for video question answering},
\newblock \bibinfo{journal}{Neurocomputing} \bibinfo{volume}{445} (\bibinfo{year}{2021}) \bibinfo{pages}{121--133}.
\bibitem[{Biten et~al.(2021)Biten, Litman, Xie, Appalaraju, and Manmatha}]{biten2021latr}
\bibinfo{author}{A.~F. Biten}, \bibinfo{author}{R.~Litman}, \bibinfo{author}{Y.~Xie}, \bibinfo{author}{S.~Appalaraju}, \bibinfo{author}{R.~Manmatha},
\newblock \bibinfo{title}{Latr: Layout-aware transformer for scene-text vqa},
\newblock \bibinfo{journal}{arXiv preprint arXiv:2112.12494}  (\bibinfo{year}{2021}).
\bibitem[{Yang et~al.(2021)Yang, Lu, Wang, Yin, Florencio, Wang, Zhang, Zhang, and Luo}]{yang2021tap}
\bibinfo{author}{Z.~Yang}, \bibinfo{author}{Y.~Lu}, \bibinfo{author}{J.~Wang}, \bibinfo{author}{X.~Yin}, \bibinfo{author}{D.~Florencio}, \bibinfo{author}{L.~Wang}, \bibinfo{author}{C.~Zhang}, \bibinfo{author}{L.~Zhang}, \bibinfo{author}{J.~Luo},
\newblock \bibinfo{title}{Tap: Text-aware pre-training for text-vqa and text-caption},
\newblock in: \bibinfo{booktitle}{Proceedings of the IEEE/CVF Conference on Computer Vision and Pattern Recognition}, \bibinfo{year}{2021}, pp. \bibinfo{pages}{8751--8761}.
\bibitem[{Chen et~al.(2015)Chen, Wang, Chen, Gao, Xu, and Nevatia}]{chen2015abc}
\bibinfo{author}{K.~Chen}, \bibinfo{author}{J.~Wang}, \bibinfo{author}{L.-C. Chen}, \bibinfo{author}{H.~Gao}, \bibinfo{author}{W.~Xu}, \bibinfo{author}{R.~Nevatia},
\newblock \bibinfo{title}{Abc-cnn: An attention based convolutional neural network for visual question answering},
\newblock \bibinfo{journal}{arXiv preprint arXiv:1511.05960}  (\bibinfo{year}{2015}).
\bibitem[{Zhou et~al.(2015)Zhou, Tian, Sukhbaatar, Szlam, and Fergus}]{zhou2015simple}
\bibinfo{author}{B.~Zhou}, \bibinfo{author}{Y.~Tian}, \bibinfo{author}{S.~Sukhbaatar}, \bibinfo{author}{A.~Szlam}, \bibinfo{author}{R.~Fergus},
\newblock \bibinfo{title}{Simple baseline for visual question answering},
\newblock \bibinfo{journal}{arXiv preprint arXiv:1512.02167}  (\bibinfo{year}{2015}).
\bibitem[{Jabri et~al.(2016)Jabri, Joulin, and van~der Maaten}]{jabri2016revisiting}
\bibinfo{author}{A.~Jabri}, \bibinfo{author}{A.~Joulin}, \bibinfo{author}{L.~van~der Maaten},
\newblock \bibinfo{title}{Revisiting visual question answering baselines},
\newblock in: \bibinfo{editor}{B.~Leibe}, \bibinfo{editor}{J.~Matas}, \bibinfo{editor}{N.~Sebe}, \bibinfo{editor}{M.~Welling} (Eds.), \bibinfo{booktitle}{Computer Vision -- ECCV 2016}, \bibinfo{publisher}{Springer International Publishing}, \bibinfo{address}{Cham}, \bibinfo{year}{2016}, pp. \bibinfo{pages}{727--739}.
\bibitem[{Huang et~al.(2019)Huang, Dao, Alfadly, and Ghanem}]{huang2019novel}
\bibinfo{author}{J.-H. Huang}, \bibinfo{author}{C.~D. Dao}, \bibinfo{author}{M.~Alfadly}, \bibinfo{author}{B.~Ghanem},
\newblock \bibinfo{title}{A novel framework for robustness analysis of visual qa models},
\newblock in: \bibinfo{booktitle}{Proceedings of the AAAI Conference on Artificial Intelligence}, volume \bibinfo{volume}{33-01}, \bibinfo{year}{2019}, pp. \bibinfo{pages}{8449--8456}.
\bibitem[{Teney et~al.(2018)Teney, Anderson, He, and Van Den~Hengel}]{teney2018tips}
\bibinfo{author}{D.~Teney}, \bibinfo{author}{P.~Anderson}, \bibinfo{author}{X.~He}, \bibinfo{author}{A.~Van Den~Hengel},
\newblock \bibinfo{title}{Tips and tricks for visual question answering: Learnings from the 2017 challenge},
\newblock in: \bibinfo{booktitle}{Proceedings of the IEEE conference on computer vision and pattern recognition}, \bibinfo{year}{2018}, pp. \bibinfo{pages}{4223--4232}.
\bibitem[{Yang et~al.(2016)Yang, He, Gao, Deng, and Smola}]{yang2016stacked}
\bibinfo{author}{Z.~Yang}, \bibinfo{author}{X.~He}, \bibinfo{author}{J.~Gao}, \bibinfo{author}{L.~Deng}, \bibinfo{author}{A.~Smola},
\newblock \bibinfo{title}{Stacked attention networks for image question answering},
\newblock in: \bibinfo{booktitle}{Proceedings of the IEEE conference on computer vision and pattern recognition}, \bibinfo{year}{2016}, pp. \bibinfo{pages}{21--29}.
\bibitem[{Fukui et~al.(2016)Fukui, Park, Yang, Rohrbach, Darrell, and Rohrbach}]{fukui2016multimodalVQAVisualGrounding}
\bibinfo{author}{A.~Fukui}, \bibinfo{author}{D.~H. Park}, \bibinfo{author}{D.~Yang}, \bibinfo{author}{A.~Rohrbach}, \bibinfo{author}{T.~Darrell}, \bibinfo{author}{M.~Rohrbach},
\newblock \bibinfo{title}{Multimodal compact bilinear pooling for visual question answering and visual grounding},
\newblock \bibinfo{journal}{CoRR} \bibinfo{volume}{abs/1606.01847} (\bibinfo{year}{2016}). \URLprefix \url{http://arxiv.org/abs/1606.01847}. \href{http://arxiv.org/abs/1606.01847}{{\tt arXiv:1606.01847}}.
\bibitem[{Ben-Younes et~al.(2017)Ben-Younes, Cadene, Cord, and Thome}]{ben2017mutan}
\bibinfo{author}{H.~Ben-Younes}, \bibinfo{author}{R.~Cadene}, \bibinfo{author}{M.~Cord}, \bibinfo{author}{N.~Thome},
\newblock \bibinfo{title}{Mutan: Multimodal tucker fusion for visual question answering},
\newblock in: \bibinfo{booktitle}{Proceedings of the IEEE international conference on computer vision}, \bibinfo{year}{2017}, pp. \bibinfo{pages}{2612--2620}.
\bibitem[{Ma et~al.(2016)Ma, Lu, and Li}]{ma2016learning}
\bibinfo{author}{L.~Ma}, \bibinfo{author}{Z.~Lu}, \bibinfo{author}{H.~Li},
\newblock \bibinfo{title}{Learning to answer questions from image using convolutional neural network},
\newblock in: \bibinfo{booktitle}{Proceedings of the AAAI Conference on Artificial Intelligence}, volume~\bibinfo{volume}{30} of \textit{\bibinfo{series}{AAAI'16}}, \bibinfo{publisher}{AAAI Press}, \bibinfo{year}{2016}, p. \bibinfo{pages}{3567–3573}.
\bibitem[{Xu and Saenko(2016)}]{xu2016ask}
\bibinfo{author}{H.~Xu}, \bibinfo{author}{K.~Saenko},
\newblock \bibinfo{title}{Ask, attend and answer: Exploring question-guided spatial attention for visual question answering},
\newblock in: \bibinfo{booktitle}{Computer Vision--ECCV 2016: 14th European Conference, Amsterdam, The Netherlands, October 11--14, 2016, Proceedings, Part VII 14}, \bibinfo{organization}{Springer}, \bibinfo{year}{2016}, pp. \bibinfo{pages}{451--466}.
\bibitem[{Noh et~al.(2016)Noh, Seo, and Han}]{noh2016image}
\bibinfo{author}{H.~Noh}, \bibinfo{author}{P.~H. Seo}, \bibinfo{author}{B.~Han},
\newblock \bibinfo{title}{Image question answering using convolutional neural network with dynamic parameter prediction},
\newblock in: \bibinfo{booktitle}{Proceedings of the IEEE conference on computer vision and pattern recognition}, \bibinfo{year}{2016}, pp. \bibinfo{pages}{30--38}.
\bibitem[{Shih et~al.(2016)Shih, Singh, and Hoiem}]{shih2016look}
\bibinfo{author}{K.~J. Shih}, \bibinfo{author}{S.~Singh}, \bibinfo{author}{D.~Hoiem},
\newblock \bibinfo{title}{Where to look: Focus regions for visual question answering},
\newblock in: \bibinfo{booktitle}{Proceedings of the IEEE conference on computer vision and pattern recognition}, \bibinfo{year}{2016}, pp. \bibinfo{pages}{4613--4621}.
\bibitem[{Kim et~al.(2016)Kim, Lee, Kwak, Heo, Kim, Ha, and Zhang}]{kim2016multimodal}
\bibinfo{author}{J.-H. Kim}, \bibinfo{author}{S.-W. Lee}, \bibinfo{author}{D.~Kwak}, \bibinfo{author}{M.-O. Heo}, \bibinfo{author}{J.~Kim}, \bibinfo{author}{J.-W. Ha}, \bibinfo{author}{B.-T. Zhang},
\newblock \bibinfo{title}{Multimodal residual learning for visual qa},
\newblock \bibinfo{journal}{Advances in neural information processing systems} \bibinfo{volume}{29} (\bibinfo{year}{2016}).
\bibitem[{Nam et~al.(2017)Nam, Ha, and Kim}]{nam2017dual}
\bibinfo{author}{H.~Nam}, \bibinfo{author}{J.-W. Ha}, \bibinfo{author}{J.~Kim},
\newblock \bibinfo{title}{Dual attention networks for multimodal reasoning and matching},
\newblock in: \bibinfo{booktitle}{Proceedings of the IEEE conference on computer vision and pattern recognition}, \bibinfo{year}{2017}, pp. \bibinfo{pages}{299--307}.
\bibitem[{Kim et~al.(2016)Kim, On, Lim, Kim, Ha, and Zhang}]{kim2016hadamard}
\bibinfo{author}{J.-H. Kim}, \bibinfo{author}{K.-W. On}, \bibinfo{author}{W.~Lim}, \bibinfo{author}{J.~Kim}, \bibinfo{author}{J.-W. Ha}, \bibinfo{author}{B.-T. Zhang},
\newblock \bibinfo{title}{Hadamard product for low-rank bilinear pooling},
\newblock \bibinfo{journal}{arXiv preprint arXiv:1610.04325}  (\bibinfo{year}{2016}).
\bibitem[{Lu et~al.(2016)Lu, Yang, Batra, and Parikh}]{lu2016hierarchical}
\bibinfo{author}{J.~Lu}, \bibinfo{author}{J.~Yang}, \bibinfo{author}{D.~Batra}, \bibinfo{author}{D.~Parikh},
\newblock \bibinfo{title}{Hierarchical question-image co-attention for visual question answering},
\newblock \bibinfo{journal}{Advances in neural information processing systems} \bibinfo{volume}{29} (\bibinfo{year}{2016}).
\bibitem[{Xiong et~al.(2016)Xiong, Merity, and Socher}]{xiong2016dynamic}
\bibinfo{author}{C.~Xiong}, \bibinfo{author}{S.~Merity}, \bibinfo{author}{R.~Socher},
\newblock \bibinfo{title}{Dynamic memory networks for visual and textual question answering},
\newblock in: \bibinfo{booktitle}{International conference on machine learning}, \bibinfo{organization}{PMLR}, \bibinfo{year}{2016}, pp. \bibinfo{pages}{2397--2406}.
\bibitem[{Wu et~al.(2017)Wu, Shen, Wang, Dick, and Van Den~Hengel}]{wu2017image}
\bibinfo{author}{Q.~Wu}, \bibinfo{author}{C.~Shen}, \bibinfo{author}{P.~Wang}, \bibinfo{author}{A.~Dick}, \bibinfo{author}{A.~Van Den~Hengel},
\newblock \bibinfo{title}{Image captioning and visual question answering based on attributes and external knowledge},
\newblock \bibinfo{journal}{IEEE transactions on pattern analysis and machine intelligence} \bibinfo{volume}{40} (\bibinfo{year}{2017}) \bibinfo{pages}{1367--1381}.
\bibitem[{Yu et~al.(2017)Yu, Fu, Mei, and Rui}]{yu2017multi}
\bibinfo{author}{D.~Yu}, \bibinfo{author}{J.~Fu}, \bibinfo{author}{T.~Mei}, \bibinfo{author}{Y.~Rui},
\newblock \bibinfo{title}{Multi-level attention networks for visual question answering},
\newblock in: \bibinfo{booktitle}{Proceedings of the IEEE conference on computer vision and pattern recognition}, \bibinfo{year}{2017}, pp. \bibinfo{pages}{4709--4717}.
\bibitem[{Kazemi and Elqursh(2017)}]{kazemi2017show}
\bibinfo{author}{V.~Kazemi}, \bibinfo{author}{A.~Elqursh},
\newblock \bibinfo{title}{Show, ask, attend, and answer: A strong baseline for visual question answering},
\newblock \bibinfo{journal}{arXiv preprint arXiv:1704.03162}  (\bibinfo{year}{2017}).
\bibitem[{Anderson et~al.(2018)Anderson, He, Buehler, Teney, Johnson, Gould, and Zhang}]{anderson2018bottom}
\bibinfo{author}{P.~Anderson}, \bibinfo{author}{X.~He}, \bibinfo{author}{C.~Buehler}, \bibinfo{author}{D.~Teney}, \bibinfo{author}{M.~Johnson}, \bibinfo{author}{S.~Gould}, \bibinfo{author}{L.~Zhang},
\newblock \bibinfo{title}{Bottom-up and top-down attention for image captioning and visual question answering},
\newblock in: \bibinfo{booktitle}{Proceedings of the IEEE conference on computer vision and pattern recognition}, \bibinfo{year}{2018}, pp. \bibinfo{pages}{6077--6086}.
\bibitem[{Yu et~al.(2018)Yu, Yu, Xiang, Fan, and Tao}]{yu2018beyond}
\bibinfo{author}{Z.~Yu}, \bibinfo{author}{J.~Yu}, \bibinfo{author}{C.~Xiang}, \bibinfo{author}{J.~Fan}, \bibinfo{author}{D.~Tao},
\newblock \bibinfo{title}{Beyond bilinear: Generalized multimodal factorized high-order pooling for visual question answering},
\newblock \bibinfo{journal}{IEEE transactions on neural networks and learning systems} \bibinfo{volume}{29} (\bibinfo{year}{2018}) \bibinfo{pages}{5947--5959}.
\bibitem[{Nguyen and Okatani(2018)}]{nguyen2018improved}
\bibinfo{author}{D.-K. Nguyen}, \bibinfo{author}{T.~Okatani},
\newblock \bibinfo{title}{Improved fusion of visual and language representations by dense symmetric co-attention for visual question answering},
\newblock in: \bibinfo{booktitle}{Proceedings of the IEEE conference on computer vision and pattern recognition}, \bibinfo{year}{2018}, pp. \bibinfo{pages}{6087--6096}.
\bibitem[{Kim et~al.(2018)Kim, Jun, and Zhang}]{kim2018bilinear}
\bibinfo{author}{J.-H. Kim}, \bibinfo{author}{J.~Jun}, \bibinfo{author}{B.-T. Zhang},
\newblock \bibinfo{title}{Bilinear attention networks},
\newblock \bibinfo{journal}{Advances in neural information processing systems} \bibinfo{volume}{31} (\bibinfo{year}{2018}).
\bibitem[{Yu et~al.(2019)Yu, Yu, Cui, Tao, and Tian}]{yu2019deep}
\bibinfo{author}{Z.~Yu}, \bibinfo{author}{J.~Yu}, \bibinfo{author}{Y.~Cui}, \bibinfo{author}{D.~Tao}, \bibinfo{author}{Q.~Tian},
\newblock \bibinfo{title}{Deep modular co-attention networks for visual question answering},
\newblock in: \bibinfo{booktitle}{Proceedings of the IEEE/CVF conference on computer vision and pattern recognition}, \bibinfo{year}{2019}, pp. \bibinfo{pages}{6281--6290}.
\bibitem[{Ba et~al.(2014)Ba, Mnih, and Kavukcuoglu}]{ba2014multipleObjRecogAttention}
\bibinfo{author}{J.~Ba}, \bibinfo{author}{V.~Mnih}, \bibinfo{author}{K.~Kavukcuoglu},
\newblock \bibinfo{title}{Multiple object recognition with visual attention},
\newblock \bibinfo{journal}{arXiv preprint arXiv:1412.7755}  (\bibinfo{year}{2014}).
\bibitem[{Jin et~al.(2015)Jin, Fu, Cui, Sha, and Zhang}]{jin2015aligningImgCapAttention}
\bibinfo{author}{J.~Jin}, \bibinfo{author}{K.~Fu}, \bibinfo{author}{R.~Cui}, \bibinfo{author}{F.~Sha}, \bibinfo{author}{C.~Zhang},
\newblock \bibinfo{title}{Aligning where to see and what to tell: image caption with region-based attention and scene factorization},
\newblock \bibinfo{journal}{arXiv preprint arXiv:1506.06272}  (\bibinfo{year}{2015}).
\bibitem[{Peng et~al.(2019)Peng, Yang, Bin, Xie, Shen, Ji, and Xu}]{peng2019word}
\bibinfo{author}{L.~Peng}, \bibinfo{author}{Y.~Yang}, \bibinfo{author}{Y.~Bin}, \bibinfo{author}{N.~Xie}, \bibinfo{author}{F.~Shen}, \bibinfo{author}{Y.~Ji}, \bibinfo{author}{X.~Xu},
\newblock \bibinfo{title}{Word-to-region attention network for visual question answering},
\newblock \bibinfo{journal}{Multimedia Tools and Applications} \bibinfo{volume}{78} (\bibinfo{year}{2019}) \bibinfo{pages}{3843--3858}.
\bibitem[{Malinowski et~al.(2018)Malinowski, Doersch, Santoro, and Battaglia}]{malinowski2018learning}
\bibinfo{author}{M.~Malinowski}, \bibinfo{author}{C.~Doersch}, \bibinfo{author}{A.~Santoro}, \bibinfo{author}{P.~Battaglia},
\newblock \bibinfo{title}{Learning visual question answering by bootstrapping hard attention},
\newblock in: \bibinfo{booktitle}{Proceedings of the European Conference on Computer Vision (ECCV)}, \bibinfo{year}{2018}, pp. \bibinfo{pages}{3--20}.
\bibitem[{Rahman et~al.(2021)Rahman, Chou, Sigal, and Carenini}]{rahman2021improved}
\bibinfo{author}{T.~Rahman}, \bibinfo{author}{S.-H. Chou}, \bibinfo{author}{L.~Sigal}, \bibinfo{author}{G.~Carenini},
\newblock \bibinfo{title}{An improved attention for visual question answering},
\newblock in: \bibinfo{booktitle}{Proceedings of the IEEE/CVF Conference on Computer Vision and Pattern Recognition}, \bibinfo{year}{2021}, pp. \bibinfo{pages}{1653--1662}.
\bibitem[{Devlin et~al.(2018)Devlin, Chang, Lee, and Toutanova}]{devlin2018bert}
\bibinfo{author}{J.~Devlin}, \bibinfo{author}{M.-W. Chang}, \bibinfo{author}{K.~Lee}, \bibinfo{author}{K.~Toutanova},
\newblock \bibinfo{title}{Bert: Pre-training of deep bidirectional transformers for language understanding},
\newblock \bibinfo{journal}{arXiv preprint arXiv:1810.04805}  (\bibinfo{year}{2018}).
\bibitem[{Liu et~al.(2019)Liu, Ott, Goyal, Du, Joshi, Chen, Levy, Lewis, Zettlemoyer, and Stoyanov}]{liu2019roberta}
\bibinfo{author}{Y.~Liu}, \bibinfo{author}{M.~Ott}, \bibinfo{author}{N.~Goyal}, \bibinfo{author}{J.~Du}, \bibinfo{author}{M.~Joshi}, \bibinfo{author}{D.~Chen}, \bibinfo{author}{O.~Levy}, \bibinfo{author}{M.~Lewis}, \bibinfo{author}{L.~Zettlemoyer}, \bibinfo{author}{V.~Stoyanov},
\newblock \bibinfo{title}{Roberta: A robustly optimized bert pretraining approach},
\newblock \bibinfo{journal}{arXiv preprint arXiv:1907.11692}  (\bibinfo{year}{2019}).
\bibitem[{Lu et~al.(2019)Lu, Batra, Parikh, and Lee}]{lu2019vilbert}
\bibinfo{author}{J.~Lu}, \bibinfo{author}{D.~Batra}, \bibinfo{author}{D.~Parikh}, \bibinfo{author}{S.~Lee},
\newblock \bibinfo{title}{Vilbert: Pretraining task-agnostic visiolinguistic representations for vision-and-language tasks},
\newblock \bibinfo{journal}{Advances in neural information processing systems} \bibinfo{volume}{32} (\bibinfo{year}{2019}).
\bibitem[{Taylor(1953)}]{taylor1953clozeMLM}
\bibinfo{author}{W.~L. Taylor},
\newblock \bibinfo{title}{“cloze procedure”: A new tool for measuring readability},
\newblock \bibinfo{journal}{Journalism quarterly} \bibinfo{volume}{30} (\bibinfo{year}{1953}) \bibinfo{pages}{415--433}.
\bibitem[{Chen et~al.(2020)Chen, Li, Yu, El~Kholy, Ahmed, Gan, Cheng, and Liu}]{chen2020uniter}
\bibinfo{author}{Y.-C. Chen}, \bibinfo{author}{L.~Li}, \bibinfo{author}{L.~Yu}, \bibinfo{author}{A.~El~Kholy}, \bibinfo{author}{F.~Ahmed}, \bibinfo{author}{Z.~Gan}, \bibinfo{author}{Y.~Cheng}, \bibinfo{author}{J.~Liu},
\newblock \bibinfo{title}{Uniter: Universal image-text representation learning},
\newblock in: \bibinfo{booktitle}{European conference on computer vision}, \bibinfo{organization}{Springer}, \bibinfo{year}{2020}, pp. \bibinfo{pages}{104--120}.
\bibitem[{Li et~al.(2021)Li, Selvaraju, Gotmare, Joty, Xiong, and Hoi}]{li2021alignALBEF}
\bibinfo{author}{J.~Li}, \bibinfo{author}{R.~Selvaraju}, \bibinfo{author}{A.~Gotmare}, \bibinfo{author}{S.~Joty}, \bibinfo{author}{C.~Xiong}, \bibinfo{author}{S.~C.~H. Hoi},
\newblock \bibinfo{title}{Align before fuse: Vision and language representation learning with momentum distillation},
\newblock \bibinfo{journal}{Advances in neural information processing systems} \bibinfo{volume}{34} (\bibinfo{year}{2021}) \bibinfo{pages}{9694--9705}.
\bibitem[{Vinyals et~al.(2015)Vinyals, Toshev, Bengio, and Erhan}]{vinyals2015show}
\bibinfo{author}{O.~Vinyals}, \bibinfo{author}{A.~Toshev}, \bibinfo{author}{S.~Bengio}, \bibinfo{author}{D.~Erhan},
\newblock \bibinfo{title}{Show and tell: A neural image caption generator},
\newblock in: \bibinfo{booktitle}{Proceedings of the IEEE conference on computer vision and pattern recognition}, \bibinfo{year}{2015}, pp. \bibinfo{pages}{3156--3164}.
\bibitem[{Xie et~al.(2019)Xie, Lai, Doran, and Kadav}]{xie2019visualEntailment}
\bibinfo{author}{N.~Xie}, \bibinfo{author}{F.~Lai}, \bibinfo{author}{D.~Doran}, \bibinfo{author}{A.~Kadav},
\newblock \bibinfo{title}{Visual entailment: A novel task for fine-grained image understanding},
\newblock \bibinfo{journal}{arXiv preprint arXiv:1901.06706}  (\bibinfo{year}{2019}).
\bibitem[{Tan and Bansal(2019)}]{tan2019lxmert}
\bibinfo{author}{H.~Tan}, \bibinfo{author}{M.~Bansal},
\newblock \bibinfo{title}{Lxmert: Learning cross-modality encoder representations from transformers},
\newblock \bibinfo{journal}{arXiv preprint arXiv:1908.07490}  (\bibinfo{year}{2019}).
\bibitem[{Su et~al.(2019)Su, Zhu, Cao, Li, Lu, Wei, and Dai}]{su2019vlBert}
\bibinfo{author}{W.~Su}, \bibinfo{author}{X.~Zhu}, \bibinfo{author}{Y.~Cao}, \bibinfo{author}{B.~Li}, \bibinfo{author}{L.~Lu}, \bibinfo{author}{F.~Wei}, \bibinfo{author}{J.~Dai},
\newblock \bibinfo{title}{Vl-bert: Pre-training of generic visual-linguistic representations},
\newblock \bibinfo{journal}{arXiv preprint arXiv:1908.08530}  (\bibinfo{year}{2019}).
\bibitem[{Zhou et~al.(2020)Zhou, Palangi, Zhang, Hu, Corso, and Gao}]{zhou2020unified}
\bibinfo{author}{L.~Zhou}, \bibinfo{author}{H.~Palangi}, \bibinfo{author}{L.~Zhang}, \bibinfo{author}{H.~Hu}, \bibinfo{author}{J.~Corso}, \bibinfo{author}{J.~Gao},
\newblock \bibinfo{title}{Unified vision-language pre-training for image captioning and vqa},
\newblock in: \bibinfo{booktitle}{Proceedings of the AAAI conference on artificial intelligence}, volume \bibinfo{volume}{34-07}, \bibinfo{year}{2020}, pp. \bibinfo{pages}{13041--13049}.
\bibitem[{Kim et~al.(2021)Kim, Son, and Kim}]{kim2021vilt}
\bibinfo{author}{W.~Kim}, \bibinfo{author}{B.~Son}, \bibinfo{author}{I.~Kim},
\newblock \bibinfo{title}{Vilt: Vision-and-language transformer without convolution or region supervision},
\newblock in: \bibinfo{booktitle}{International Conference on Machine Learning}, \bibinfo{organization}{PMLR}, \bibinfo{year}{2021}, pp. \bibinfo{pages}{5583--5594}.
\bibitem[{Bao et~al.(2022)Bao, Wang, Dong, Liu, Mohammed, Aggarwal, Som, Piao, and Wei}]{bao2022vlmo}
\bibinfo{author}{H.~Bao}, \bibinfo{author}{W.~Wang}, \bibinfo{author}{L.~Dong}, \bibinfo{author}{Q.~Liu}, \bibinfo{author}{O.~K. Mohammed}, \bibinfo{author}{K.~Aggarwal}, \bibinfo{author}{S.~Som}, \bibinfo{author}{S.~Piao}, \bibinfo{author}{F.~Wei},
\newblock \bibinfo{title}{Vlmo: Unified vision-language pre-training with mixture-of-modality-experts},
\newblock \bibinfo{journal}{Advances in Neural Information Processing Systems} \bibinfo{volume}{35} (\bibinfo{year}{2022}) \bibinfo{pages}{32897--32912}.
\bibitem[{Wu et~al.(2016)Wu, Schuster, Chen, Le, Norouzi, Macherey, Krikun, Cao, Gao, Macherey et~al.}]{wu2016googleWordPiece}
\bibinfo{author}{Y.~Wu}, \bibinfo{author}{M.~Schuster}, \bibinfo{author}{Z.~Chen}, \bibinfo{author}{Q.~V. Le}, \bibinfo{author}{M.~Norouzi}, \bibinfo{author}{W.~Macherey}, \bibinfo{author}{M.~Krikun}, \bibinfo{author}{Y.~Cao}, \bibinfo{author}{Q.~Gao}, \bibinfo{author}{K.~Macherey}, et~al.,
\newblock \bibinfo{title}{Google's neural machine translation system: Bridging the gap between human and machine translation},
\newblock \bibinfo{journal}{arXiv preprint arXiv:1609.08144}  (\bibinfo{year}{2016}).
\bibitem[{Wang et~al.(2022)Wang, Yang, Men, Lin, Bai, Li, Ma, Zhou, Zhou, and Yang}]{wang2022ofa}
\bibinfo{author}{P.~Wang}, \bibinfo{author}{A.~Yang}, \bibinfo{author}{R.~Men}, \bibinfo{author}{J.~Lin}, \bibinfo{author}{S.~Bai}, \bibinfo{author}{Z.~Li}, \bibinfo{author}{J.~Ma}, \bibinfo{author}{C.~Zhou}, \bibinfo{author}{J.~Zhou}, \bibinfo{author}{H.~Yang},
\newblock \bibinfo{title}{Ofa: Unifying architectures, tasks, and modalities through a simple sequence-to-sequence learning framework},
\newblock in: \bibinfo{booktitle}{International Conference on Machine Learning}, \bibinfo{organization}{PMLR}, \bibinfo{year}{2022}, pp. \bibinfo{pages}{23318--23340}.
\bibitem[{Lu et~al.(2022)Lu, Clark, Zellers, Mottaghi, and Kembhavi}]{lu2022unified}
\bibinfo{author}{J.~Lu}, \bibinfo{author}{C.~Clark}, \bibinfo{author}{R.~Zellers}, \bibinfo{author}{R.~Mottaghi}, \bibinfo{author}{A.~Kembhavi},
\newblock \bibinfo{title}{Unified-io: A unified model for vision, language, and multi-modal tasks},
\newblock \bibinfo{journal}{arXiv preprint arXiv:2206.08916}  (\bibinfo{year}{2022}).
\bibitem[{Kudo and Richardson(2018)}]{kudo2018sentencepiece}
\bibinfo{author}{T.~Kudo}, \bibinfo{author}{J.~Richardson},
\newblock \bibinfo{title}{Sentencepiece: A simple and language independent subword tokenizer and detokenizer for neural text processing},
\newblock \bibinfo{journal}{arXiv preprint arXiv:1808.06226}  (\bibinfo{year}{2018}).
\bibitem[{Esser et~al.(2021)Esser, Rombach, and Ommer}]{esser2021taming}
\bibinfo{author}{P.~Esser}, \bibinfo{author}{R.~Rombach}, \bibinfo{author}{B.~Ommer},
\newblock \bibinfo{title}{Taming transformers for high-resolution image synthesis},
\newblock in: \bibinfo{booktitle}{Proceedings of the IEEE/CVF conference on computer vision and pattern recognition}, \bibinfo{year}{2021}, pp. \bibinfo{pages}{12873--12883}.
\bibitem[{Xue et~al.(2020)Xue, Constant, Roberts, Kale, Al-Rfou, Siddhant, Barua, and Raffel}]{xue2020mt5}
\bibinfo{author}{L.~Xue}, \bibinfo{author}{N.~Constant}, \bibinfo{author}{A.~Roberts}, \bibinfo{author}{M.~Kale}, \bibinfo{author}{R.~Al-Rfou}, \bibinfo{author}{A.~Siddhant}, \bibinfo{author}{A.~Barua}, \bibinfo{author}{C.~Raffel},
\newblock \bibinfo{title}{mt5: A massively multilingual pre-trained text-to-text transformer},
\newblock \bibinfo{journal}{arXiv preprint arXiv:2010.11934}  (\bibinfo{year}{2020}).
\bibitem[{Xie et~al.(2017)Xie, Girshick, Doll{\'a}r, Tu, and He}]{xie2017aggregatedResNext}
\bibinfo{author}{S.~Xie}, \bibinfo{author}{R.~Girshick}, \bibinfo{author}{P.~Doll{\'a}r}, \bibinfo{author}{Z.~Tu}, \bibinfo{author}{K.~He},
\newblock \bibinfo{title}{Aggregated residual transformations for deep neural networks},
\newblock in: \bibinfo{booktitle}{Proceedings of the IEEE conference on computer vision and pattern recognition}, \bibinfo{year}{2017}, pp. \bibinfo{pages}{1492--1500}.
\bibitem[{Ilievski et~al.(2016)Ilievski, Yan, and Feng}]{ilievski2016focused}
\bibinfo{author}{I.~Ilievski}, \bibinfo{author}{S.~Yan}, \bibinfo{author}{J.~Feng},
\newblock \bibinfo{title}{A focused dynamic attention model for visual question answering},
\newblock \bibinfo{journal}{arXiv preprint arXiv:1604.01485}  (\bibinfo{year}{2016}).
\bibitem[{Lu et~al.(2015)Lu, Lin, Batra, and Parikh}]{lu2015deeper}
\bibinfo{author}{J.~Lu}, \bibinfo{author}{X.~Lin}, \bibinfo{author}{D.~Batra}, \bibinfo{author}{D.~Parikh},
\newblock \bibinfo{title}{Deeper lstm and normalized cnn visual question answering model},
\newblock \bibinfo{journal}{GitHub repository} \bibinfo{volume}{6} (\bibinfo{year}{2015}).
\bibitem[{Wu et~al.(2016)Wu, Wang, Shen, Dick, and Van Den~Hengel}]{wu2016ask}
\bibinfo{author}{Q.~Wu}, \bibinfo{author}{P.~Wang}, \bibinfo{author}{C.~Shen}, \bibinfo{author}{A.~Dick}, \bibinfo{author}{A.~Van Den~Hengel},
\newblock \bibinfo{title}{Ask me anything: Free-form visual question answering based on knowledge from external sources},
\newblock in: \bibinfo{booktitle}{Proceedings of the IEEE conference on computer vision and pattern recognition}, \bibinfo{year}{2016}, pp. \bibinfo{pages}{4622--4630}.
\bibitem[{Lu et~al.(2018)Lu, Li, Zhang, Wang, and Wang}]{lu2018co}
\bibinfo{author}{P.~Lu}, \bibinfo{author}{H.~Li}, \bibinfo{author}{W.~Zhang}, \bibinfo{author}{J.~Wang}, \bibinfo{author}{X.~Wang},
\newblock \bibinfo{title}{Co-attending free-form regions and detections with multi-modal multiplicative feature embedding for visual question answering},
\newblock in: \bibinfo{booktitle}{Proceedings of the AAAI conference on artificial intelligence}, volume \bibinfo{volume}{32-1}, \bibinfo{year}{2018}, pp. \bibinfo{pages}{7218–--7225}.
\bibitem[{Wang et~al.(2017)Wang, Wu, Shen, and van~den Hengel}]{wang2017vqa}
\bibinfo{author}{P.~Wang}, \bibinfo{author}{Q.~Wu}, \bibinfo{author}{C.~Shen}, \bibinfo{author}{A.~van~den Hengel},
\newblock \bibinfo{title}{The vqa-machine: Learning how to use existing vision algorithms to answer new questions},
\newblock in: \bibinfo{booktitle}{Proceedings of the IEEE Conference on Computer Vision and Pattern Recognition}, \bibinfo{year}{2017}, pp. \bibinfo{pages}{1173--1182}.
\bibitem[{Kumar et~al.(2016)Kumar, Irsoy, Ondruska, Iyyer, Bradbury, Gulrajani, Zhong, Paulus, and Socher}]{kumar2016ask}
\bibinfo{author}{A.~Kumar}, \bibinfo{author}{O.~Irsoy}, \bibinfo{author}{P.~Ondruska}, \bibinfo{author}{M.~Iyyer}, \bibinfo{author}{J.~Bradbury}, \bibinfo{author}{I.~Gulrajani}, \bibinfo{author}{V.~Zhong}, \bibinfo{author}{R.~Paulus}, \bibinfo{author}{R.~Socher},
\newblock \bibinfo{title}{Ask me anything: Dynamic memory networks for natural language processing},
\newblock in: \bibinfo{booktitle}{International conference on machine learning}, \bibinfo{organization}{PMLR}, \bibinfo{year}{2016}, pp. \bibinfo{pages}{1378--1387}.
\bibitem[{Gao et~al.(2018)Gao, Li, Li, Lu, Li, Hoi, and Wang}]{gao2018question}
\bibinfo{author}{P.~Gao}, \bibinfo{author}{H.~Li}, \bibinfo{author}{S.~Li}, \bibinfo{author}{P.~Lu}, \bibinfo{author}{Y.~Li}, \bibinfo{author}{S.~C. Hoi}, \bibinfo{author}{X.~Wang},
\newblock \bibinfo{title}{Question-guided hybrid convolution for visual question answering},
\newblock in: \bibinfo{booktitle}{Proceedings of the European Conference on Computer Vision (ECCV)}, \bibinfo{year}{2018}, pp. \bibinfo{pages}{469--485}.
\bibitem[{Andreas et~al.(2016)Andreas, Rohrbach, Darrell, and Klein}]{andreas2016learning}
\bibinfo{author}{J.~Andreas}, \bibinfo{author}{M.~Rohrbach}, \bibinfo{author}{T.~Darrell}, \bibinfo{author}{D.~Klein},
\newblock \bibinfo{title}{Learning to compose neural networks for question answering},
\newblock \bibinfo{journal}{arXiv preprint arXiv:1601.01705}  (\bibinfo{year}{2016}).
\bibitem[{Huang et~al.(2020)Huang, Zeng, Liu, Fu, and Fu}]{huang2020pixel}
\bibinfo{author}{Z.~Huang}, \bibinfo{author}{Z.~Zeng}, \bibinfo{author}{B.~Liu}, \bibinfo{author}{D.~Fu}, \bibinfo{author}{J.~Fu},
\newblock \bibinfo{title}{Pixel-bert: Aligning image pixels with text by deep multi-modal transformers},
\newblock \bibinfo{journal}{arXiv preprint arXiv:2004.00849}  (\bibinfo{year}{2020}).
\bibitem[{Gan et~al.(2020)Gan, Chen, Li, Zhu, Cheng, and Liu}]{gan2020large}
\bibinfo{author}{Z.~Gan}, \bibinfo{author}{Y.-C. Chen}, \bibinfo{author}{L.~Li}, \bibinfo{author}{C.~Zhu}, \bibinfo{author}{Y.~Cheng}, \bibinfo{author}{J.~Liu},
\newblock \bibinfo{title}{Large-scale adversarial training for vision-and-language representation learning},
\newblock \bibinfo{journal}{Advances in Neural Information Processing Systems} \bibinfo{volume}{33} (\bibinfo{year}{2020}) \bibinfo{pages}{6616--6628}.
\bibitem[{Li et~al.(2020)Li, Gao, Niu, Xiao, Liu, Liu, Wu, and Wang}]{li2020unimo}
\bibinfo{author}{W.~Li}, \bibinfo{author}{C.~Gao}, \bibinfo{author}{G.~Niu}, \bibinfo{author}{X.~Xiao}, \bibinfo{author}{H.~Liu}, \bibinfo{author}{J.~Liu}, \bibinfo{author}{H.~Wu}, \bibinfo{author}{H.~Wang},
\newblock \bibinfo{title}{Unimo: Towards unified-modal understanding and generation via cross-modal contrastive learning},
\newblock \bibinfo{journal}{arXiv preprint arXiv:2012.15409}  (\bibinfo{year}{2020}).
\bibitem[{Zhang et~al.(2021)Zhang, Li, Hu, Yang, Zhang, Wang, Choi, and Gao}]{zhang2021vinvl}
\bibinfo{author}{P.~Zhang}, \bibinfo{author}{X.~Li}, \bibinfo{author}{X.~Hu}, \bibinfo{author}{J.~Yang}, \bibinfo{author}{L.~Zhang}, \bibinfo{author}{L.~Wang}, \bibinfo{author}{Y.~Choi}, \bibinfo{author}{J.~Gao},
\newblock \bibinfo{title}{Vinvl: Revisiting visual representations in vision-language models},
\newblock in: \bibinfo{booktitle}{Proceedings of the IEEE/CVF conference on computer vision and pattern recognition}, \bibinfo{year}{2021}, pp. \bibinfo{pages}{5579--5588}.
\bibitem[{Dou et~al.(2022)Dou, Xu, Gan, Wang, Wang, Wang, Zhu, Zhang, Yuan, Peng et~al.}]{dou2022empirical}
\bibinfo{author}{Z.-Y. Dou}, \bibinfo{author}{Y.~Xu}, \bibinfo{author}{Z.~Gan}, \bibinfo{author}{J.~Wang}, \bibinfo{author}{S.~Wang}, \bibinfo{author}{L.~Wang}, \bibinfo{author}{C.~Zhu}, \bibinfo{author}{P.~Zhang}, \bibinfo{author}{L.~Yuan}, \bibinfo{author}{N.~Peng}, et~al.,
\newblock \bibinfo{title}{An empirical study of training end-to-end vision-and-language transformers},
\newblock in: \bibinfo{booktitle}{Proceedings of the IEEE/CVF Conference on Computer Vision and Pattern Recognition}, \bibinfo{year}{2022}, pp. \bibinfo{pages}{18166--18176}.
\bibitem[{Wang et~al.(2022)Wang, Yang, Hu, Li, Lin, Gan, Liu, Liu, and Wang}]{wang2022git}
\bibinfo{author}{J.~Wang}, \bibinfo{author}{Z.~Yang}, \bibinfo{author}{X.~Hu}, \bibinfo{author}{L.~Li}, \bibinfo{author}{K.~Lin}, \bibinfo{author}{Z.~Gan}, \bibinfo{author}{Z.~Liu}, \bibinfo{author}{C.~Liu}, \bibinfo{author}{L.~Wang},
\newblock \bibinfo{title}{Git: A generative image-to-text transformer for vision and language},
\newblock \bibinfo{journal}{arXiv preprint arXiv:2205.14100}  (\bibinfo{year}{2022}).
\bibitem[{Wang et~al.(2021)Wang, Yu, Yu, Dai, Tsvetkov, and Cao}]{wang2021simvlm}
\bibinfo{author}{Z.~Wang}, \bibinfo{author}{J.~Yu}, \bibinfo{author}{A.~W. Yu}, \bibinfo{author}{Z.~Dai}, \bibinfo{author}{Y.~Tsvetkov}, \bibinfo{author}{Y.~Cao},
\newblock \bibinfo{title}{Simvlm: Simple visual language model pretraining with weak supervision},
\newblock \bibinfo{journal}{arXiv preprint arXiv:2108.10904}  (\bibinfo{year}{2021}).
\bibitem[{Yuan et~al.(2021)Yuan, Chen, Chen, Codella, Dai, Gao, Hu, Huang, Li, Li et~al.}]{yuan2021florence}
\bibinfo{author}{L.~Yuan}, \bibinfo{author}{D.~Chen}, \bibinfo{author}{Y.-L. Chen}, \bibinfo{author}{N.~Codella}, \bibinfo{author}{X.~Dai}, \bibinfo{author}{J.~Gao}, \bibinfo{author}{H.~Hu}, \bibinfo{author}{X.~Huang}, \bibinfo{author}{B.~Li}, \bibinfo{author}{C.~Li}, et~al.,
\newblock \bibinfo{title}{Florence: A new foundation model for computer vision},
\newblock \bibinfo{journal}{arXiv preprint arXiv:2111.11432}  (\bibinfo{year}{2021}).
\bibitem[{Li et~al.(2022)Li, Xu, Tian, Wang, Yan, Bi, Ye, Chen, Xu, Cao et~al.}]{li2022mplug}
\bibinfo{author}{C.~Li}, \bibinfo{author}{H.~Xu}, \bibinfo{author}{J.~Tian}, \bibinfo{author}{W.~Wang}, \bibinfo{author}{M.~Yan}, \bibinfo{author}{B.~Bi}, \bibinfo{author}{J.~Ye}, \bibinfo{author}{H.~Chen}, \bibinfo{author}{G.~Xu}, \bibinfo{author}{Z.~Cao}, et~al.,
\newblock \bibinfo{title}{mplug: Effective and efficient vision-language learning by cross-modal skip-connections},
\newblock \bibinfo{journal}{arXiv preprint arXiv:2205.12005}  (\bibinfo{year}{2022}).
\bibitem[{Yu et~al.(2022)Yu, Wang, Vasudevan, Yeung, Seyedhosseini, and Wu}]{yu2022coca}
\bibinfo{author}{J.~Yu}, \bibinfo{author}{Z.~Wang}, \bibinfo{author}{V.~Vasudevan}, \bibinfo{author}{L.~Yeung}, \bibinfo{author}{M.~Seyedhosseini}, \bibinfo{author}{Y.~Wu},
\newblock \bibinfo{title}{Coca: Contrastive captioners are image-text foundation models},
\newblock \bibinfo{journal}{arXiv preprint arXiv:2205.01917}  (\bibinfo{year}{2022}).
\bibitem[{Li et~al.(2023)Li, Li, Savarese, and Hoi}]{li2023blip2}
\bibinfo{author}{J.~Li}, \bibinfo{author}{D.~Li}, \bibinfo{author}{S.~Savarese}, \bibinfo{author}{S.~Hoi},
\newblock \bibinfo{title}{Blip-2: Bootstrapping language-image pre-training with frozen image encoders and large language models},
\newblock \bibinfo{journal}{arXiv preprint arXiv:2301.12597}  (\bibinfo{year}{2023}).
\bibitem[{Wang et~al.(2023)Wang, Wang, Lin, Bai, Zhou, Zhou, Wang, and Zhou}]{wang2023onePeace}
\bibinfo{author}{P.~Wang}, \bibinfo{author}{S.~Wang}, \bibinfo{author}{J.~Lin}, \bibinfo{author}{S.~Bai}, \bibinfo{author}{X.~Zhou}, \bibinfo{author}{J.~Zhou}, \bibinfo{author}{X.~Wang}, \bibinfo{author}{C.~Zhou},
\newblock \bibinfo{title}{One-peace: Exploring one general representation model toward unlimited modalities},
\newblock \bibinfo{journal}{arXiv preprint arXiv:2305.11172}  (\bibinfo{year}{2023}).
\bibitem[{Wang et~al.(2022)Wang, Bao, Dong, Bjorck, Peng, Liu, Aggarwal, Mohammed, Singhal, Som et~al.}]{wang2022imageBeit}
\bibinfo{author}{W.~Wang}, \bibinfo{author}{H.~Bao}, \bibinfo{author}{L.~Dong}, \bibinfo{author}{J.~Bjorck}, \bibinfo{author}{Z.~Peng}, \bibinfo{author}{Q.~Liu}, \bibinfo{author}{K.~Aggarwal}, \bibinfo{author}{O.~K. Mohammed}, \bibinfo{author}{S.~Singhal}, \bibinfo{author}{S.~Som}, et~al.,
\newblock \bibinfo{title}{Image as a foreign language: Beit pretraining for all vision and vision-language tasks},
\newblock \bibinfo{journal}{arXiv preprint arXiv:2208.10442}  (\bibinfo{year}{2022}).
\bibitem[{Papineni et~al.(2002)Papineni, Roukos, Ward, and Zhu}]{papineni2002bleu}
\bibinfo{author}{K.~Papineni}, \bibinfo{author}{S.~Roukos}, \bibinfo{author}{T.~Ward}, \bibinfo{author}{W.-J. Zhu},
\newblock \bibinfo{title}{Bleu: a method for automatic evaluation of machine translation},
\newblock in: \bibinfo{booktitle}{Proceedings of the 40th annual meeting of the Association for Computational Linguistics}, \bibinfo{year}{2002}, pp. \bibinfo{pages}{311--318}.
\bibitem[{Wang et~al.(2023)Wang, Lin, Zhang, Lei, and Shou}]{wang2023too}
\bibinfo{author}{A.~J. Wang}, \bibinfo{author}{K.~Q. Lin}, \bibinfo{author}{D.~J. Zhang}, \bibinfo{author}{S.~W. Lei}, \bibinfo{author}{M.~Z. Shou},
\newblock \bibinfo{title}{Too large; data reduction for vision-language pre-training},
\newblock \bibinfo{journal}{arXiv preprint arXiv:2305.20087}  (\bibinfo{year}{2023}).
\bibitem[{Thapliyal et~al.(2022)Thapliyal, Pont-Tuset, Chen, and Soricut}]{thapliyal2022crossmodal}
\bibinfo{author}{A.~V. Thapliyal}, \bibinfo{author}{J.~Pont-Tuset}, \bibinfo{author}{X.~Chen}, \bibinfo{author}{R.~Soricut},
\newblock \bibinfo{title}{Crossmodal-3600: A massively multilingual multimodal evaluation dataset},
\newblock \bibinfo{journal}{arXiv preprint arXiv:2205.12522}  (\bibinfo{year}{2022}).
\bibitem[{Lau et~al.(2018)Lau, Gayen, Ben~Abacha, and Demner-Fushman}]{lau2018datasetVQARAD}
\bibinfo{author}{J.~J. Lau}, \bibinfo{author}{S.~Gayen}, \bibinfo{author}{A.~Ben~Abacha}, \bibinfo{author}{D.~Demner-Fushman},
\newblock \bibinfo{title}{A dataset of clinically generated visual questions and answers about radiology images},
\newblock \bibinfo{journal}{Scientific data} \bibinfo{volume}{5} (\bibinfo{year}{2018}) \bibinfo{pages}{1--10}.
\bibitem[{He et~al.(2020)He, Zhang, Mou, Xing, and Xie}]{he2020pathvqa}
\bibinfo{author}{X.~He}, \bibinfo{author}{Y.~Zhang}, \bibinfo{author}{L.~Mou}, \bibinfo{author}{E.~Xing}, \bibinfo{author}{P.~Xie}, \bibinfo{title}{Pathvqa: 30000+ questions for medical visual question answering}, \bibinfo{year}{2020}. \href{http://arxiv.org/abs/2003.10286}{{\tt arXiv:2003.10286}}.
\bibitem[{Zhang et~al.(2023)Zhang, Yu, Yan, Liu, Adhikarla, Fu, Chen, Chen, Zhou, Li et~al.}]{zhang2023biomedgpt}
\bibinfo{author}{K.~Zhang}, \bibinfo{author}{J.~Yu}, \bibinfo{author}{Z.~Yan}, \bibinfo{author}{Y.~Liu}, \bibinfo{author}{E.~Adhikarla}, \bibinfo{author}{S.~Fu}, \bibinfo{author}{X.~Chen}, \bibinfo{author}{C.~Chen}, \bibinfo{author}{Y.~Zhou}, \bibinfo{author}{X.~Li}, et~al.,
\newblock \bibinfo{title}{Biomedgpt: A unified and generalist biomedical generative pre-trained transformer for vision, language, and multimodal tasks},
\newblock \bibinfo{journal}{arXiv preprint arXiv:2305.17100}  (\bibinfo{year}{2023}).
\bibitem[{Li et~al.(2023{\natexlab{a}})Li, Wong, Zhang, Usuyama, Liu, Yang, Naumann, Poon, and Gao}]{li2023llava}
\bibinfo{author}{C.~Li}, \bibinfo{author}{C.~Wong}, \bibinfo{author}{S.~Zhang}, \bibinfo{author}{N.~Usuyama}, \bibinfo{author}{H.~Liu}, \bibinfo{author}{J.~Yang}, \bibinfo{author}{T.~Naumann}, \bibinfo{author}{H.~Poon}, \bibinfo{author}{J.~Gao},
\newblock \bibinfo{title}{Llava-med: Training a large language-and-vision assistant for biomedicine in one day},
\newblock \bibinfo{journal}{arXiv preprint arXiv:2306.00890}  (\bibinfo{year}{2023}{\natexlab{a}}).
\bibitem[{Li et~al.(2023{\natexlab{b}})Li, Du, Zhou, Wang, Zhao, and Wen}]{li2023evaluatingHallucination}
\bibinfo{author}{Y.~Li}, \bibinfo{author}{Y.~Du}, \bibinfo{author}{K.~Zhou}, \bibinfo{author}{J.~Wang}, \bibinfo{author}{W.~X. Zhao}, \bibinfo{author}{J.-R. Wen},
\newblock \bibinfo{title}{Evaluating object hallucination in large vision-language models},
\newblock \bibinfo{journal}{arXiv preprint arXiv:2305.10355}  (\bibinfo{year}{2023}{\natexlab{b}}).
\bibitem[{Gupta et~al.(2022)Gupta, Marten, Kembhavi, and Hoiem}]{gupta2022grit}
\bibinfo{author}{T.~Gupta}, \bibinfo{author}{R.~Marten}, \bibinfo{author}{A.~Kembhavi}, \bibinfo{author}{D.~Hoiem},
\newblock \bibinfo{title}{Grit: General robust image task benchmark},
\newblock \bibinfo{journal}{arXiv preprint arXiv:2204.13653}  (\bibinfo{year}{2022}).
\bibitem[{Huang et~al.(2019)Huang, Alfadly, Ghanem, and Worring}]{huang2019assessing}
\bibinfo{author}{J.-H. Huang}, \bibinfo{author}{M.~Alfadly}, \bibinfo{author}{B.~Ghanem}, \bibinfo{author}{M.~Worring},
\newblock \bibinfo{title}{Assessing the robustness of visual question answering},
\newblock \bibinfo{journal}{ArXiv} \bibinfo{volume}{abs/1912.01452} (\bibinfo{year}{2019}). \URLprefix \url{https://api.semanticscholar.org/CorpusID:208548469}.
\bibitem[{Jimenez et~al.(2022)Jimenez, Russakovsky, and Narasimhan}]{jimenez2022carets}
\bibinfo{author}{C.~E. Jimenez}, \bibinfo{author}{O.~Russakovsky}, \bibinfo{author}{K.~Narasimhan},
\newblock \bibinfo{title}{Carets: A consistency and robustness evaluative test suite for vqa},
\newblock \bibinfo{journal}{arXiv preprint arXiv:2203.07613}  (\bibinfo{year}{2022}).
\bibitem[{Zhao et~al.(2023)Zhao, Pang, Du, Yang, Li, Cheung, and Lin}]{zhao2023evaluatingRobustnessVLP}
\bibinfo{author}{Y.~Zhao}, \bibinfo{author}{T.~Pang}, \bibinfo{author}{C.~Du}, \bibinfo{author}{X.~Yang}, \bibinfo{author}{C.~Li}, \bibinfo{author}{N.-M. Cheung}, \bibinfo{author}{M.~Lin},
\newblock \bibinfo{title}{On evaluating adversarial robustness of large vision-language models},
\newblock \bibinfo{journal}{arXiv preprint arXiv:2305.16934}  (\bibinfo{year}{2023}).
\bibitem[{Chao et~al.(2018)Chao, Hu, and Sha}]{chao2018crossDataset}
\bibinfo{author}{W.-L. Chao}, \bibinfo{author}{H.~Hu}, \bibinfo{author}{F.~Sha},
\newblock \bibinfo{title}{Cross-dataset adaptation for visual question answering},
\newblock in: \bibinfo{booktitle}{Proceedings of the IEEE Conference on Computer Vision and Pattern Recognition}, \bibinfo{year}{2018}, pp. \bibinfo{pages}{5716--5725}.
\bibitem[{Li et~al.(2018)Li, Fu, Yu, Mei, and Luo}]{li2018tellExplainableVQA}
\bibinfo{author}{Q.~Li}, \bibinfo{author}{J.~Fu}, \bibinfo{author}{D.~Yu}, \bibinfo{author}{T.~Mei}, \bibinfo{author}{J.~Luo},
\newblock \bibinfo{title}{Tell-and-answer: Towards explainable visual question answering using attributes and captions},
\newblock \bibinfo{journal}{arXiv preprint arXiv:1801.09041}  (\bibinfo{year}{2018}).
\bibitem[{Goyal et~al.(2016)Goyal, Mohapatra, Parikh, and Batra}]{goyal2016towardsExlainableVQA}
\bibinfo{author}{Y.~Goyal}, \bibinfo{author}{A.~Mohapatra}, \bibinfo{author}{D.~Parikh}, \bibinfo{author}{D.~Batra},
\newblock \bibinfo{title}{Towards transparent ai systems: Interpreting visual question answering models},
\newblock \bibinfo{journal}{arXiv preprint arXiv:1608.08974}  (\bibinfo{year}{2016}).
\bibitem[{Goertzel and Pennachin(2007)}]{goertzel2007artificial}
\bibinfo{author}{B.~Goertzel}, \bibinfo{author}{C.~Pennachin}, \bibinfo{title}{Artificial general intelligence}, volume~\bibinfo{volume}{2}, \bibinfo{publisher}{Springer}, \bibinfo{year}{2007}.
\bibitem[{Farazi et~al.(2020)Farazi, Khan, and Barnes}]{farazi2020known}
\bibinfo{author}{M.~R. Farazi}, \bibinfo{author}{S.~H. Khan}, \bibinfo{author}{N.~Barnes},
\newblock \bibinfo{title}{From known to the unknown: Transferring knowledge to answer questions about novel visual and semantic concepts},
\newblock \bibinfo{journal}{Image and Vision Computing} \bibinfo{volume}{103} (\bibinfo{year}{2020}) \bibinfo{pages}{103985}.
\bibitem[{Jin et~al.(2021)Jin, Cheng, Shen, Chen, and Ren}]{jin2021good}
\bibinfo{author}{W.~Jin}, \bibinfo{author}{Y.~Cheng}, \bibinfo{author}{Y.~Shen}, \bibinfo{author}{W.~Chen}, \bibinfo{author}{X.~Ren},
\newblock \bibinfo{title}{A good prompt is worth millions of parameters: Low-resource prompt-based learning for vision-language models},
\newblock \bibinfo{journal}{arXiv preprint arXiv:2110.08484}  (\bibinfo{year}{2021}).
\bibitem[{Chuang et~al.(2019)Chuang, Liu, Lee, and Lee}]{chuang2019speechbert}
\bibinfo{author}{Y.-S. Chuang}, \bibinfo{author}{C.-L. Liu}, \bibinfo{author}{H.-Y. Lee}, \bibinfo{author}{L.-s. Lee},
\newblock \bibinfo{title}{Speechbert: An audio-and-text jointly learned language model for end-to-end spoken question answering},
\newblock \bibinfo{journal}{arXiv preprint arXiv:1910.11559}  (\bibinfo{year}{2019}).
\bibitem[{Drossos et~al.(2020)Drossos, Lipping, and Virtanen}]{drossos2020clothoAudioCaptioning}
\bibinfo{author}{K.~Drossos}, \bibinfo{author}{S.~Lipping}, \bibinfo{author}{T.~Virtanen},
\newblock \bibinfo{title}{Clotho: An audio captioning dataset},
\newblock in: \bibinfo{booktitle}{ICASSP 2020-2020 IEEE International Conference on Acoustics, Speech and Signal Processing (ICASSP)}, \bibinfo{organization}{IEEE}, \bibinfo{year}{2020}, pp. \bibinfo{pages}{736--740}.
\bibitem[{Iashin and Rahtu(2020)}]{iashin2020multiVideoCaptioning}
\bibinfo{author}{V.~Iashin}, \bibinfo{author}{E.~Rahtu},
\newblock \bibinfo{title}{Multi-modal dense video captioning},
\newblock in: \bibinfo{booktitle}{Proceedings of the IEEE/CVF conference on computer vision and pattern recognition workshops}, \bibinfo{year}{2020}, pp. \bibinfo{pages}{958--959}.
\bibitem[{Maaz et~al.(2023)Maaz, Rasheed, Khan, and Khan}]{maaz2023video}
\bibinfo{author}{M.~Maaz}, \bibinfo{author}{H.~Rasheed}, \bibinfo{author}{S.~Khan}, \bibinfo{author}{F.~S. Khan},
\newblock \bibinfo{title}{Video-chatgpt: Towards detailed video understanding via large vision and language models},
\newblock \bibinfo{journal}{arXiv preprint arXiv:2306.05424}  (\bibinfo{year}{2023}).
\bibitem[{Gan et~al.(2017)Gan, Gan, He, Pu, Tran, Gao, Carin, and Deng}]{gan2017semanticVisualCaptioning}
\bibinfo{author}{Z.~Gan}, \bibinfo{author}{C.~Gan}, \bibinfo{author}{X.~He}, \bibinfo{author}{Y.~Pu}, \bibinfo{author}{K.~Tran}, \bibinfo{author}{J.~Gao}, \bibinfo{author}{L.~Carin}, \bibinfo{author}{L.~Deng},
\newblock \bibinfo{title}{Semantic compositional networks for visual captioning},
\newblock in: \bibinfo{booktitle}{Proceedings of the IEEE conference on computer vision and pattern recognition}, \bibinfo{year}{2017}, pp. \bibinfo{pages}{5630--5639}.
\bibitem[{Gao et~al.(2023)Gao, Chen, Chen, Wang, and Lu}]{gao2023champion}
\bibinfo{author}{S.~Gao}, \bibinfo{author}{Z.~Chen}, \bibinfo{author}{G.~Chen}, \bibinfo{author}{W.~Wang}, \bibinfo{author}{T.~Lu},
\newblock \bibinfo{title}{Champion solution for the wsdm2023 toloka vqa challenge},
\newblock \bibinfo{journal}{arXiv preprint arXiv:2301.09045}  (\bibinfo{year}{2023}).
\bibitem[{Liu et~al.(2018)Liu, Xiang, Hospedales, Yang, and Sun}]{liu2018ivqa}
\bibinfo{author}{F.~Liu}, \bibinfo{author}{T.~Xiang}, \bibinfo{author}{T.~M. Hospedales}, \bibinfo{author}{W.~Yang}, \bibinfo{author}{C.~Sun},
\newblock \bibinfo{title}{ivqa: Inverse visual question answering},
\newblock in: \bibinfo{booktitle}{Proceedings of the IEEE Conference on Computer Vision and Pattern Recognition}, \bibinfo{year}{2018}, pp. \bibinfo{pages}{8611--8619}.
\bibitem[{Mostafazadeh et~al.(2016)Mostafazadeh, Misra, Devlin, Mitchell, He, and Vanderwende}]{mostafazadeh2016generating}
\bibinfo{author}{N.~Mostafazadeh}, \bibinfo{author}{I.~Misra}, \bibinfo{author}{J.~Devlin}, \bibinfo{author}{M.~Mitchell}, \bibinfo{author}{X.~He}, \bibinfo{author}{L.~Vanderwende},
\newblock \bibinfo{title}{Generating natural questions about an image},
\newblock \bibinfo{journal}{arXiv preprint arXiv:1603.06059}  (\bibinfo{year}{2016}).
\bibitem[{Zeng et~al.(2017)Zeng, Chen, Chuang, Liao, Niebles, and Sun}]{zeng2017leveraging}
\bibinfo{author}{K.-H. Zeng}, \bibinfo{author}{T.-H. Chen}, \bibinfo{author}{C.-Y. Chuang}, \bibinfo{author}{Y.-H. Liao}, \bibinfo{author}{J.~C. Niebles}, \bibinfo{author}{M.~Sun},
\newblock \bibinfo{title}{Leveraging video descriptions to learn video question answering},
\newblock in: \bibinfo{booktitle}{Proceedings of the AAAI Conference on Artificial Intelligence}, volume \bibinfo{volume}{31-1}, \bibinfo{year}{2017}, pp. \bibinfo{pages}{4334–--4340}.
\bibitem[{Changpinyo et~al.(2022)Changpinyo, Kukliansky, Szpektor, Chen, Ding, and Soricut}]{changpinyo2022allVQAImageCaptions}
\bibinfo{author}{S.~Changpinyo}, \bibinfo{author}{D.~Kukliansky}, \bibinfo{author}{I.~Szpektor}, \bibinfo{author}{X.~Chen}, \bibinfo{author}{N.~Ding}, \bibinfo{author}{R.~Soricut},
\newblock \bibinfo{title}{All you may need for vqa are image captions},
\newblock \bibinfo{journal}{arXiv preprint arXiv:2205.01883}  (\bibinfo{year}{2022}).
\bibitem[{Hu et~al.(2017)Hu, Lin, Liu, Cheng, Chang, and Sun}]{hu2017deep360}
\bibinfo{author}{H.-N. Hu}, \bibinfo{author}{Y.-C. Lin}, \bibinfo{author}{M.-Y. Liu}, \bibinfo{author}{H.-T. Cheng}, \bibinfo{author}{Y.-J. Chang}, \bibinfo{author}{M.~Sun},
\newblock \bibinfo{title}{Deep 360 pilot: Learning a deep agent for piloting through 360deg sports videos},
\newblock in: \bibinfo{booktitle}{Proceedings of the IEEE Conference on Computer Vision and Pattern Recognition}, \bibinfo{year}{2017}, pp. \bibinfo{pages}{3451--3460}.
\bibitem[{Gordon et~al.(2018)Gordon, Kembhavi, Rastegari, Redmon, Fox, and Farhadi}]{gordon2018iqa}
\bibinfo{author}{D.~Gordon}, \bibinfo{author}{A.~Kembhavi}, \bibinfo{author}{M.~Rastegari}, \bibinfo{author}{J.~Redmon}, \bibinfo{author}{D.~Fox}, \bibinfo{author}{A.~Farhadi},
\newblock \bibinfo{title}{Iqa: Visual question answering in interactive environments},
\newblock in: \bibinfo{booktitle}{Proceedings of the IEEE conference on computer vision and pattern recognition}, \bibinfo{year}{2018}, pp. \bibinfo{pages}{4089--4098}.
\bibitem[{Zhuge et~al.(2021)Zhuge, Gao, Fan, Jin, Chen, Zhou, Qiu, and Shao}]{zhuge2021kaleido}
\bibinfo{author}{M.~Zhuge}, \bibinfo{author}{D.~Gao}, \bibinfo{author}{D.-P. Fan}, \bibinfo{author}{L.~Jin}, \bibinfo{author}{B.~Chen}, \bibinfo{author}{H.~Zhou}, \bibinfo{author}{M.~Qiu}, \bibinfo{author}{L.~Shao},
\newblock \bibinfo{title}{Kaleido-bert: Vision-language pre-training on fashion domain},
\newblock in: \bibinfo{booktitle}{Proceedings of the IEEE/CVF Conference on Computer Vision and Pattern Recognition}, \bibinfo{year}{2021}, pp. \bibinfo{pages}{12647--12657}.
\bibitem[{Ghosal et~al.(2018)Ghosal, Akhtar, Chauhan, Poria, Ekbal, and Bhattacharyya}]{ghosal2018contextual}
\bibinfo{author}{D.~Ghosal}, \bibinfo{author}{M.~S. Akhtar}, \bibinfo{author}{D.~Chauhan}, \bibinfo{author}{S.~Poria}, \bibinfo{author}{A.~Ekbal}, \bibinfo{author}{P.~Bhattacharyya},
\newblock \bibinfo{title}{Contextual inter-modal attention for multi-modal sentiment analysis},
\newblock in: \bibinfo{booktitle}{proceedings of the 2018 conference on empirical methods in natural language processing}, \bibinfo{year}{2018}, pp. \bibinfo{pages}{3454--3466}.
\bibitem[{Wang et~al.(2016)Wang, Yin, Wang, Wu, and Wang}]{wang2016comprehensive}
\bibinfo{author}{K.~Wang}, \bibinfo{author}{Q.~Yin}, \bibinfo{author}{W.~Wang}, \bibinfo{author}{S.~Wu}, \bibinfo{author}{L.~Wang},
\newblock \bibinfo{title}{A comprehensive survey on cross-modal retrieval},
\newblock \bibinfo{journal}{arXiv preprint arXiv:1607.06215}  (\bibinfo{year}{2016}).
\bibitem[{Chen et~al.(2022)Chen, Chen, Shi, Zhang, Chang, and Tian}]{chen2022hivlp}
\bibinfo{author}{F.~Chen}, \bibinfo{author}{X.~Chen}, \bibinfo{author}{J.~Shi}, \bibinfo{author}{D.~Zhang}, \bibinfo{author}{J.~Chang}, \bibinfo{author}{Q.~Tian},
\newblock \bibinfo{title}{Hivlp: Hierarchical vision-language pre-training for fast image-text retrieval},
\newblock \bibinfo{journal}{arXiv preprint arXiv:2205.12105}  (\bibinfo{year}{2022}).
\bibitem[{Specia et~al.(2016)Specia, Frank, Sima’An, and Elliott}]{specia2016shared}
\bibinfo{author}{L.~Specia}, \bibinfo{author}{S.~Frank}, \bibinfo{author}{K.~Sima’An}, \bibinfo{author}{D.~Elliott},
\newblock \bibinfo{title}{A shared task on multimodal machine translation and crosslingual image description},
\newblock in: \bibinfo{booktitle}{Proceedings of the First Conference on Machine Translation: Volume 2, Shared Task Papers}, \bibinfo{year}{2016}, pp. \bibinfo{pages}{543--553}.
\bibitem[{Shi et~al.(2020)Shi, Zhang, Zhang, Chen, and Zhan}]{shi2020change}
\bibinfo{author}{W.~Shi}, \bibinfo{author}{M.~Zhang}, \bibinfo{author}{R.~Zhang}, \bibinfo{author}{S.~Chen}, \bibinfo{author}{Z.~Zhan},
\newblock \bibinfo{title}{Change detection based on artificial intelligence: State-of-the-art and challenges},
\newblock \bibinfo{journal}{Remote Sensing} \bibinfo{volume}{12} (\bibinfo{year}{2020}) \bibinfo{pages}{1688}.
\bibitem[{Yun et~al.(2021)Yun, Yu, Yang, Lee, and Kim}]{yun2021pano}
\bibinfo{author}{H.~Yun}, \bibinfo{author}{Y.~Yu}, \bibinfo{author}{W.~Yang}, \bibinfo{author}{K.~Lee}, \bibinfo{author}{G.~Kim},
\newblock \bibinfo{title}{Pano-avqa: Grounded audio-visual question answering on 360deg videos},
\newblock in: \bibinfo{booktitle}{Proceedings of the IEEE/CVF International Conference on Computer Vision}, \bibinfo{year}{2021}, pp. \bibinfo{pages}{2031--2041}.
\bibitem[{Sharma et~al.(2018)Sharma, Ding, Goodman, and Soricut}]{sharma2018conceptual}
\bibinfo{author}{P.~Sharma}, \bibinfo{author}{N.~Ding}, \bibinfo{author}{S.~Goodman}, \bibinfo{author}{R.~Soricut},
\newblock \bibinfo{title}{Conceptual captions: A cleaned, hypernymed, image alt-text dataset for automatic image captioning},
\newblock in: \bibinfo{booktitle}{Proceedings of the 56th Annual Meeting of the Association for Computational Linguistics (Volume 1: Long Papers)}, \bibinfo{year}{2018}, pp. \bibinfo{pages}{2556--2565}.
\bibitem[{Pfeiffer et~al.(2021)Pfeiffer, Geigle, Kamath, Steitz, Roth, Vuli{\'c}, and Gurevych}]{pfeiffer2021xgqa}
\bibinfo{author}{J.~Pfeiffer}, \bibinfo{author}{G.~Geigle}, \bibinfo{author}{A.~Kamath}, \bibinfo{author}{J.-M.~O. Steitz}, \bibinfo{author}{S.~Roth}, \bibinfo{author}{I.~Vuli{\'c}}, \bibinfo{author}{I.~Gurevych},
\newblock \bibinfo{title}{xgqa: Cross-lingual visual question answering},
\newblock \bibinfo{journal}{arXiv preprint arXiv:2109.06082}  (\bibinfo{year}{2021}).
\bibitem[{Changpinyo et~al.(2022)Changpinyo, Xue, Szpektor, Thapliyal, Amelot, Chen, and Soricut}]{changpinyo2022towardsMAXM}
\bibinfo{author}{S.~Changpinyo}, \bibinfo{author}{L.~Xue}, \bibinfo{author}{I.~Szpektor}, \bibinfo{author}{A.~V. Thapliyal}, \bibinfo{author}{J.~Amelot}, \bibinfo{author}{X.~Chen}, \bibinfo{author}{R.~Soricut},
\newblock \bibinfo{title}{Towards multi-lingual visual question answering},
\newblock \bibinfo{journal}{arXiv preprint arXiv:2209.05401}  (\bibinfo{year}{2022}).
\bibitem[{Liu et~al.(2022)Liu, Pfeiffer, Korhonen, Vulic, and Gurevych}]{liu2022delving}
\bibinfo{author}{C.~Liu}, \bibinfo{author}{J.~Pfeiffer}, \bibinfo{author}{A.~Korhonen}, \bibinfo{author}{I.~Vulic}, \bibinfo{author}{I.~Gurevych},
\newblock \bibinfo{title}{Delving deeper into cross-lingual visual question answering},
\newblock \bibinfo{journal}{arXiv preprint arXiv:2202.07630}  (\bibinfo{year}{2022}).
\bibitem[{Chen et~al.(2022)Chen, Duan, Wang, He, Lu, Dai, and Qiao}]{chen2022visionAdapter}
\bibinfo{author}{Z.~Chen}, \bibinfo{author}{Y.~Duan}, \bibinfo{author}{W.~Wang}, \bibinfo{author}{J.~He}, \bibinfo{author}{T.~Lu}, \bibinfo{author}{J.~Dai}, \bibinfo{author}{Y.~Qiao},
\newblock \bibinfo{title}{Vision transformer adapter for dense predictions},
\newblock \bibinfo{journal}{arXiv preprint arXiv:2205.08534}  (\bibinfo{year}{2022}).
\bibitem[{Yin et~al.(2023)Yin, Fu, Zhao, Li, Sun, Xu, and Chen}]{yin2023survey}
\bibinfo{author}{S.~Yin}, \bibinfo{author}{C.~Fu}, \bibinfo{author}{S.~Zhao}, \bibinfo{author}{K.~Li}, \bibinfo{author}{X.~Sun}, \bibinfo{author}{T.~Xu}, \bibinfo{author}{E.~Chen},
\newblock \bibinfo{title}{A survey on multimodal large language models},
\newblock \bibinfo{journal}{arXiv preprint arXiv:2306.13549}  (\bibinfo{year}{2023}).
\bibitem[{Liu et~al.(2023)Liu, Li, Wu, and Lee}]{liu2023visual}
\bibinfo{author}{H.~Liu}, \bibinfo{author}{C.~Li}, \bibinfo{author}{Q.~Wu}, \bibinfo{author}{Y.~J. Lee},
\newblock \bibinfo{title}{Visual instruction tuning},
\newblock \bibinfo{journal}{arXiv preprint arXiv:2304.08485}  (\bibinfo{year}{2023}).
\bibitem[{Zhang et~al.(2023)Zhang, Han, Zhou, Hu, Yan, Lu, Li, Gao, and Qiao}]{zhang2023llama}
\bibinfo{author}{R.~Zhang}, \bibinfo{author}{J.~Han}, \bibinfo{author}{A.~Zhou}, \bibinfo{author}{X.~Hu}, \bibinfo{author}{S.~Yan}, \bibinfo{author}{P.~Lu}, \bibinfo{author}{H.~Li}, \bibinfo{author}{P.~Gao}, \bibinfo{author}{Y.~Qiao},
\newblock \bibinfo{title}{Llama-adapter: Efficient fine-tuning of language models with zero-init attention},
\newblock \bibinfo{journal}{arXiv preprint arXiv:2303.16199}  (\bibinfo{year}{2023}).
\bibitem[{Gao et~al.(2023)Gao, Han, Zhang, Lin, Geng, Zhou, Zhang, Lu, He, Yue et~al.}]{gao2023llama}
\bibinfo{author}{P.~Gao}, \bibinfo{author}{J.~Han}, \bibinfo{author}{R.~Zhang}, \bibinfo{author}{Z.~Lin}, \bibinfo{author}{S.~Geng}, \bibinfo{author}{A.~Zhou}, \bibinfo{author}{W.~Zhang}, \bibinfo{author}{P.~Lu}, \bibinfo{author}{C.~He}, \bibinfo{author}{X.~Yue}, et~al.,
\newblock \bibinfo{title}{Llama-adapter v2: Parameter-efficient visual instruction model},
\newblock \bibinfo{journal}{arXiv preprint arXiv:2304.15010}  (\bibinfo{year}{2023}).
\bibitem[{Rohrbach et~al.(2018)Rohrbach, Hendricks, Burns, Darrell, and Saenko}]{rohrbach2018objectHallucination}
\bibinfo{author}{A.~Rohrbach}, \bibinfo{author}{L.~A. Hendricks}, \bibinfo{author}{K.~Burns}, \bibinfo{author}{T.~Darrell}, \bibinfo{author}{K.~Saenko},
\newblock \bibinfo{title}{Object hallucination in image captioning},
\newblock \bibinfo{journal}{arXiv preprint arXiv:1809.02156}  (\bibinfo{year}{2018}).
\bibitem[{Kurp(2008)}]{kurp2008green}
\bibinfo{author}{P.~Kurp},
\newblock \bibinfo{title}{Green computing},
\newblock \bibinfo{journal}{Communications of the ACM} \bibinfo{volume}{51} (\bibinfo{year}{2008}) \bibinfo{pages}{11--13}.
\bibitem[{Ahmad et~al.(2021)Ahmad, Zhang, Huang, Zhang, Dai, Song, and Chen}]{ahmad2021artificial}
\bibinfo{author}{T.~Ahmad}, \bibinfo{author}{D.~Zhang}, \bibinfo{author}{C.~Huang}, \bibinfo{author}{H.~Zhang}, \bibinfo{author}{N.~Dai}, \bibinfo{author}{Y.~Song}, \bibinfo{author}{H.~Chen},
\newblock \bibinfo{title}{Artificial intelligence in sustainable energy industry: Status quo, challenges and opportunities},
\newblock \bibinfo{journal}{Journal of Cleaner Production} \bibinfo{volume}{289} (\bibinfo{year}{2021}) \bibinfo{pages}{125834}.

\end{thebibliography}
